\documentclass[10pt,twocolumn,letterpaper]{article}

\usepackage{cvpr}
\usepackage{times}
\usepackage{epsfig}
\usepackage{graphicx}
\usepackage{amsmath}
\usepackage{amssymb}
\usepackage{multirow}
\usepackage{comment}
\usepackage{algorithm}
\usepackage{algorithmic}

\newcommand{\argmin}{\mathop{\rm argmin}\limits}
\newlength\myindent
\usepackage[caption=false]{subfig}

\usepackage[pagebackref=true,breaklinks=true,letterpaper=true,colorlinks,bookmarks=false]{hyperref}

\cvprfinalcopy 

\def\cvprPaperID{5990} 

\ifcvprfinal\fi
\begin{document}

\title{Dehazing Cost Volume for Deep Multi-view Stereo in Scattering Media with Airlight and Scattering Coefficient Estimation}

\author{
Yuki Fujimura \quad Motoharu Sonogashira \quad Masaaki Iiyama \\
Kyoto Univeristy\\
{\tt\small \{fujimura, sonogashira, iiyama\}@mm.media.kyoto-u.ac.jp}
}

\maketitle

\begin{abstract}
We propose a learning-based multi-view stereo (MVS) method in scattering media, such as fog or smoke, with a novel cost volume, called the dehazing cost volume.
Images captured in scattering media are degraded due to light scattering and attenuation caused by suspended particles.
This degradation depends on scene depth; thus, it is difficult for traditional MVS methods to evaluate photometric consistency because the depth is unknown before three-dimensional (3D) reconstruction.
The dehazing cost volume can solve this chicken-and-egg problem of depth estimation and image restoration by computing the scattering effect using swept planes in the cost volume.
We also propose a method of estimating scattering parameters, such as airlight, and a scattering coefficient, which are required for our dehazing cost volume.
The output depth of a network with our dehazing cost volume can be regarded as a function of these parameters; thus, they are geometrically optimized with a sparse 3D point cloud obtained at a structure-from-motion step.
Experimental results on synthesized hazy images indicate the effectiveness of our dehazing cost volume against the ordinary cost volume regarding scattering media. We also demonstrated the applicability of our dehazing cost volume to real foggy scenes.
\end{abstract}

\section{Introduction}
\begin{figure}[tb]
\centering
\subfloat[]{\includegraphics[width=0.14\textwidth]{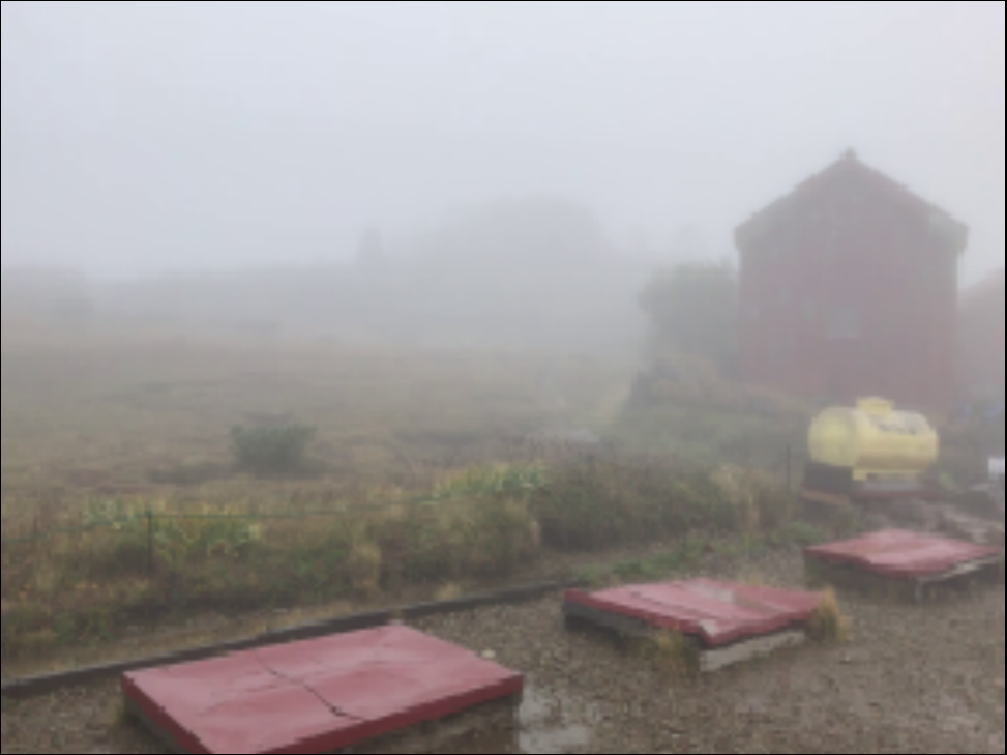}} \,
\subfloat[]{\includegraphics[width=0.14\textwidth]{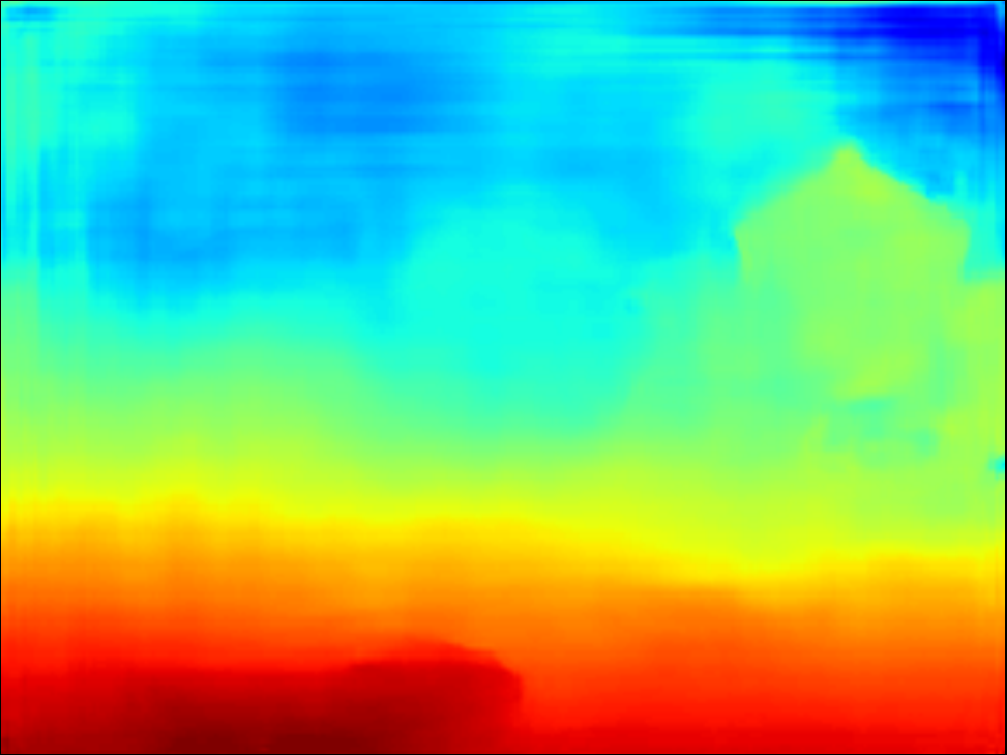}} \,
\subfloat[]{\includegraphics[width=0.14\textwidth]{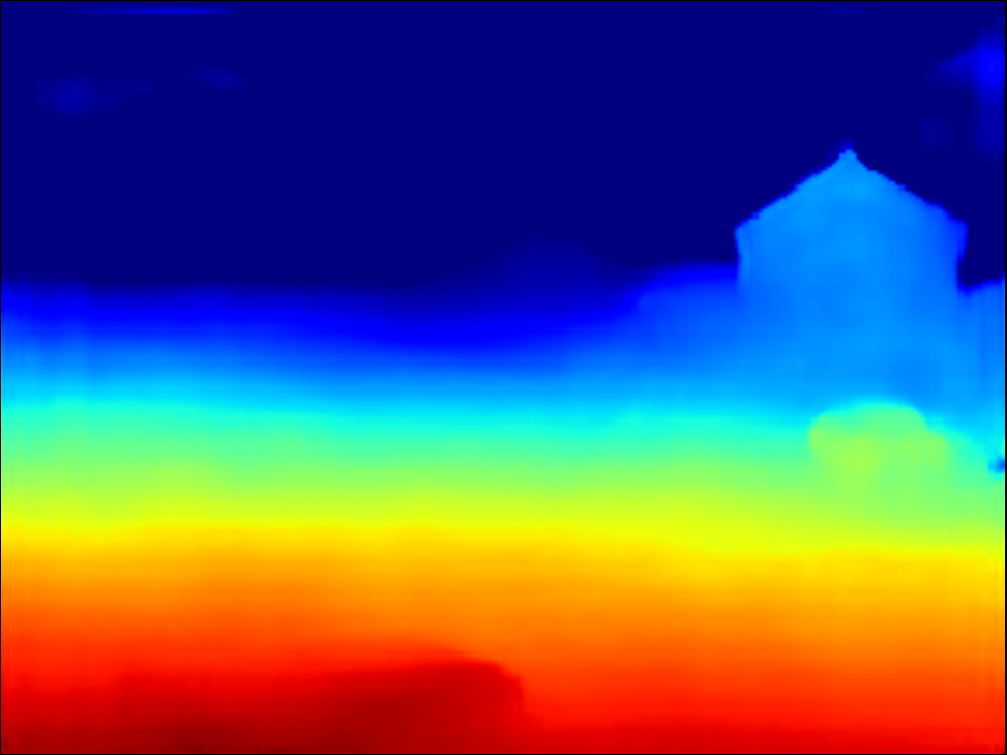}}
\caption{(a) Image captured in actual foggy scene. (b) Output depth of fine-tuned MVDepthNet \cite{wang18} with ordinary cost volume. (c) Output depth of network with our dehazing cost volume.}
\label{fig:cover_figure}
\end{figure}

Three-dimensional (3D) reconstruction from 2D images is important in computer vision.
However, images captured in scattering media, such as fog or smoke, degrade due to light scattering and attenuation caused by suspended particles. 
For example, Fig. \ref{fig:cover_figure}(a) shows images captured in an actual foggy scene, the contrast of which is reduced due to light scattering. 
Traditional 3D reconstruction methods that exploit observed pixel intensity cannot work in such environments.

We propose a learning-based multi-view stereo (MVS) method in scattering media.
MVS methods \cite{furukawa15} are used for reconstructing the 3D geometry of a scene from multiple images.
Learning-based MVS methods have recently been proposed and provided highly accurate results \cite{yao18,huang18,im19}.
The proposed method is based on MVDepthNet \cite{wang18}, which is one such MVS method.

MVDepthNet estimates scene depth by taking a cost volume as input for the network.
The cost volume is based on a plane sweep volume \cite{collins96}, i.e., it is constructed by sweeping a fronto-parallel plane to a camera in the scene and evaluates the photometric consistency between multiple cameras under the assumptions that the scene lies on each plane.
As described above, however, an image captured in scattering media degrades; thus, using the ordinary cost volume leads to undesirable results, as shown in Fig. \ref{fig:cover_figure}(b).

To solve this problem, we propose a novel cost volume for scattering media, called {\it the dehazing cost volume}.
In scattering media, light bouncing off a scene is attenuated exponentially relative to the depth.
On the other hand, scattered light observed with a camera increases with depth.
This means that degradation due to a scattering medium depends on the scene depth.
Our dehazing cost volume can restore images with such depth-dependent degradation and compute the effective cost of photometric consistency simultaneously.
It enables robust 3D reconstruction in scattering media, as shown in Fig. \ref{fig:cover_figure}(c).

Image degradation in scattering media depends on not only the scene depth but also scattering parameters such as airlight and a scattering coefficient, which determine scattering intensity.
Our dehazing cost volume thus requires these parameters to compute photometric consistency in addition to the depth of the swept plane.
Li et al. \cite{li15} estimated the scattering coefficient under a multi-view setting at a structure-from-motion (SfM) step.
However, this method is not necessarily numerically stable because it directly uses pixel intensity.
We also propose a scattering parameter estimation method with our dehazing cost volume.
Our dehazing cost volume requires the scattering parameters; in other words, the output depth of a network with our dehazing cost volume can be regarded as a function of scattering parameters.
These parameters are thus optimized so that the output depth matches a sparse 3D point cloud obtained by SfM that is less affected by light scattering.
This geometry-based optimization without using pixel intensity is stable and ensures the correctness of the final output depth with the estimated parameters.

The primary contributions of this paper are summarized as follows:
\begin{itemize}
  \item A novel cost volume is introduced to consider photometric consistency and image degradation in scattering media simultaneously. It enables the avoidance of the chicken-and-egg problem of depth estimation and image restoration by computing degradation with the depth of each swept plane in the cost volume.
  \item A method of estimating scattering parameters, such as airlight and a scattering coefficient, is proposed. This method is stable because it uses a 3D point cloud obtained at an SfM step that is less affected by light scattering without using pixel intensity.
  \item We evaluated the effectiveness of our dehazing cost volume against the ordinary cost volume on synthesized hazy images and also demonstrated the applicability to real scenes. We captured a video in actual foggy scenes, which is made available to the public at \url{https://github.com/yfujimura/DCV-release}.
\end{itemize}
This is an extended version of our previous study \cite{fujimura20}.
We additionally provided the details and evaluation of our scattering parameter estimation method and presented new real-world data captured in foggy scenes for further evaluation.
This paper is under consideration at Computer Vision and Image Understanding.

\section{Related work}
\subsection{Multi-view stereo}
As mentioned above, MVS methods \cite{furukawa15} are used for reconstructing 3D geometry using multiple cameras.
In general, it exploits the dense pixel correspondence between multiple images for 3D reconstruction.
The correspondence is referred to as photometric consistency and computed on the basis of the similarity measure of pixel intensity.
One of the difficulties in the computation of photometric consistency is occlusion, i.e., the surface of a target object is occluded from certain cameras.
This leads to incorrect correspondence and inaccurate 3D reconstruction.
To solve this problem, methods have been proposed for simultaneous view selection to compute effective photometric consistency and 3D reconstruction with MVS, achieving highly accurate 3D reconstruction \cite{zheng14,schonberger16}.

Along with the above problem, there are many cases in which it is difficult to obtain accurate 3D geometry with traditional MVS methods.
A textureless surface and an object with a view-dependent reflectance property, such as specular reflection, are typical cases.
Learning-based MVS methods have recently been used to learn semantic information on large-scale training data and enable robust 3D reconstruction in such scenes.

Learning-based MVS methods often construct a cost volume to constrain 3D geometry between multiple cameras.
For example, Wang and Shen \cite{wang18} proposed MVDepthNet, which constructs a cost volume from multi-view images by setting one of the images as a reference image.
It can take an arbitrary number of input images to construct the cost volume.
The convolutional neural network (CNN) takes the reference image and cost volume as input then estimates the depth map of the reference camera.
DeepMVS proposed by Huang et al. \cite{huang18} first constructs a plane sweep volume, then the patch matching network is applied to the reference image and each slice of the volume to extract features to measure the correspondence, which is followed by feature aggregation networks and depth refinement with a fully connected conditional random field.
Yao et al. \cite{yao18} and Im et al. \cite{im19} respectively proposed MVSNet and DPSNet, in which input images are first passed through the networks to extract features, then the features are warped instead of constructing the cost volume in the image space.
Our proposed method is based on MVDepthNet \cite{wang18}, which is the simplest and light-weight method, and we extended the ordinary cost volume and constructs our dehazing cost volume for scattering media.

\subsection{Dehazing}
In scattering media, captured images degraded due to light scattering and attenuation.
To enhance the quality of an image captured in scattering media, dehazing and defogging methods have been proposed \cite{he11,nishino12,fattal14,berman16}.
These studies introduced the priors of latent clear images to solve the ill-posed nature of the problem.
For example, He et al. \cite{he11} proposed a dark channel prior with which a clear image having a dark pixel in a local image patch is assumed.
Berman et al. \cite{berman16} proposed a haze-line prior with which the same intensity pixels of the latent clear image forms a line in RGB space.
Many learning-based methods using neural networks have also recently been proposed \cite{cai16,ren16,zhang18,d_yang18,liu19,qin20,deng20}.
Dehazing can improve computer vision tasks in scattering media such as object detection tasks \cite{li17}.

\subsection{3D reconstruction in scattering media}
Our goal is to reconstruct 3D geometry directly from degraded images by scattering media instead of recovering the latent clear images.
There has been research focusing on the same problem as in our study.
For example, Narasimhan et al. \cite{narasimhan05} proposed a 3D reconstruction method using structured light in scattering media.
Photometric stereo methods have also been proposed for scattering media \cite{tsiotsios14,murez17,fujimura18}.
However, these methods require active light sources, which limits real-world applicability.
Instead of using an ordinary camera, Heide et al. \cite{heide14} and Satat et al. \cite{satat18} respectively used a time-of-flight camera and single photon avalanche diode for scattering media.
Wang et al. \cite{j_wang18} combined a line sensor and line laser to generate a programmable light curtain that can suppress the backscatter effect.
However, the use of these methods is hindered due to the requirement of expensive sensors or special hardware settings.

The proposed method is based on stereo 3D reconstruction requiring neither active light sources nor special hardware settings.
Caraffa et al. \cite{caraffa12} proposed a binocular stereo method in scattering media.
With this method, image enhancement and stereo reconstruction are simultaneously modeled on the basis of a Markov random field. 
Song et al. \cite{song18} proposed a learning-based binocular stereo method in scattering media, where dehazing and stereo reconstruction are trained as multi-task learning.
The features from the networks of each task are simply concatenated at the intermediate layer.
The most related method to ours is the MVS method proposed by Li et al. \cite{li15}. 
They modeled dehazing and MVS simultaneously and regularized the output depth using an ordering constraint, which was based on a transmission map that was the output of dehazing with Laplacian smoothing.
With all these methods, homogeneous scattering media is assumed; thus, we followed the same assumption. It is left open to apply these methods to inhomogeneous media.

These previous studies \cite{caraffa12,li15} designed photometric consistency measures considering the scattering effect.
However, this requires scene depth because degradation due to scattering media depends on this depth.
Thus, they relied on iterative implementation of an MVS method and dehazing, which leads to large computation cost.
In contrast, our dehazing cost volume can solve this chicken-and-egg problem by computing the scattering effect in the cost volume. 
The scene depth is then estimated effectively by taking the cost volume as input for a CNN, making fast inference possible.

\section{Multi-view stereo in scattering media} \label{sec:mvs_in_scattering_media}
In this section, we describe MVS in scattering media with our dehazing cost volume.
First, we introduce an image formation model in scattering media, followed by an overview of the proposed method, finally a discussion on the ordinary cost volume and our dehazing cost volume.

\subsection{Image formation model} \label{sec:image_formation_model}
We use an atmospheric scattering model \cite{tan08} for image observation in scattering media.
This model is used for many dehazing methods and describes the degradation of an observed image in scattering media in daylight.
Let an RGB value at the pixel $(u,v)$ of a degraded image captured in scattering media and its latent clear image be $I(u,v) \in \mathbb{R}^3$ and $J(u,v) \in \mathbb{R}^3$, respectively.
We assume that the pixel value of each color channel is within $0$ and $1$.
The observation process of this model is given as
\begin{equation}
I(u,v) = J(u,v) e^{-\beta z(u,v)} + \mathbf{A}(1 - e^{-\beta z(u,v)}),
\label{eq:image_formation_model}
\end{equation}
where $z(u,v) \in \mathbb{R}$ is the depth at pixel $(u,v)$, $\beta \in \mathbb{R}$ is a scattering coefficient that represents the density of a medium, and $\mathbf{A} \in \mathbb{R}^3$ is global airlight.
For simplicity, we assume that $\mathbf{A}$ is given by $\mathbf{A} = [A, A, A]^\top, A \in \mathbb{R}$, i.e., the color of scattering media is achromatic (gray or white).
The first term is a component describing reflected light in a scene. 
This reflected component becomes attenuated exponentially with respect to the scene depth.
The second term is a scattering component, which consists of scattered light that arrives at a camera without reflecting on objects.
In contrast to the reflected component, this component increases with depth.
Therefore, image degradation due to scattering media depends on the scene depth.

In the context of image restoration, we aim to estimate unknown parameters $J$, $z$, scattering coefficient $\beta$, and airlight $A$ from an observed image $I$, and the estimation of all these parameters at the same time is an ill-posed problem.
Previous studies developed methods for estimating $A$ from a single image \cite{he11,berman17}, and Li et al. \cite{li15} estimated $\beta$ under a multi-view setting at an SfM step.
Thus, it is assumed with their MVS method in scattering media that $A$ and $\beta$ can be estimated beforehand. (Our dehazing cost volume also requires these parameters.)
However, such an assumption is sometimes too strict especially for $\beta$, and the estimation error of these parameters affects the following 3D reconstruction.
With our dehazing cost volume, the estimation of these parameters is achieved in the same framework as the following depth estimation; thus, the correctness of the estimated depth is ensured.

\subsection{Overview}
MVS methods are roughly categorized by output representations, e.g., point-cloud, volume, or mesh-based reconstruction.
The proposed method is formulated as depth-map estimation, i.e., given multiple cameras, we estimate a depth map for one of the cameras.
We refer to a target camera to estimate a depth map as a reference camera $r$ and the other cameras as source cameras $s \in \{1, \cdots, S \}$, and images captured with these cameras are denoted as a reference image $I_r$ and source images $I_s$, respectively. 
We assume that the camera parameters are calibrated beforehand. 

\begin{figure}[tb]
\centering
\includegraphics[width=0.48\textwidth]{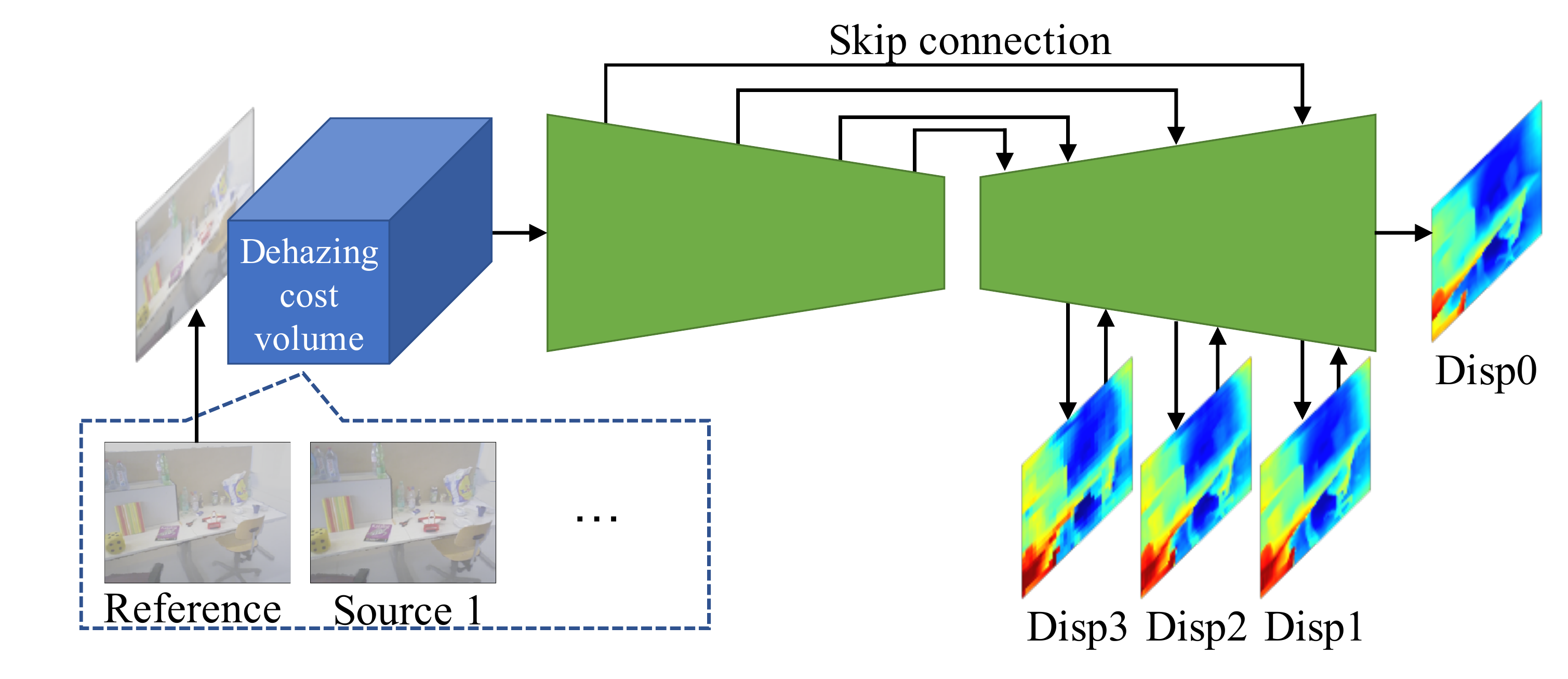}
\caption{Input of network is reference image captured in scattering medium and our dehazing cost volume. Our dehazing cost volume is constructed from reference image and source images. Network architecture of our method is same as that of MVDepthNet \cite{wang18}, which has encoder-decoder with skip connections. Output of network is disparity maps (inverse depth maps) at different resolutions.}
\label{fig:architecture}
\end{figure}

An overview of the proposed method is shown in Fig. \ref{fig:architecture}.
Our dehazing cost volume is constructed from a hazy reference image and source images captured in a scattering medium.
The network takes the reference image and our dehazing cost volume as input then outputs a disparity map (inverse depth map) of the reference image.
The network architecture is the same as that of MVDepthNet \cite{wang18}, while the ordinary cost volume used in MVDepthNet is replaced with our dehazing cost volume for scattering media.

\subsection{Dehazing cost volume}
In this section, we explain our dehazing cost volume, which is taken as input to the network.
The dehazing cost volume enables effective computation of photometric consistency in scattering media.

\begin{figure*}[tb]
\begin{minipage}[b]{0.45\hsize}
\centering
\subfloat[Cost volume]{\includegraphics[scale=0.3]{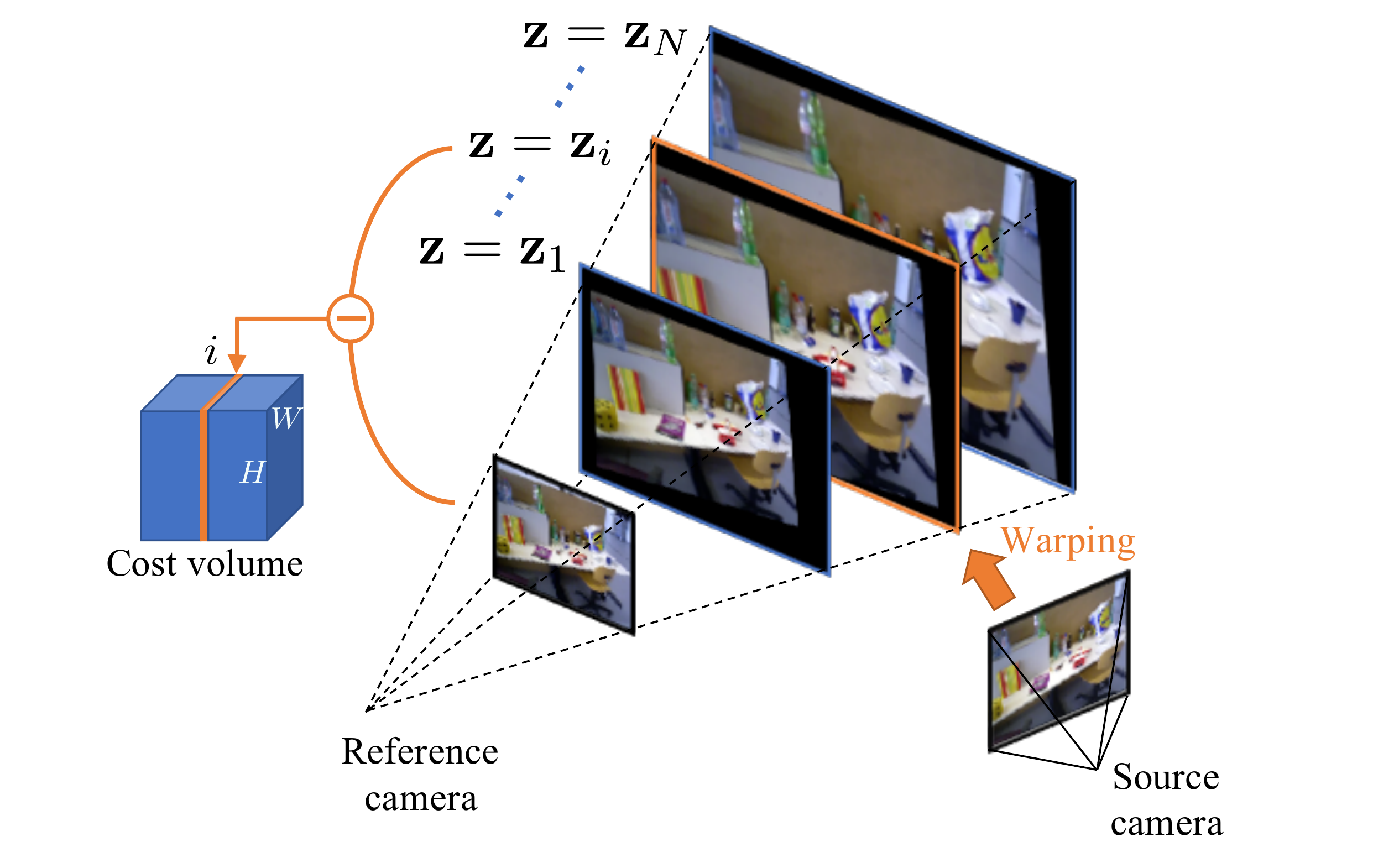}} 
\end{minipage}
\begin{minipage}[b]{0.5\hsize}
\centering
\subfloat[Dehazing cost volume]{\includegraphics[scale=0.3]{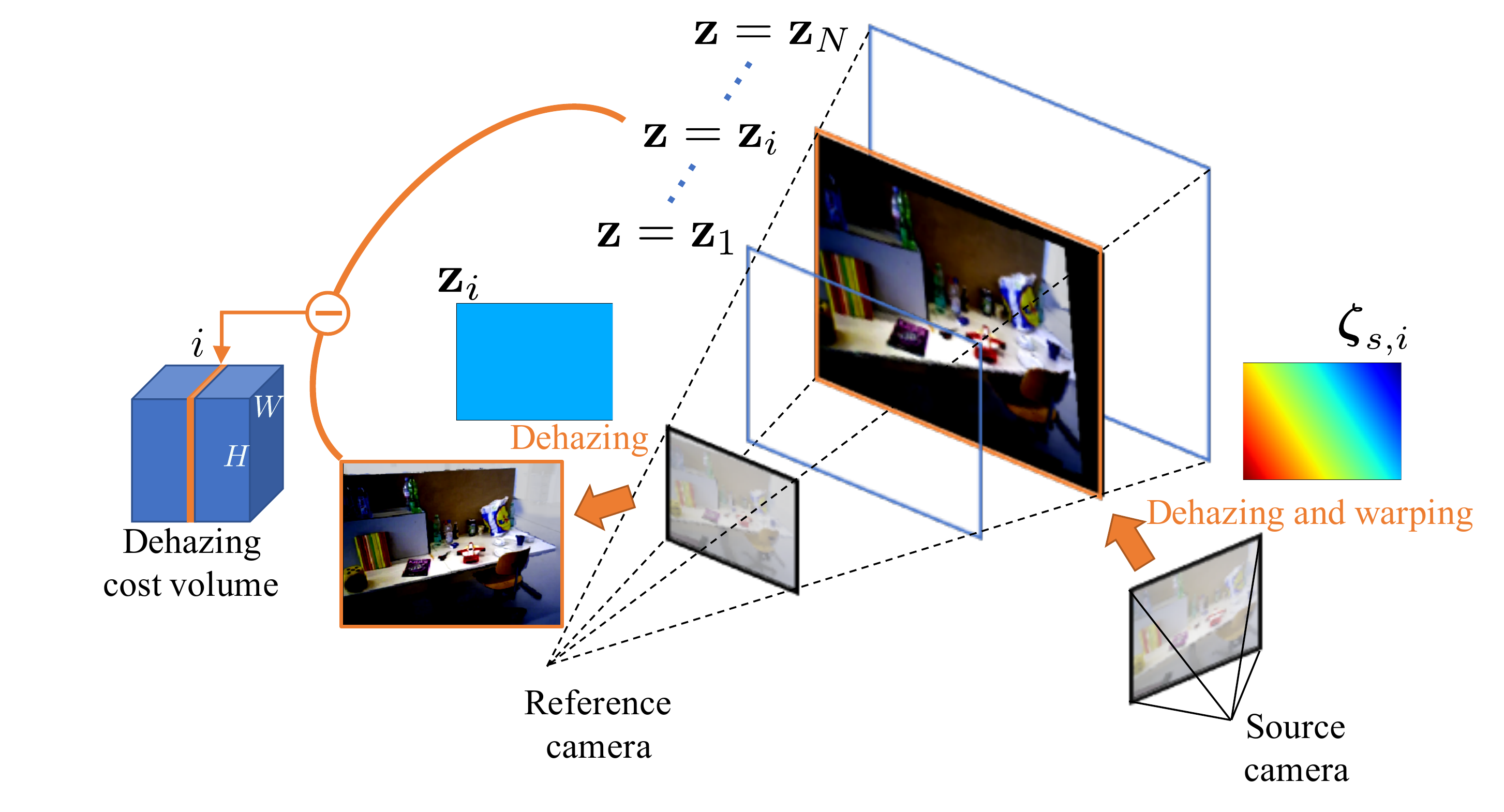}}
\end{minipage}
\caption{(a) Ordinary cost volume is constructed by sweeping fronto-parallel plane in reference-camera coordinate. Cost of photometric consistency is simply computed as residual between reference image and warped source image on each swept plane $\mathbf{z}=\mathbf{z}_i$. (b) In our dehazing cost volume, reference image is dehazed using sampled depth, $\mathbf{z}_i$, which is constant over all pixels. Source image is dehazed using depth of swept plane from source-camera view, then dehazed source image is back-projected onto plane. Cost is computed by taking residual between both dehazed images.}
\label{fig:cost_volume}
\end{figure*}

Before explaining our dehazing cost volume, we show the computation of the ordinary cost volume in Fig. \ref{fig:cost_volume}(a).
We first sample the 3D space in the reference-camera coordinate system by sweeping a fronto-parallel plane.
We then back-project source images onto each sampled plane.
Finally, we take the residual between the reference image and each warped source image, which corresponds to the cost of photometric consistency on the hypothesis that the scene exists on the plane.
Let the image size be $W \times H$ and number of sampled depths be $N$.
We denote the cost volume as $\mathcal{V}: \{1,\cdots,W\} \times \{1,\cdots,H\} \times \{1,\cdots,N\} \to \mathbb{R}$, and each element of the cost volume is given as follows:
\begin{equation}
\mathcal{V}(u,v,i) = \frac{1}{S} \sum_{s} \| I_r(u,v) - I_s(\pi_{r \to s}(u,v; z_i)) \|_1,
\label{eq:cost_volume}
\end{equation}
where $z_i$ is the depth value of the $i$-th plane.
The operator $\pi_{r \to s}:\mathbb{R}^2\to \mathbb{R}^2$ projects the camera pixel $(u,v)$ of the reference camera $r$ onto the source image $I_s$ with the given depth, which is defined as follows:
\begin{equation}
\left[
\begin{array}{c}
\pi_{r \to s}(u,v; z) \\
1
\end{array}
\right] \sim z \mathbf{K}_s \mathbf{R}_{r\to s} \mathbf{K}_r^{-1}
\left[
\begin{array}{c}
u \\
v \\
1
\end{array}
\right]
+ \mathbf{K}_s \mathbf{t}_{r \to s},
\end{equation}
where $\mathbf{K}_r$ and $\mathbf{K}_s$ are the intrinsic parameters of the reference camera $r$ and the source camera $s$, and $\mathbf{R}_{r\to s}$ and $\mathbf{t}_{r \to s}$ are a rotation matrix and translation vector from $r$ to $s$, respectively.
The cost volume evaluates the photometric consistency of each pixel with respect to the sampled depth; thus, the element of the cost volume with correct depth ideally becomes zero. 

\begin{figure*}[tb]
\centering
\subfloat[]{\includegraphics[width=0.18\textwidth]{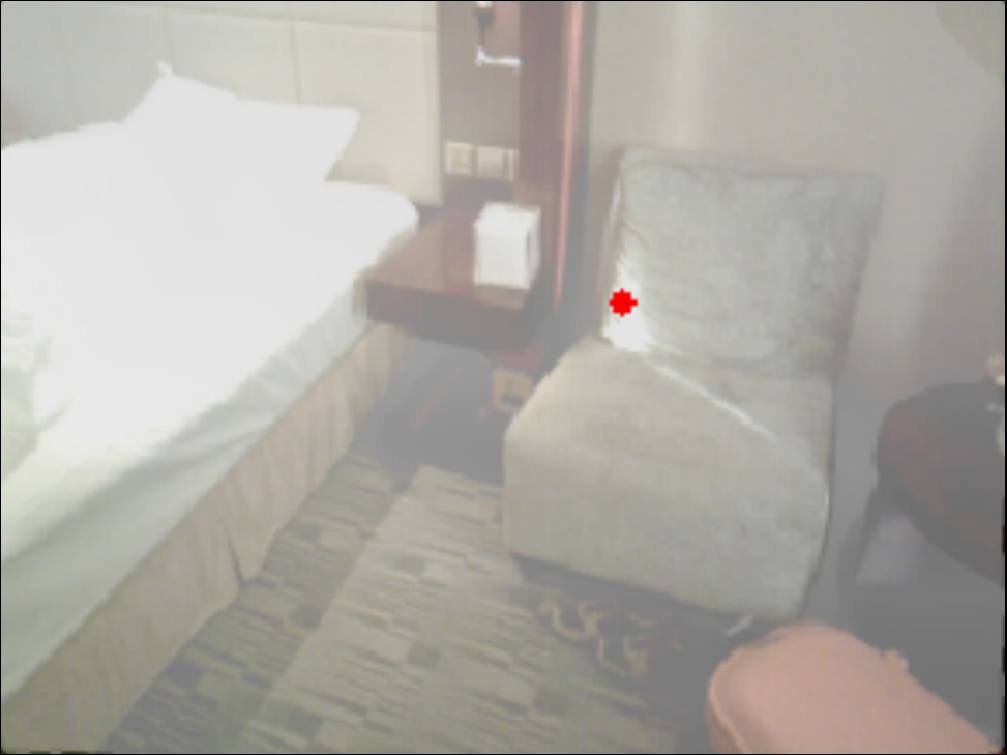}} \quad
\subfloat[]{\includegraphics[width=0.2\textwidth]{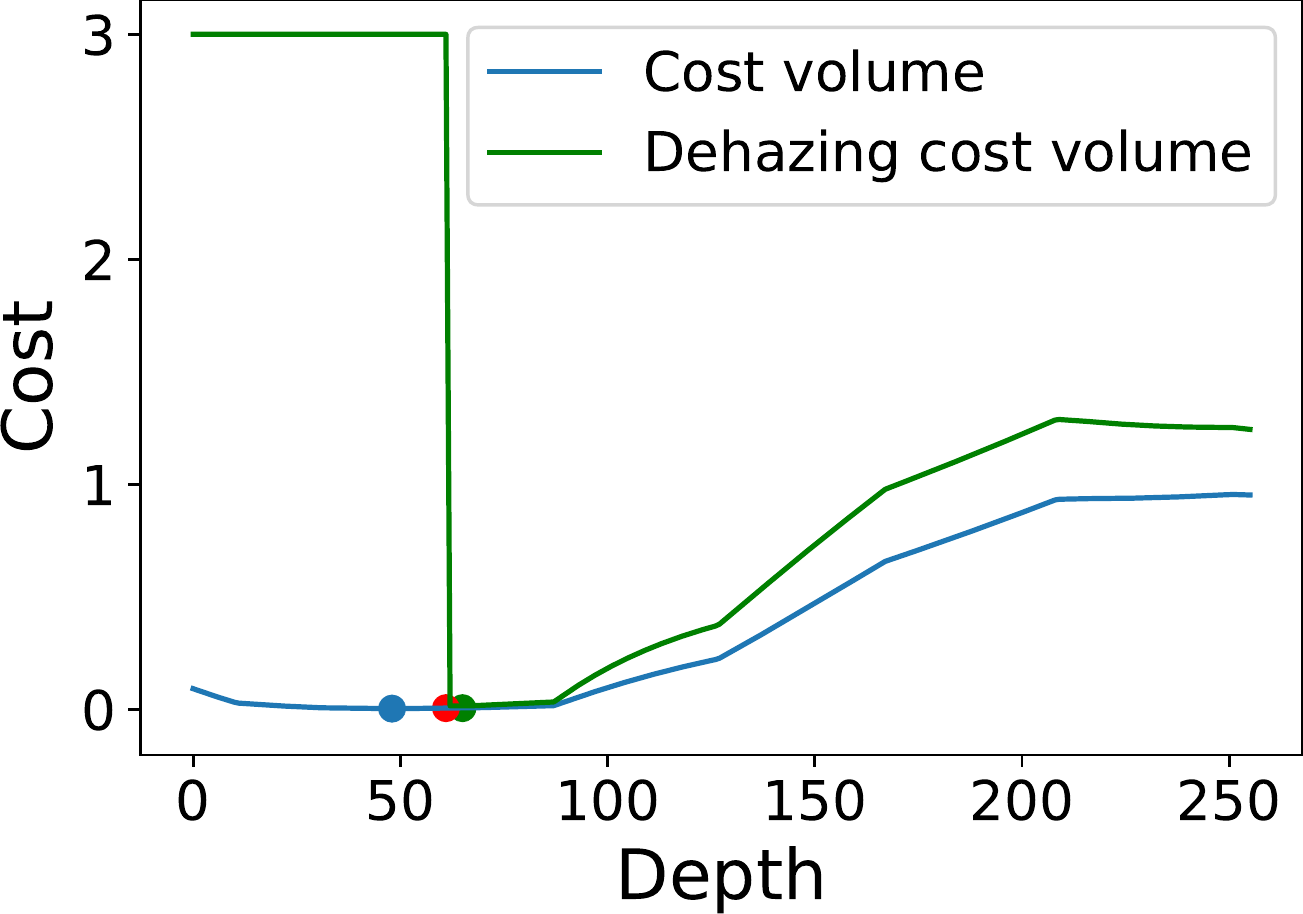}}\quad
\subfloat[]{\includegraphics[width=0.18\textwidth]{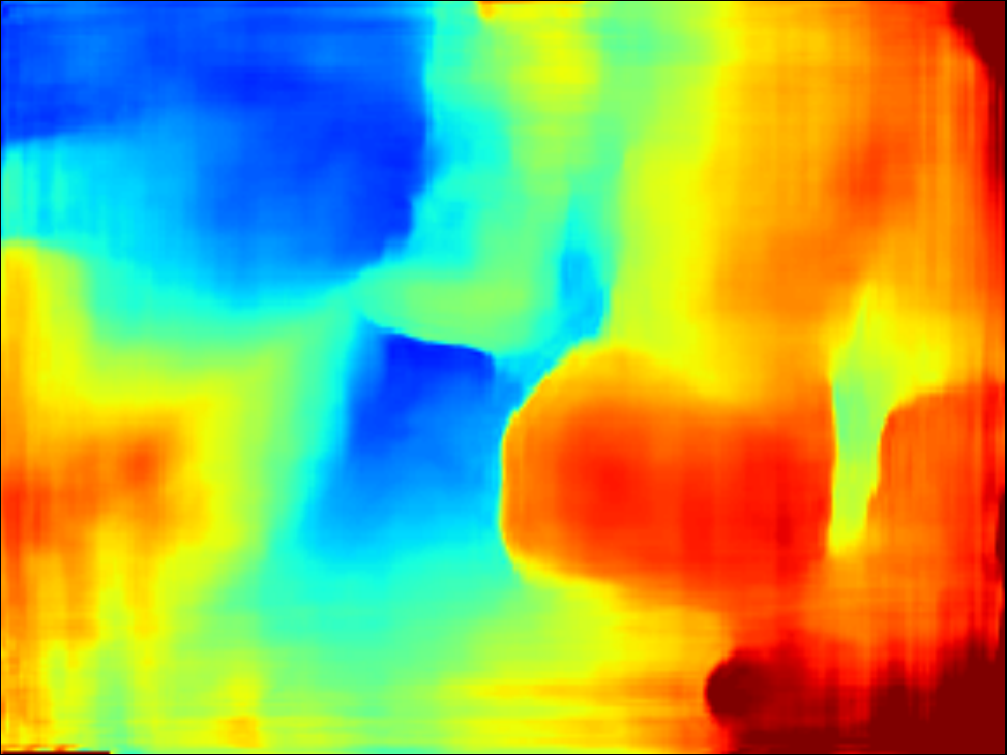}} \quad
\subfloat[]{\includegraphics[width=0.18\textwidth]{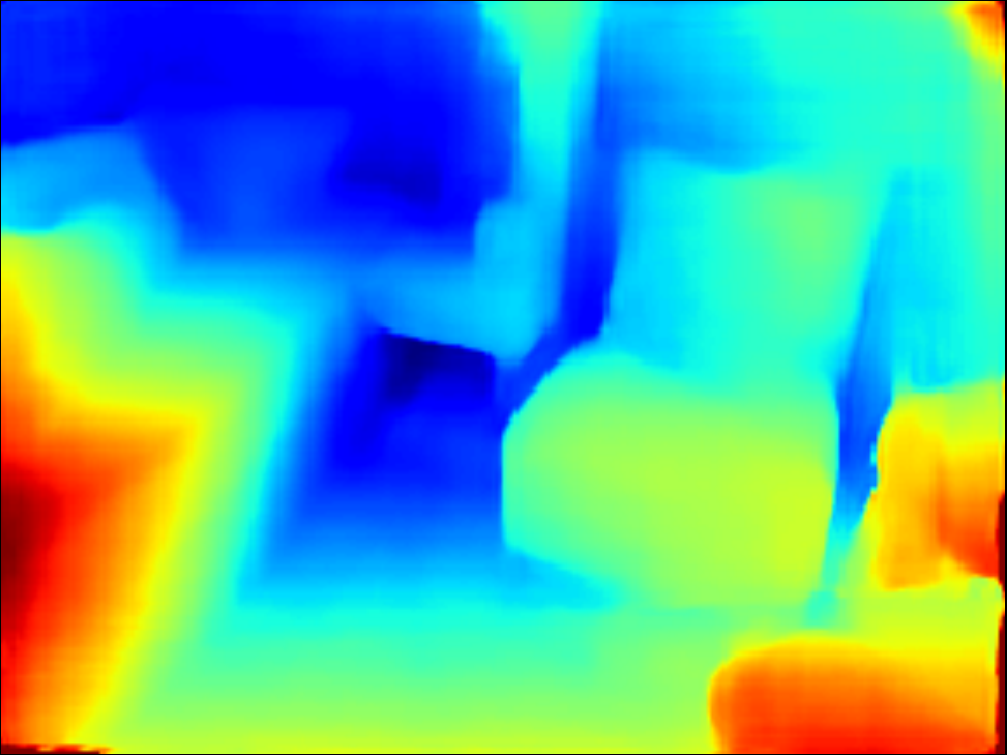}}
\caption{Visualization of our dehazing cost volume. (b) Computed ordinary cost volume and our dehazing cost volume at red point in (a). In (b), red dot indicates location of ground-truth, and blue and green dots indicate minimum value of ordinary cost volume and our dehazing cost volume, respectively. (c) and (d) Output depth of MVDepthNet (\cite{wang18}) with ordinary cost volume and our dehazing cost volume, respectively.}
\label{fig:visualize_dcv}
\end{figure*}

An observed image captured in scattering media degrades in the manner described in Eq. (\ref{eq:image_formation_model}), and the ordinary cost volume defined in Eq. (\ref{eq:cost_volume}) leads to undesirable results.
In contrast, our dehazing cost volume dehazes the image and computes photometric consistency cost simultaneously.
As described in Section \ref{sec:image_formation_model}, degradation due to scattering media depends on scene depth; thus, our dehazing cost volume restores degraded images using the depth of a swept plane.

Figure \ref{fig:cost_volume}(b) shows the computation of our dehazing cost volume.
A reference image is dehazed directly using the depth of a swept plane.
A source image is dehazed using the swept plane from a source camera view, then the dehazed source image is warped to the reference-camera coordinate system.
Similar to the ordinary cost volume, we define our dehazing cost volume as $\mathcal{D}:\{ 1, \cdots, W\} \times \{ 1, \cdots, H \} \times \{ 1, \cdots, N \} \to \mathbb{R}$, and each element of our dehazing cost volume is given as
\begin{flalign}
&\mathcal{D}(u,v,i) = \frac{1}{S} \sum_{s} \| J_r(u,v; z_i) - J_s(\pi_{r \to s}(u,v; z_i)) \|_1,
\label{eq:dehazing_cost_volume_}
\end{flalign}
where $J_r(u,v;z_i)$ and $J_s(\pi_{t \to s}(u,v;z_i))$ are dehazed reference and source images.
From Eq. (\ref{eq:image_formation_model}), if $A$ and $\beta$ are estimated beforehand, they are computed as follows:
\begin{flalign}
J_r(u,v;z_i) &= \frac{I_r(u,v) - \mathbf{A}}{e^{-\beta z_i}} + \mathbf{A},&\\
J_s(\pi_{r \to s}(u,v;z_i)) &= \frac{I_s( \pi_{r \to s}(u,v;z_i) ) - \mathbf{A}}{e^{-\beta \zeta_{s,i} (\pi_{r \to s}(u,v;z_i))}} + \mathbf{A}. &
\end{flalign}
As shown in Fig. \ref{fig:cost_volume}(b), the reference image is dehazed using the swept plane with depth $z_i$, whose depth map is denoted as $\mathbf{z}_i$.
On the other hand, the source image is dehazed using ${\boldsymbol \zeta}_{s,i}$, 
which is a depth map of the swept plane from the source camera view.
The depth $\zeta_{s,i} (\pi_{r \to s}(u,v;z_i))$ is used for the cost computation of the pixel $(u,v)$ of the reference camera because the pixel $\pi_{r \to s}(u,v;z_i)$ on the source camera corresponds to pixel $(u,v)$ of the reference camera.
Our dehazing cost volume exploits the dehazed images with much more contrast than the degraded ones; thus, the computed cost is robust even in scattering media.
In accordance with this definition of our dehazing cost volume, the photometric consistency between the latent clear images is preserved.

Our dehazing cost volume computes photometric consistency with dehazed images in the cost volume. This is similar to the previous methods \cite{caraffa12,li15} that compute photometric consistency considering scattering effect. However, this is a chicken-and-egg problem because the effect of scattering media depends on scene depth, and they rely on iterative implementation of MVS and dehazing to compute the scattering effect. Our method, on the other hand, can compute the scattering effect using a depth hypothesis of a swept plane without an explicit scene depth, which can eliminate the iterative optimization.

Our dehazing cost volume restores an image using all depth hypotheses; thus, image dehazing with depth that greatly differs from the correct scene depth results in an unexpected image.
The extreme case is when a dehazed image has negative values at certain pixels.
This includes the possibility that a computed cost using Eq. (\ref{eq:dehazing_cost_volume_}) becomes very large.
To avoid such cases, we revise the definition of our dehazing cost volume as follows:
\begin{flalign}
&\mathcal{D}(u,v,i) = & \nonumber \\ 
&\frac{1}{S} \sum_{s}\left\{
\begin{array}{ll}
 \multicolumn{2}{l}{\| J_r(u,v; z_i) - J_s(\pi_{r \to s}(u,v; z_i)) \|_1}  \\
 & if \; 0 \leq J_r^c(u,v; z_i) \leq 1 \;and \\
 & 0 \leq J_s^c(\pi_{r \to s}(u,v; z_i)) \leq 1  \;\;\;\; c \in \{ r,g,b\}\\ 
 \gamma & otherwise,
\end{array}
\right. &
\label{eq:dehazing_cost_volume}
\end{flalign}
where $J_r^c(u,v;z_i)$ and $J_s^c(\pi_{r \to s}(u,v;z_i))$ are the pixel values of the channel $c \in \{r,g,b\}$ of the reconstructed clear images.
A constant $\gamma$ is a parameter that is set as a penalty cost when the dehazed result is not contained in the domain of definition.
This makes the training of the network stable because our dehazing cost volume is upper bounded by $\gamma$. 
We can also reduce the search space of depth by explicitly giving the penalty cost.
In this study, we set $\gamma=3$, which is the maximum value of the ordinary cost volume defined in Eq. (\ref{eq:cost_volume}) when the pixel value of each color channel is within $0$ and $1$. 

Figure \ref{fig:visualize_dcv}(b) visualizes the ordinary cost volume and our dehazing cost volume at the red point in (a). 
Each dot in (b) indicates a minimum cost, and the red dot in (b) indicates ground-truth depth. 
The curve of the cost volume is smoother than that of our dehazing cost volume due to the degradation in image contrast, which leads to a depth error. 
Our dehazing cost volume can also reduce the search space with the dehazing constraint $\gamma$ on the left part in (b), where its cost value is constantly large.

\subsection{Network architecture and loss function}
As shown in Fig. \ref{fig:architecture}, a network takes a reference image and our dehazing cost volume as input.
To compute our dehazing cost volume, we should predetermine the target 3D space for scene reconstruction and number of depth hypotheses for plane sweep.
We uniformly sample the depth on the disparity space between $0.02$ and $2$ and set the number of samples to $N=256$.
The network architecture is the same as that of MVDepthNet \cite{wang18}, which has an encoder-decoder architecture with skip connections.
The network outputs disparity maps at different resolutions.
The training loss is defined as the sum of L1 loss between these estimated disparity maps and the ground-truth disparity map. (For more details, please refer to \cite{wang18}.)

\section{Scattering parameter estimation}
As mentioned in Section \ref{sec:mvs_in_scattering_media}, our dehazing cost volume requires scattering parameters, airlight $A$ and a scattering coefficient $\beta$.
In this section, we first explain the estimation of $A$ then describe the difficulty of estimating $\beta$
Finally, we discuss the simultaneous estimation of the scattering parameters and depth with our dehazing cost volume.

\subsection{Estimation of airlight $A$}\label{sec:estimation_of_airlight}
\begin{figure}[tb]
\centering
\includegraphics[width=0.35\textwidth]{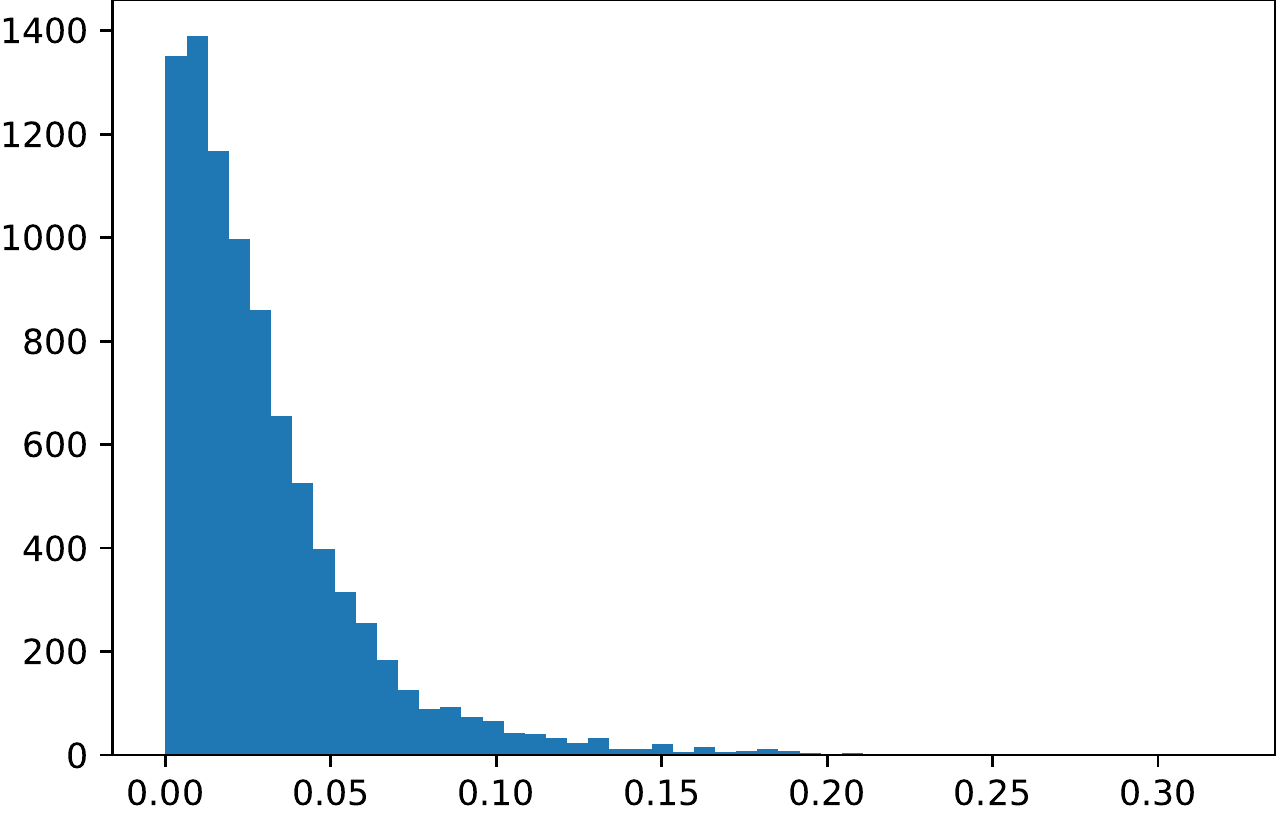}
\caption{Error histogram of our airlight estimator on synthesized test dataset. Simple L1 error is computed on each estimate. In this dataset, $A$ is randomly sampled from $[0.7, 1.0]$.}
\label{fig:error_A}
\end{figure}

We first describe the estimation of $A$.
Although methods for estimating $A$ from a single image have been proposed, we implement and evaluate a CNN-based estimator, the architecture of which is shown in Table \ref{tab:airlight_estimator}.
It takes a single RGB image as input, which is passed through several convolution layers with stride 2.
Global average pooling is then applied to generate a $256 \times 1 \times 1$ feature map.
This feature map is passed through two $1 \times 1$ convolutions to yield 1D output $A$.
Note that each convolution layer except for the final layer (conv8) is followed by batch normalization and then by rectified linear unit (ReLU) activation.
For training and test, we used the synthesized image dataset described in Section \ref{sec:dataset}.
Figure \ref{fig:error_A} shows the error histogram of $A$ on the test dataset.
In this dataset, the value of $A$ is randomly sampled from $[0.7, 1.0]$, indicating that the estimation of $A$ can be achieved from a single image.

\begin{table}[tb]
\centering
\caption{Network architecture of airlight estimator. Network takes single RGB image as input then outputs single scalar value $A$. Stride of convolution layers from conv1 to conv6 is 2. Each convolution layer except for conv8 has batch normalization and ReLU activation. glb\_avg\_pool denotes global average pooling layer.}
\label{tab:airlight_estimator}
\begin{tabular}{cccc}\hline
Layer & Kernel & Channel & Input \\ \hline
conv1 & 7 & 3/16 & $I$ \\
conv2 & 5 & 16/32 & conv1  \\
conv3 & 3 & 32/64 & conv2 \\
conv4 & 3 & 64/128 & conv3 \\
conv5 & 3 & 128/256 & conv4 \\
conv6 & 3 & 256/256 & conv5 \\
glb\_avg\_pool & - & 256/256 & conv6 \\
conv7 & 1 & 256/64 & glb\_avg\_pool \\
conv8 & 1 & 64/1 & conv7 \\ \hline
\end{tabular}
\end{table}

\subsection{Difficulty of estimating scattering coefficient $\beta$}
In contrast to $A$, it is difficult to estimate $\beta$ from a single image.
As shown in Eq. (\ref{eq:image_formation_model}), image degradation due to scattering media depends on $\beta$ and scene depth $z$ through $e^{-\beta z}$ with the scale-invariant property, i.e., the pairs of $k \beta$ and $(1/k)z$ for arbitrary $k \in \mathbb{R}$ lead to the same degradation.
Since the depth scale cannot be determined from a single image, the estimation of the scattering coefficient from a single image is infeasible.

In response to this problem, Li et al. \cite{li15} proposed a method for estimating $\beta$ from multi-view images.
With this method, it is assumed that a sparse 3D point cloud and camera parameters can be obtained by SfM from noticeable image edges even in scattering media.
From a pixel pair and corresponding 3D point, two equations can be obtained from Eq. (\ref{eq:image_formation_model}).
Additionally, if we assume that the pixel value of the latent clear image is equal to the corresponding pixel value of the other clear image, this simultaneous equations can be solved for $\beta$.
However, this multi-view-based method involves several strong assumptions.
First, the pixel value of the latent clear image should be completely equal to the corresponding pixel value of the other clear image.
Second, the values of the observed pixels should be sufficiently different to ensure numerical stability.
This assumption means the depth values of both images should be sufficiently different, and it is sometimes very difficult to find such points.
Finally, $A$ is assumed to be properly estimated beforehand.
These limitations indicate that we should avoid using the pixel values directly for $\beta$ estimation.

\subsection{Estimation with geometric information}
In this study, the scattering coefficient was estimated without using pixel intensity.
Our method ensures the correctness of the output depth with the estimated scattering coefficient.

As well as the MVS method proposed by Li et al. \cite{li15}, a sparse 3D point cloud is assumed to be obtained by SfM in advance.
Although our dehazing cost volume, which is taken as input for a network, requires $A$ and $\beta$, this means that the network can be regarded as a function that takes $A$ and $\beta$ as variables and outputs a depth map.
Now, the network with fixed parameters is denoted by $\mathcal{F}$, and the output depth can be written by $\mathbf{z}_{A,\beta} = \mathcal{F}(A, \beta)$ as a function of $A$ and $\beta$.
Note that for simplicity, we omitted the input image from the notation.
Let a depth map that corresponds to a sparse 3D point cloud by SfM be $\mathbf{z}_{sfm}$.
The scattering parameters are estimated by solving the following optimization problem:
\begin{equation}\label{eq:objective}
A^*, \beta^* = \argmin_{A,\beta} \sum_{u,v} m(u,v) \rho\Bigl(z_{sfm}(u,v), z_{A,\beta}(u,v)\Bigr),
\end{equation}
where $z_*(u,v)$ denotes a value at the pixel $(u,v)$ of a depth map $\mathbf{z}_*$, and $m(u,v)$ is an indicator function, where $m(u,v)=1$ if a 3D point estimated by SfM is observed at pixel $(u,v)$, and $m(u,v) = 0$ otherwise.
A function $\rho$ computes the residual between the argument 
depths.
Therefore, the solution of Eq. (\ref{eq:objective}) minimizes the difference between the output depth of the network and the sparse depth map obtained by SfM.
A final dense depth map can then be computed with the estimated $A^*$ and $\beta^*$, i.e., $\mathbf{z}^* = \mathcal{F}(A^*, \beta^*)$.
Differing from the previous method \cite{li15}, our method does not require pixel intensity because the optimization is based on only geometric information, and the final output depth is ensured to match at least the sparse depth map obtained by SfM.

\begin{figure}[tb]
\centering
\subfloat[]{\includegraphics[width=0.18\textwidth]{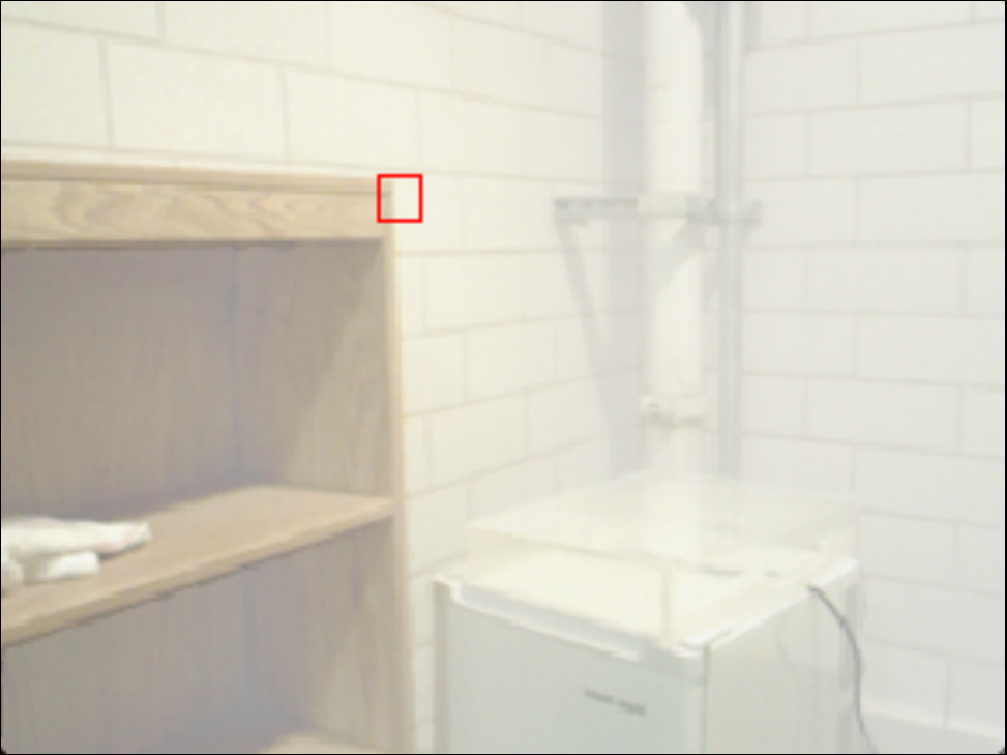}} \;
\subfloat[]{\includegraphics[width=0.18\textwidth]{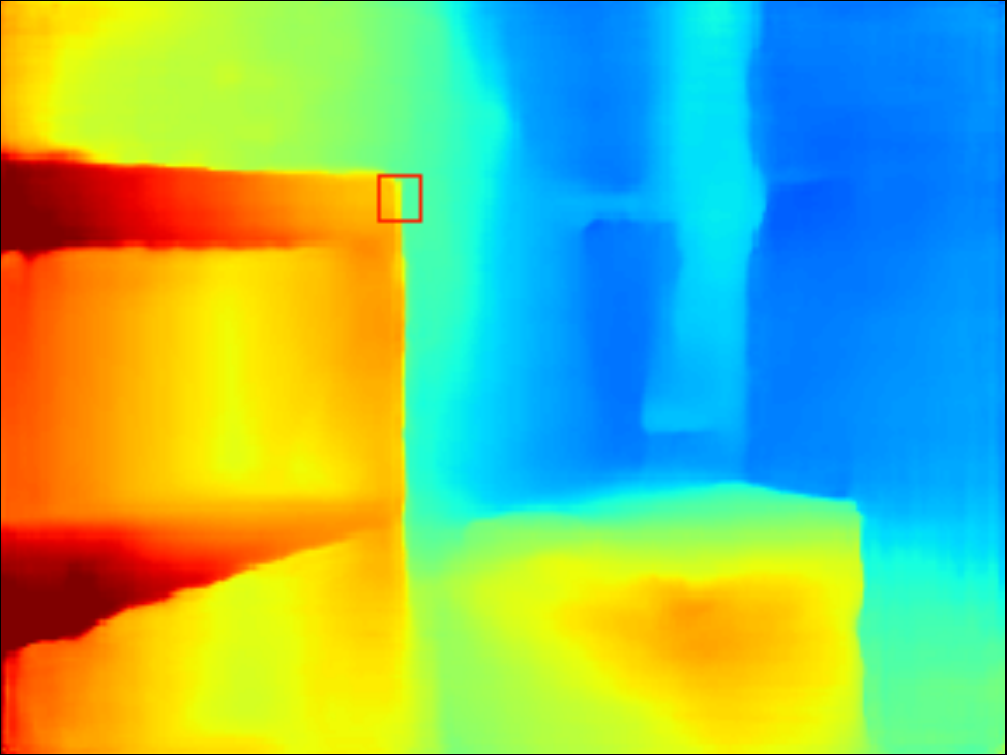}}\\
\subfloat[]{\includegraphics[width=0.1\textwidth]{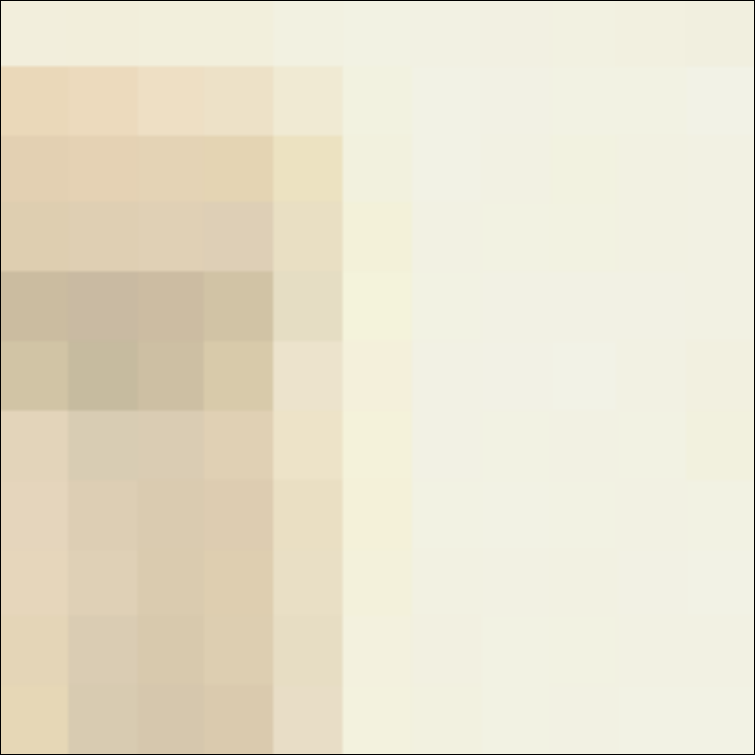}} \;
\subfloat[]{\includegraphics[width=0.1\textwidth]{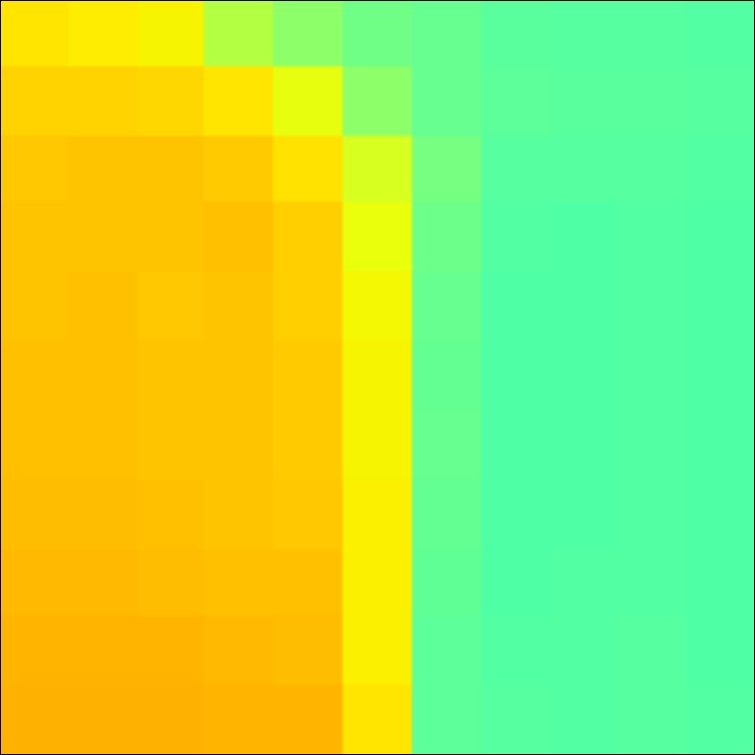}}\;
\subfloat[]{\includegraphics[width=0.1\textwidth]{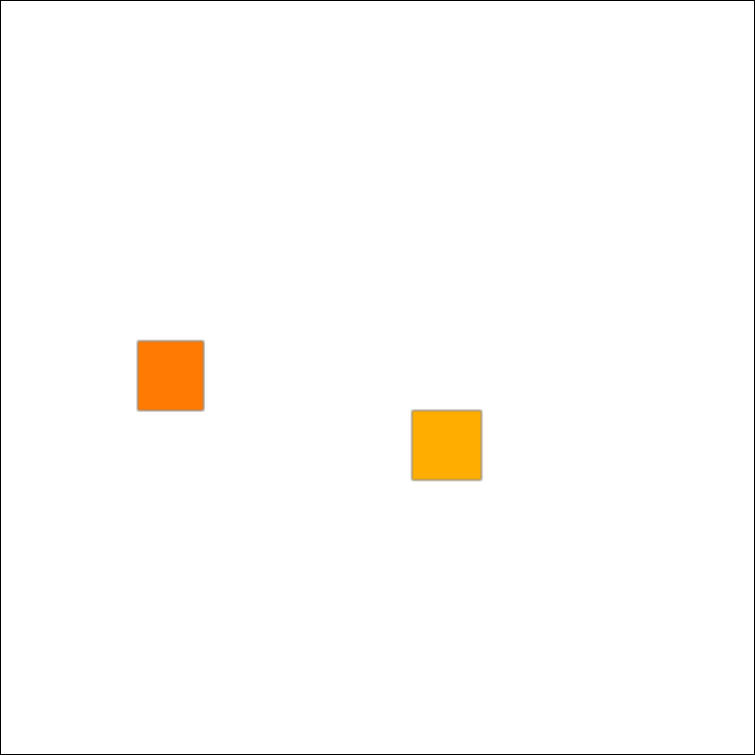}} 
\caption{(a) Input image. (b) Output depth with ground-truth scattering parameters. Depth discontinuities exist in red boxed region. Zoom of regions in (a) and (b) are shown in (c) and (d), respectively. (e) Depth map of sparse 3D point cloud obtained by SfM in this region. It is uncertain whether feature point obtained by SfM is located on background or foreground around depth discontinuities. This includes possibility that output depths of network and SfM are completely different such as right pixel in (e).}
\label{fig:depth_edge}
\end{figure}

We use the following function as $\rho$ to measure the difference between depth values:
\begin{flalign}
&\rho\Bigl(z_{sfm}(u,v), z_{A,\beta}(u,v)\Bigr) = & \nonumber \\
&\min \left \{
\begin{array}{l}
| z_{sfm}(u,v) - z_{A,\beta}(u,v)|, \\
| z_{sfm}(u,v) - z_{A,\beta}(u+\delta,v)|, \\
| z_{sfm}(u,v) - z_{A,\beta}(u-\delta,v)|, \\
| z_{sfm}(u,v) - z_{A,\beta}(u,v+\delta)|, \\
| z_{sfm}(u,v) - z_{A,\beta}(u,v-\delta)| 
\end{array}\right\} \label{eq:definition_of_rho}
\end{flalign}
As shown in Fig. \ref{fig:depth_edge}, it is uncertain whether the feature point obtained by SfM is located on the background or foreground around depth discontinuities.
This includes the possibility that the output depths of the network and SfM are completely different.
To suppress the effect of this error on the scattering parameter estimation, we use the neighboring pixels when calculating the residual of the depths.
As shown in Eq. (\ref{eq:definition_of_rho}), we use the depth values of the pixels at a distance of $\delta$ pixel in the horizontal and vertical direction.
The minimum value among these residuals is used for the optimization.
Note that we set $\delta=5$ pixels in this study.

\subsection{Solver}
\begin{figure}[tb]
\centering
\subfloat[]{\includegraphics[width=0.15\textwidth]{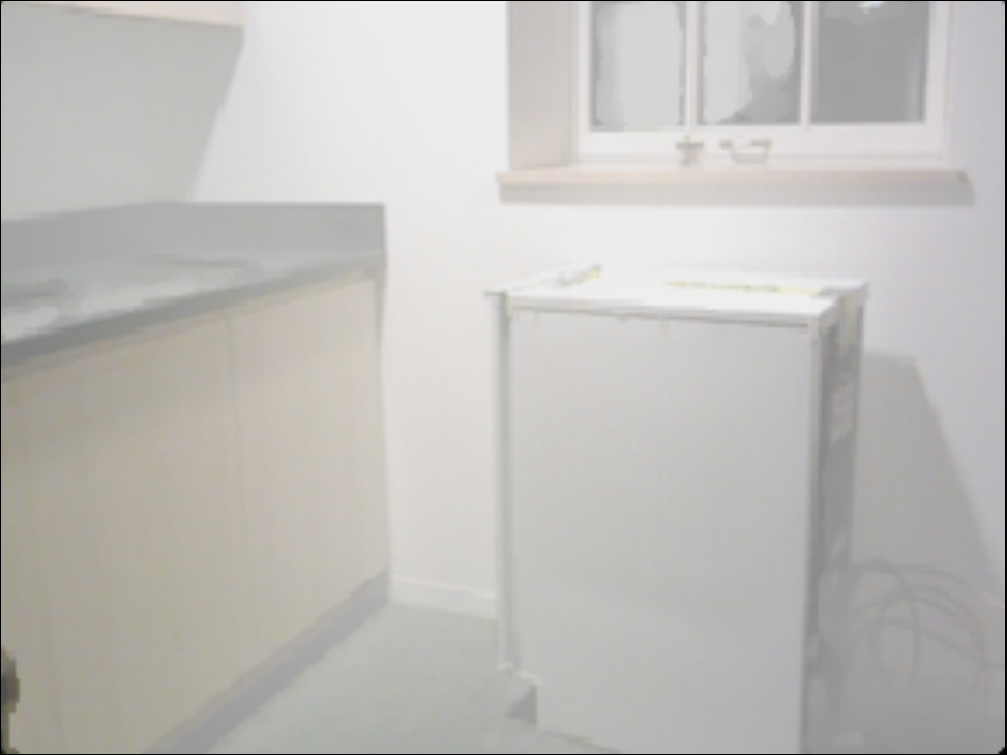}} \;
\subfloat[]{\includegraphics[width=0.15\textwidth]{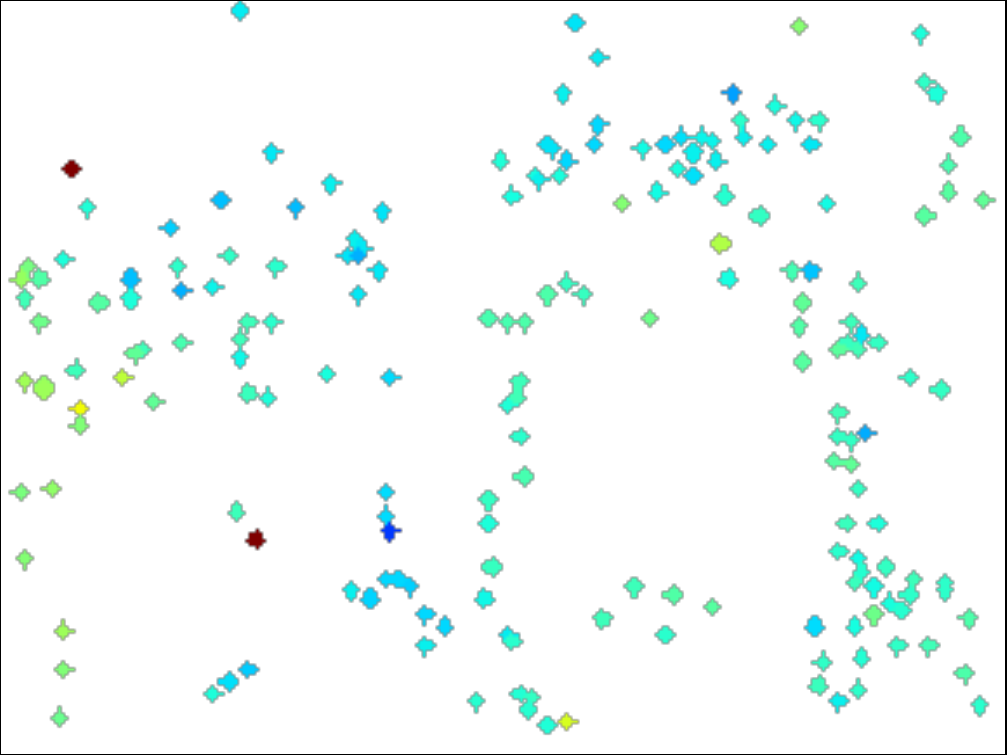}} \\
\subfloat[]{\includegraphics[width=0.15\textwidth]{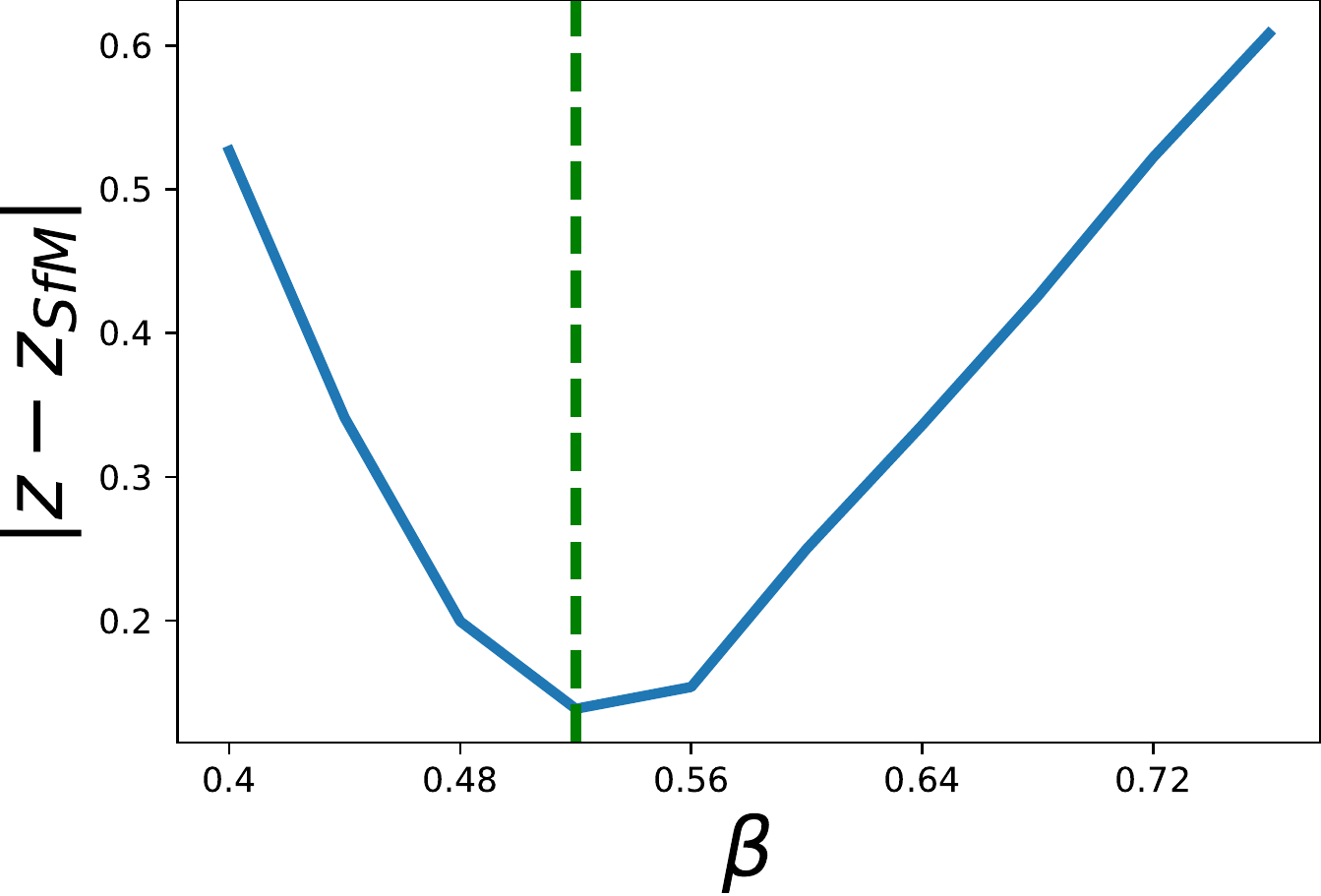}} \;
\subfloat[]{\includegraphics[width=0.15\textwidth]{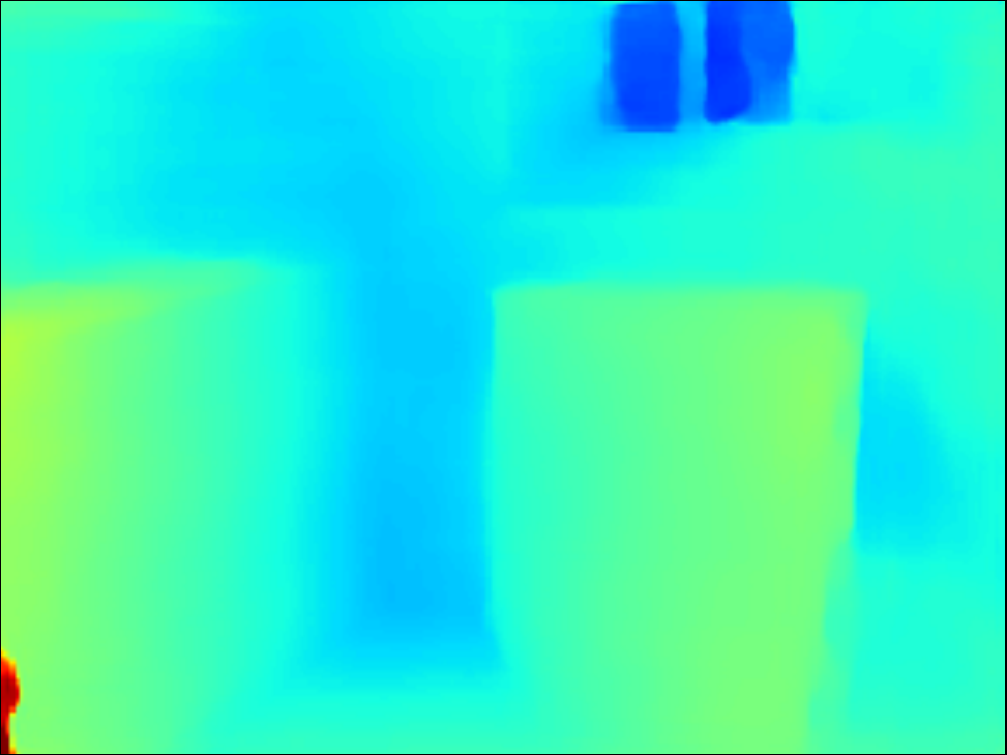}}
\caption{(a) Input image. (b) Sparse depth map obtained by SfM. (c) Error plot with respect to $\beta$. (d) Final output depth.}
\label{fig:search_beta}
\end{figure}

The network with our dehazing cost volume is differential with respect to $A$ and $\beta$.
Standard gradient-based methods can thus be adopted for the optimization problem.
However, we found that an iterative algorithm based on back-propagation easily falls into a local minimum.
Therefore, we perform grid search to find the best solution.
Figure \ref{fig:search_beta} shows an example in which we search for $\beta$ under ground-truth $A$.
Figure \ref{fig:search_beta}(a) shows an input image, and (b) shows the sparse depth map obtained by SfM.
The horizontal axis of (c) represents $\beta$, and we plot the value of Eq. (\ref{eq:objective}) with respect to each $\beta$.
The green dashed line, which represents the ground-truth $\beta$, corresponds to the global minimum.
Figure \ref{fig:search_beta}(d) shows the final output depth of the network with this global optimal solution.

As discussed in Section \ref{sec:estimation_of_airlight}, we can roughly estimate $A$ with the CNN-based estimator.
We initialize $A$ by this estimate.
Let $A_0$ be the output of this estimator, and we search for $\beta_0$ in the predetermined search space $[\beta_{min},\beta_{max}]$ as follows:
\begin{equation}
\beta_0 = \argmin_{\beta \in [\beta_{min}, \beta_{max}]} \sum_{u,v} m(u,v) \rho\Bigl(z_{sfm}(u,v), z_{A_0,\beta}(u,v)\Bigr).
\end{equation}
We then search for $A^*$ and $\beta^*$ that satisfy Eq. (\ref{eq:objective}) in the predetermined search space $[A_0 - \Delta_A, A_0 + \Delta_A]$ and $[\beta_0 - \Delta_{\beta}, \beta_0 + \Delta_{\beta}]$. 
Algorithm \ref{alg1} shows the overall procedure of depth and scattering parameter estimation.

\begin{algorithm}[tb]                      
\caption{Depth and scattering parameter estimation}         
\label{alg1}               
\begin{algorithmic}[0]
\REQUIRE Reference image $I_r$, source images $\{ I_s | s \in \{1,\cdots,S\}\}$, depth estimator $\mathcal{F}$, airlight estimator $\mathcal{G}$, $\beta_{min}$, $\beta_{max}$,$\Delta_{A}$, $\Delta_{\beta}$, and $\mathbf{z}_{sfm}$
\ENSURE $A^*$, $\beta^*$, $\mathbf{z}^*$
\STATE $A_0 \leftarrow \mathcal{G}(I_r)$
\STATE $\beta_0 \leftarrow \argmin_{\beta \in [\beta_{min}, \beta_{max}]} \sum_{u,v} m(u,v) \rho\Bigl(z_{sfm}(u,v), z_{A_0,\beta}(u,v)\Bigr)$
\STATE where $\mathbf{z}_{A,\beta} = \mathcal{F}(A, \beta; I_r, \{I_1, \cdots, I_S \})$
\STATE $A^*, \beta^* \leftarrow \argmin_{A \in \Omega_A, \beta \in \Omega_\beta} \sum_{u,v} m(u,v) \rho\Bigl(z_{sfm}(u,v), z_{A,\beta}(u,v)\Bigr)$
\STATE where $\Omega_A = [A_0 - \Delta_A, A_0 + \Delta_A]$ and $\Omega_\beta = [\beta_0 - \Delta_{\beta}, \beta_0 + \Delta_{\beta}]$
\STATE $\mathbf{z}^* \leftarrow \mathcal{F}(A^*, \beta^*; I_r, \{I_1, \cdots, I_S \})$
\end{algorithmic}
\end{algorithm}

\section{Experiments} \label{sec:experiments}
In this study, we used MVDepthNet \cite{wang18} as a baseline method.
As mentioned previously, the ordinary cost volume is replaced with our dehazing cost volume in the proposed method, so we can directly evaluate the effect of our dehazing cost volume by comparing our method with this baseline method.
We also compared the proposed method with simple sequential methods of dehazing and 3D reconstruction using the baseline method.
DPSNet \cite{im19}, the architecture of which is more complicated such as a multi-scale feature extractor, 3D convolutions, and a cost aggregation module, was also trained on hazy images for further comparison.
In addition to the experiments with synthetic data, we give an example of applying the proposed method to actual foggy scenes.

\subsection{Dataset} \label{sec:dataset}
We used the DeMoN dataset \cite{ummenhofer17} for training.
This dataset consists of the SUN3D \cite{xiao13}, RGB-D SLAM \cite{sturm12}, and MVS datasets \cite{fuhrmann14}, which have sequences of real images.
The DeMoN dataset also has the Scenes11 dataset \cite{chang15,ummenhofer17}, which consists of synthetic images.
Each image sequence in the DeMoN dataset includes RGB images, depth maps, and camera parameters.
In the real-image datasets, most of the depth maps have missing regions due to sensor sensibility.
As we discuss later, we synthesized hazy images from the clean images in the DeMoN dataset for training the proposed method, where we need dense depth maps without missing regions to compute pixel-wise degradation due to haze.
Therefore, we first trained MVDepthNet using clear images then filled the missing regions of each depth map with the output depth of MVDepthNet.
To suppress boundary discontinuities and sensor noise around missing regions, we applied a median filter after depth completion.
For the MVS dataset, which has larger noise than other datasets, we reduced the noise simply by thresholding before inpainting. 
Note that the training loss was computed using only pixels that originally had valid depth values.
We generated 419,046 and 8,842 samples for training and test data, respectively.
Each sample contained one reference image and one source image.
All images were resized to $256 \times 192$.

We synthesized a hazy-image dataset for training the proposed method from clear images.
The procedure of generating a hazy image is based on Eq. (\ref{eq:image_formation_model}).
For $A$, we randomly sampled $A \in [0.7, 1.0]$ for each data sample.
For $\beta$, we randomly sampled $\beta \in [0.4,0.8],[0.4,0.8],[0.05,0.15]$ for the SUN3D, RGB-D SLAM, and Scenes11 datasets, respectively.
We found that for the MVS dataset, it was difficult to determine the same sampling range of $\beta$ for all images because it contains various scenes with different depth scales.
Therefore, we determined the sampling range of $\beta$ for each sample of the MVS dataset as follows. We first set the range of a transmission map $e^{-\beta z}$ to $[0.2,0.4]$ for all samples then computed the median of a depth map $z_{med}$ for each sample. 
Finally, we determined the $\beta$ range for each sample as $\beta \in [-\log (0.4) / z_{med}, -\log (0.2) / z_{med}]$.

Similar to Wang and Shen \cite{wang18}, we adopted data augmentation to enable the network to reconstruct a wide depth range.
The depth of each sample was scaled by a factor between 0.5 and 1.5 together with the translation vector of the camera.
Note that when training the proposed method, $\beta$ should also be scaled by the inverse of the scale factor.

\subsection{Training details}
All networks were implemented in PyTorch. The training was done on a NVIDIA V100 GPU with 32-GB memory. 
The size of a minibatch was 32 for all training.

We first trained MVDepthNet from scratch on the clear-image dataset.
We used Adam \cite{kingma15} with a learning rate of $1.0\times10^{-4}$.
After the initial 100K iterations, the learning rate was reduced by $20\%$ after every 20K iterations.

We then fine-tuned MVDepthNet on hazy images and trained the proposed method with our dehazing cost volume.
The parameters of both methods were initialized by that of the trained MVDepthNet on clear images.
The initial learning rate was set to $1.0\times10^{-4}$ and reduced by $20\%$ after every 20K iterations.

We also trained the dehazing methods, AOD-Net \cite{li17} and FFA-Net \cite{qin20}, and the MVS method DPSNet \cite{im19} on our hazy image dataset for comparison.
The dehazing networks were followed by MVDepthNet trained on clear images for depth estimation.
DPSNet was trained with the same loss function and learning schedule as in the original paper \cite{im19}.

\begin{table*}[tb]
\centering
   \caption{Quantitative results. We compared proposed method (MVDepthNet w/ dcv) with MVDepthNet \cite{wang18} fine-tuned on hazy images (MVDepthNet), simple sequential methods of dehazing \cite{li17,qin20} and depth estimation with MVDepthNet (AOD-Net + MVDepthNet, FFA-Net + MVDepthNet), and DPSNet \cite{im19} trained on hazy images (DPSNet). Red and blue values are best and second-best, respectively. }
   \label{tab:results}
   \begin{tabular}{cccccc} \hline
   Dataset & Method & L1-rel & L1-inv &sc-inv & C.P. (\%) \\ \hline
   \multirow{5}{*}{SUN3D} & AOD-Net + MVDepthNet & 0.249 & 0.132 &0.250 & 47.8 \\ 
   & FFA-Net + MVDepthNet & 0.180 & 0.111 &0.211 & 55.5 \\ 
   & MVDepthNet & 0.155  & 0.093 & 0.184  & 60.3 \\
   & DPSNet & \textcolor{blue}{0.145} & \textcolor{blue}{0.082} & \textcolor{blue}{0.183} & \textcolor{blue}{64.7}\\
   & {\bf MVDepthNet w/ dcv} & \textcolor{red}{0.100} & \textcolor{red}{0.058}  & \textcolor{red}{0.161} & \textcolor{red}{79.0} \\ \hline
   \multirow{5}{*}{RGB-D SLAM} & AOD-Net + MVDepthNet & 0.205 & 0.127 & 0.315 & 58.9\\ 
   & FFA-Net + MVDepthNet & 0.179 & 0.114 &0.288 & 65.0 \\ 
   & MVDepthNet  & \textcolor{blue}{0.157}  & 0.091 & 0.254 & \textcolor{blue}{70.7} \\
   & DPSNet & \textcolor{red}{0.152} & \textcolor{blue}{0.090} & \textcolor{blue}{0.234} & \textcolor{red}{71.6} \\
   & {\bf MVDepthNet w/ dcv} & 0.162 & \textcolor{red}{0.089} & \textcolor{red}{0.231} & 68.8 \\ \hline
   \multirow{5}{*}{MVS} & AOD-Net + MVDepthNet & 0.323 & 0.123 & 0.309 & 51.9 \\ 
   & FFA-Net + MVDepthNet & 0.215 & 0.112 &0.288 & 55.6 \\ 
   & MVDepthNet & \textcolor{blue}{0.184} & 0.100 & 0.241 & 57.1 \\
   & DPSNet & 0.191 & \textcolor{red}{0.088} & \textcolor{blue}{0.239} & \textcolor{red}{67.9} \\
   & {\bf MVDepthNet w/ dcv} & \textcolor{red}{0.160} & \textcolor{blue}{0.091} & \textcolor{red}{0.222} & \textcolor{blue}{58.1} \\ \hline
   \multirow{5}{*}{Scenes11} & AOD-Net + MVDepthNet & 0.330 & 0.036 & 0.539 & 52.3 \\ 
   & FFA-Net + MVDepthNet  & 0.377 & 0.041 &0.600 & 51.3 \\ 
   & MVDepthNet & 0.151 & 0.022  & \textcolor{blue}{0.279} & 64.0 \\
   & DPSNet  & \textcolor{red}{0.105} & \textcolor{red}{0.018} & 0.381 & \textcolor{red}{81.8} \\
   & {\bf MVDepthNet w/ dcv} & \textcolor{blue}{0.134} & \textcolor{blue}{0.019}  & \textcolor{red}{0.216} & \textcolor{blue}{72.3} \\ \hline
   \end{tabular}
\end{table*}

\begin{figure*}[tb]
\centering
\begin{minipage}[b]{0.15\hsize}
\centering
\subfloat[][\centering Clear 

			image]{
   \begin{tabular}{c}
      \includegraphics[width=1.0\textwidth]{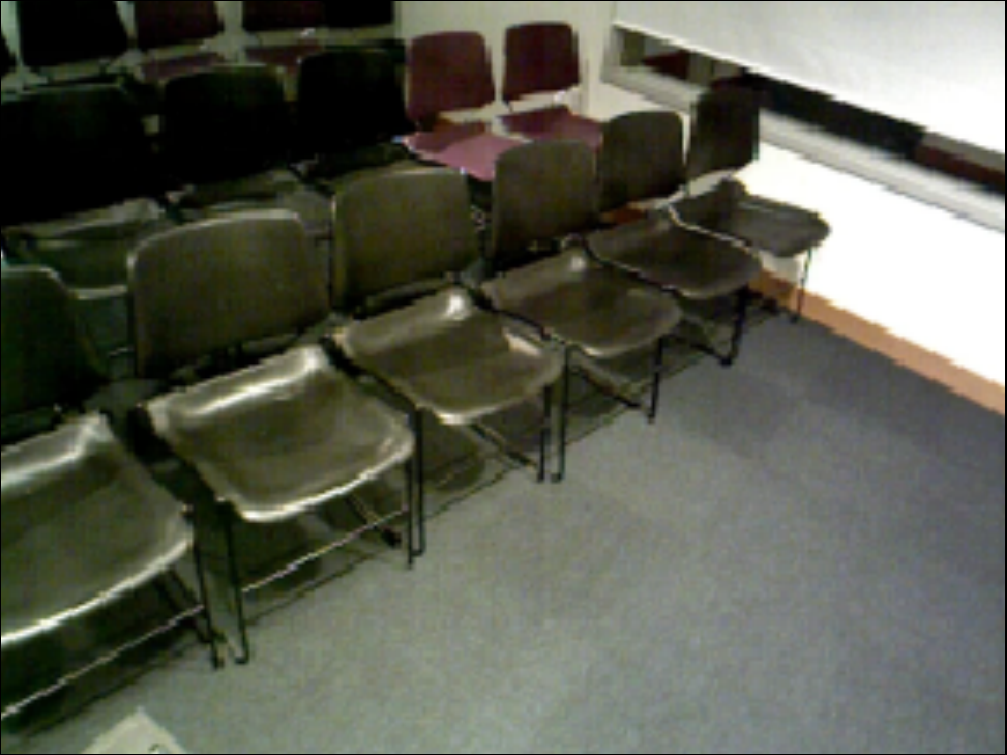}\\
      \includegraphics[width=1.0\textwidth]{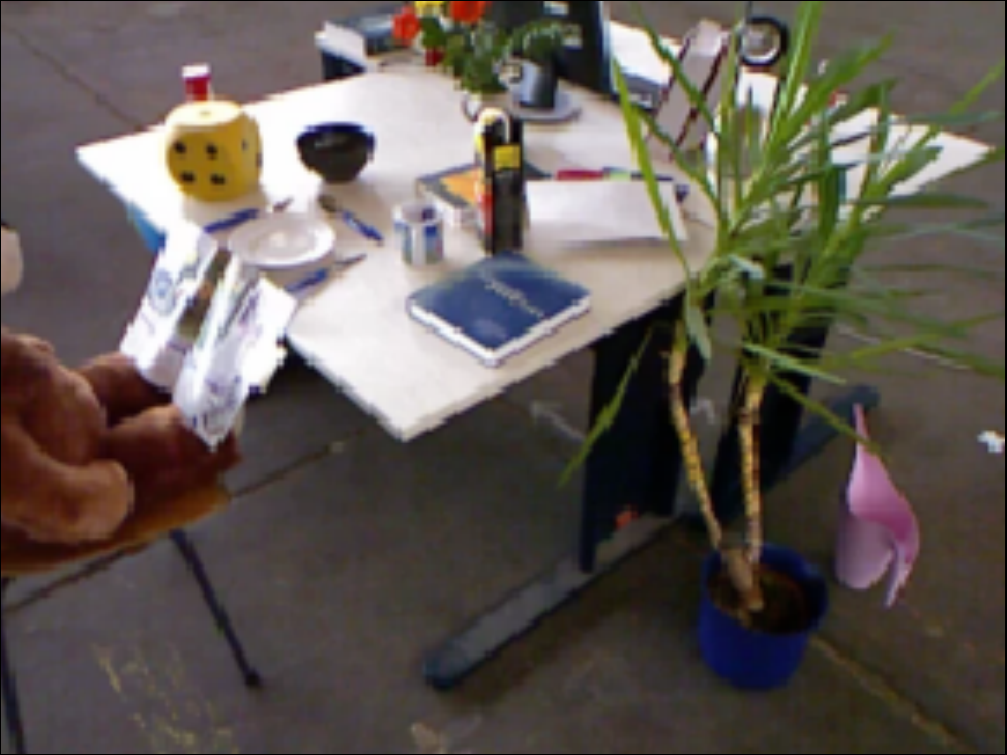}\\
      \includegraphics[width=1.0\textwidth]{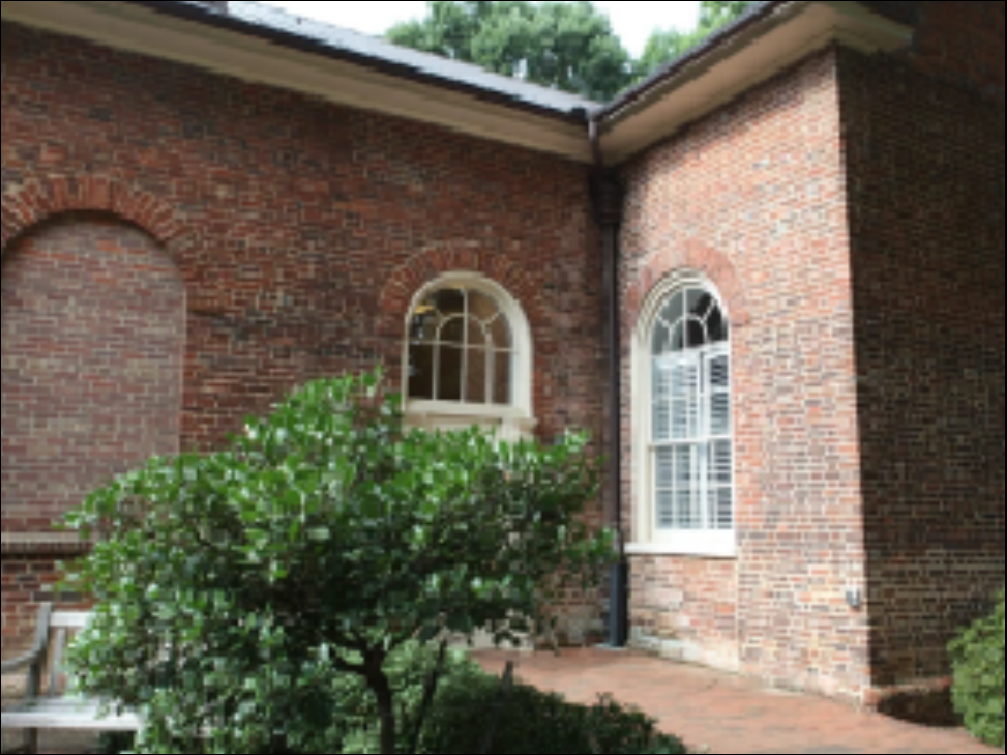}\\
      \includegraphics[width=1.0\textwidth]{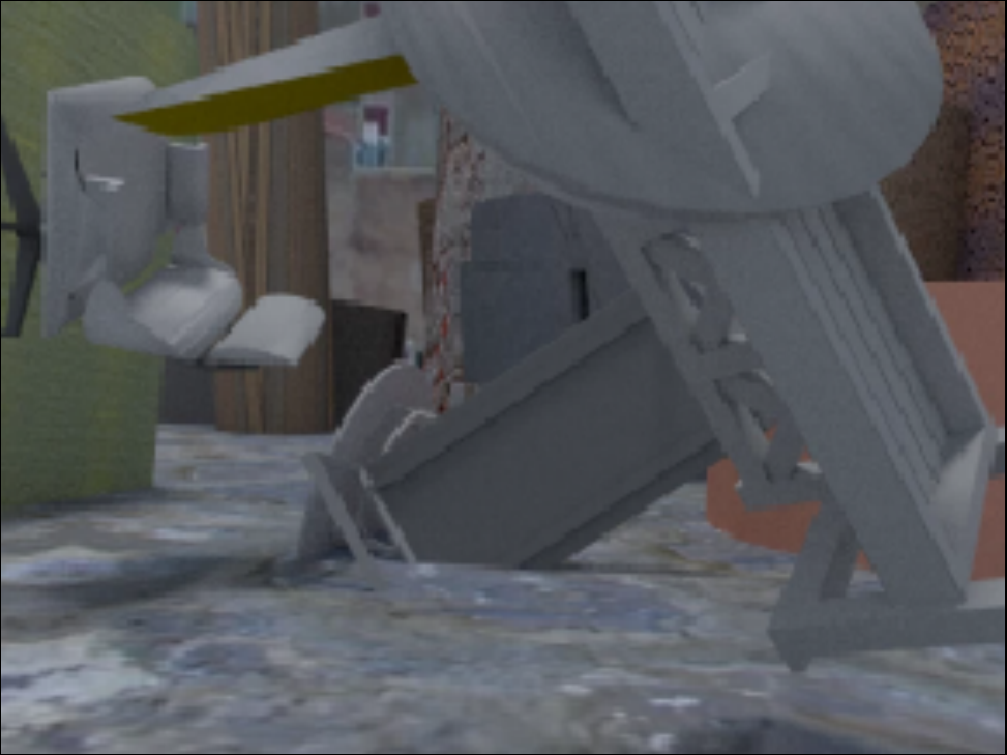}
   \end{tabular}}
\end{minipage}
\begin{minipage}[b]{0.15\hsize}
\centering
\subfloat[][\centering Hazy 

			input]{
   \begin{tabular}{c}
      \includegraphics[width=1.0\textwidth]{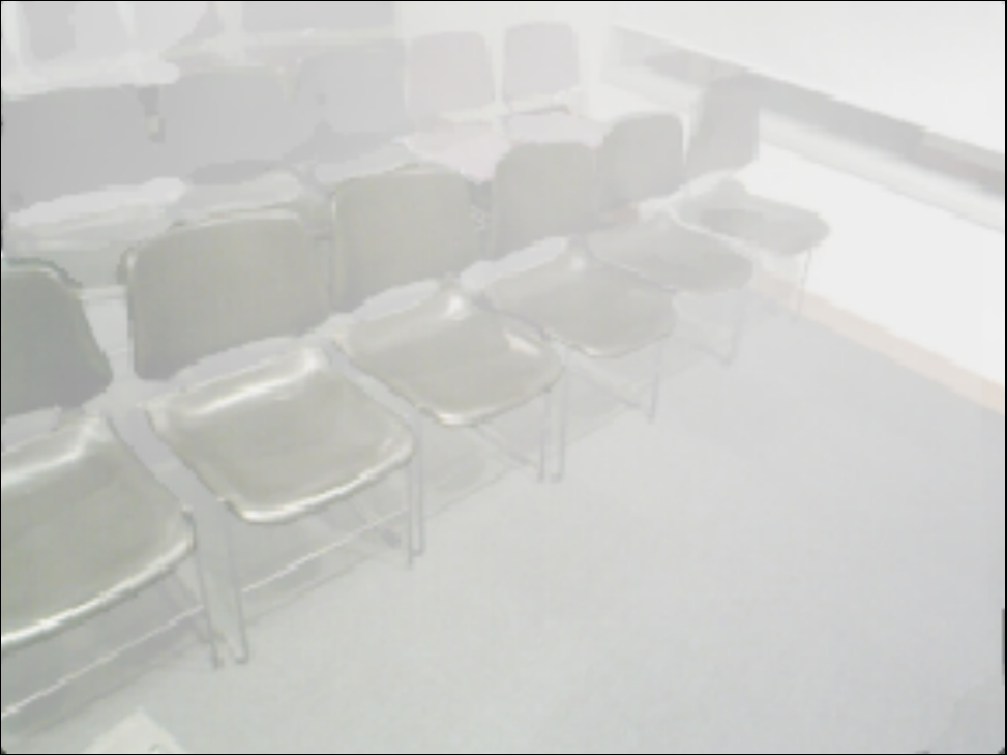}\\
      \includegraphics[width=1.0\textwidth]{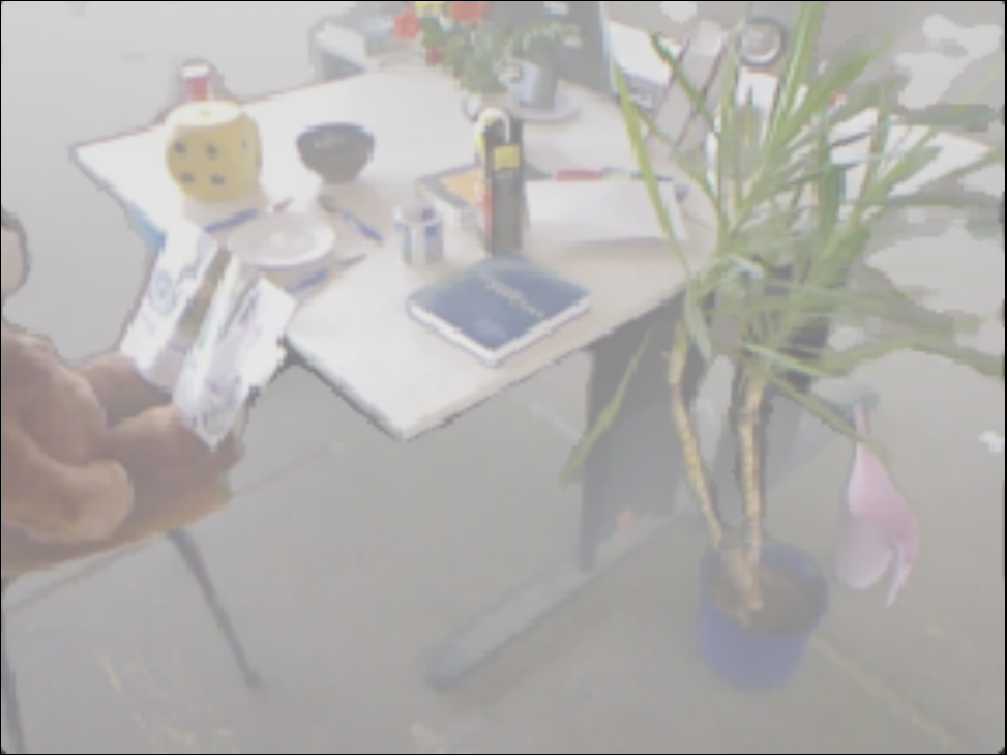}\\
      \includegraphics[width=1.0\textwidth]{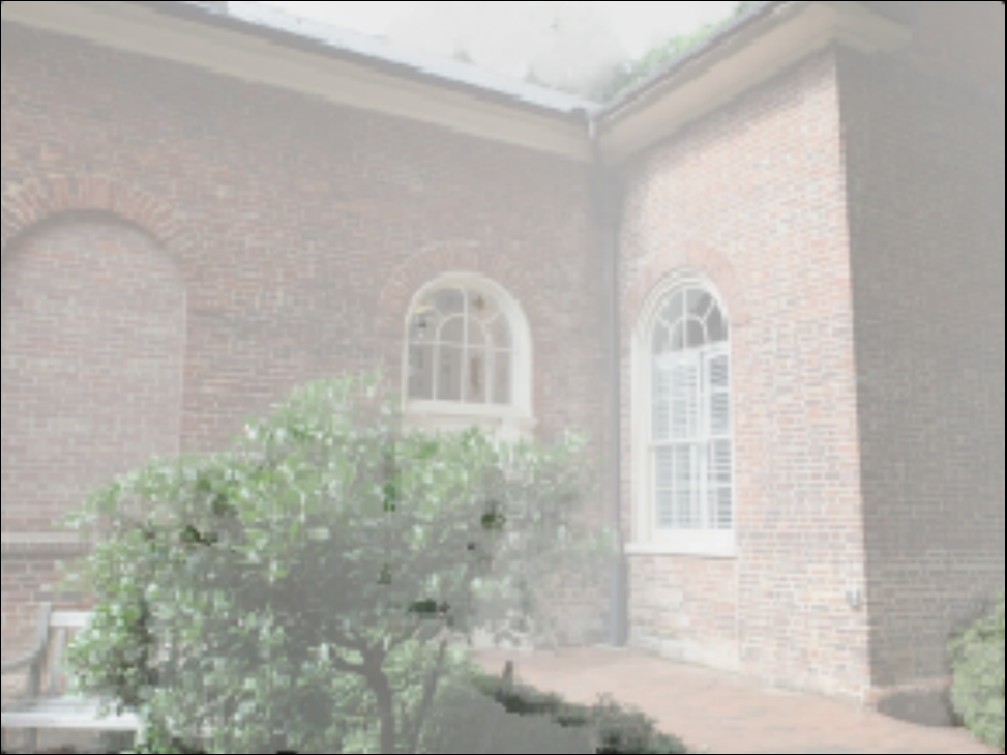}\\
      \includegraphics[width=1.0\textwidth]{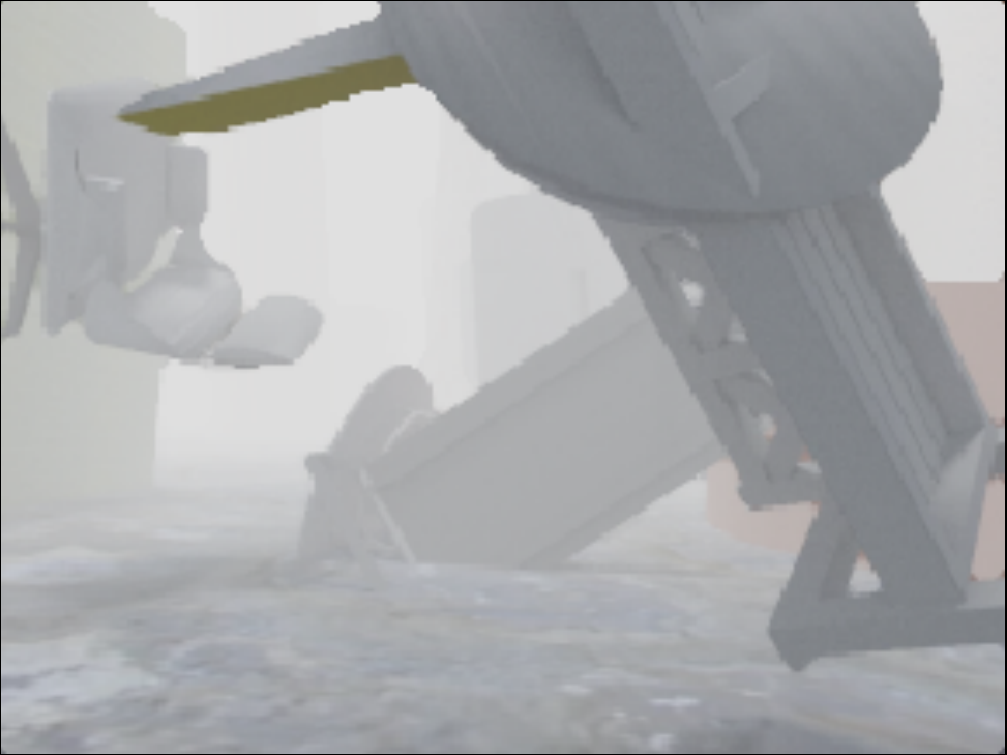}
   \end{tabular}} 
\end{minipage}
\begin{minipage}[b]{0.15\hsize}
\centering
\subfloat[][\centering Ground 

			truth]{
   \begin{tabular}{c}
      \includegraphics[width=1.0\textwidth]{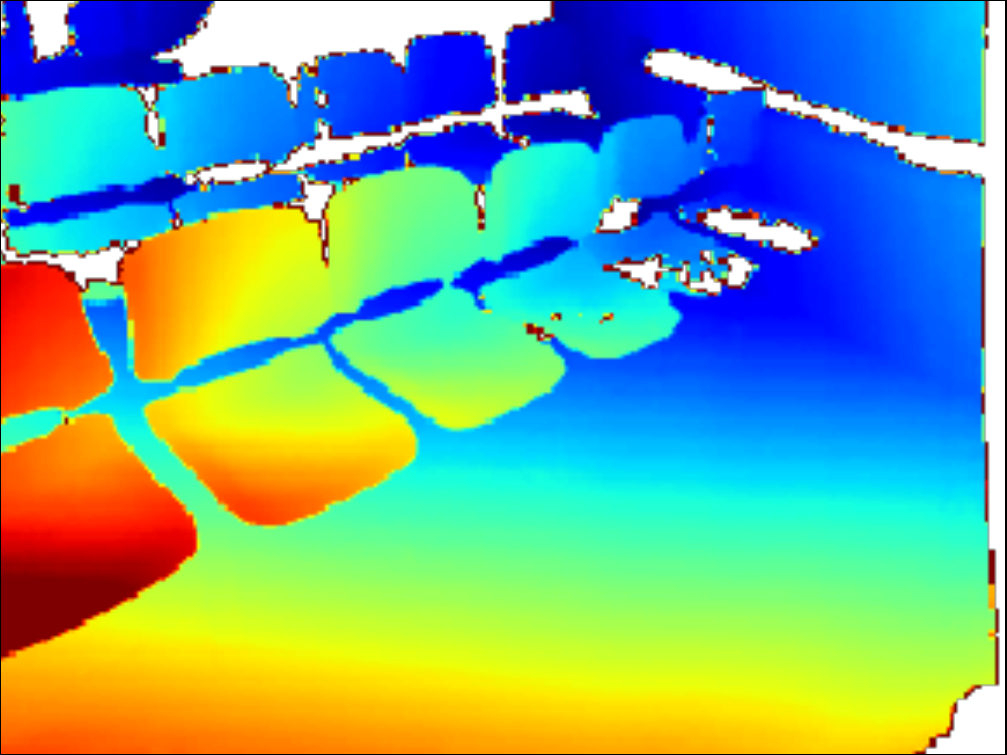}\\
      \includegraphics[width=1.0\textwidth]{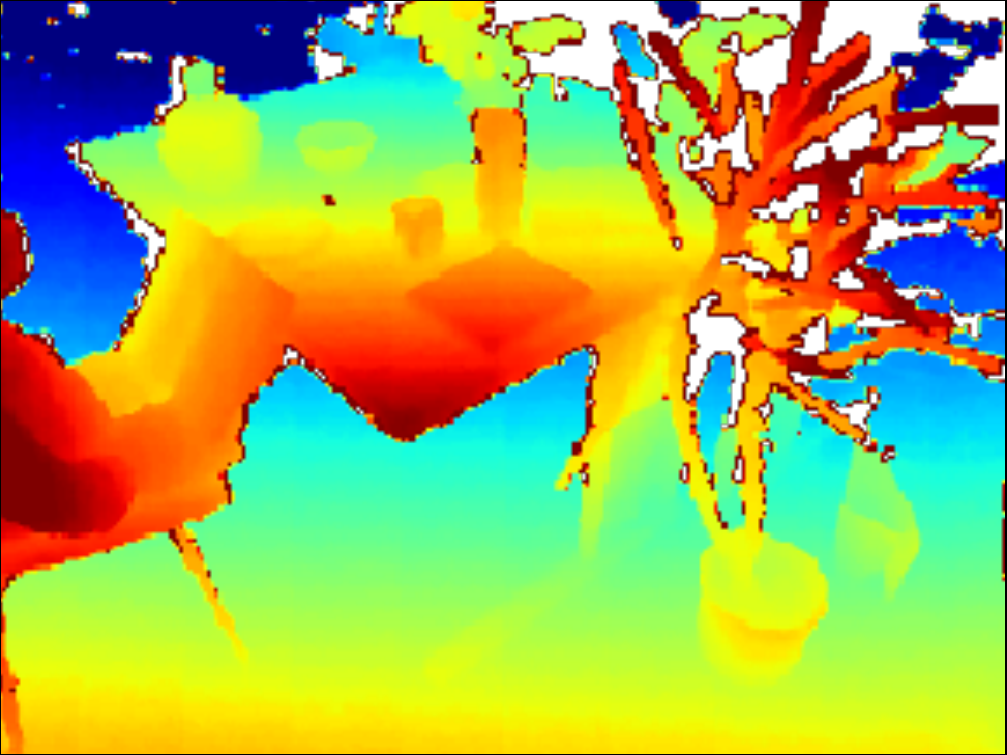}\\
      \includegraphics[width=1.0\textwidth]{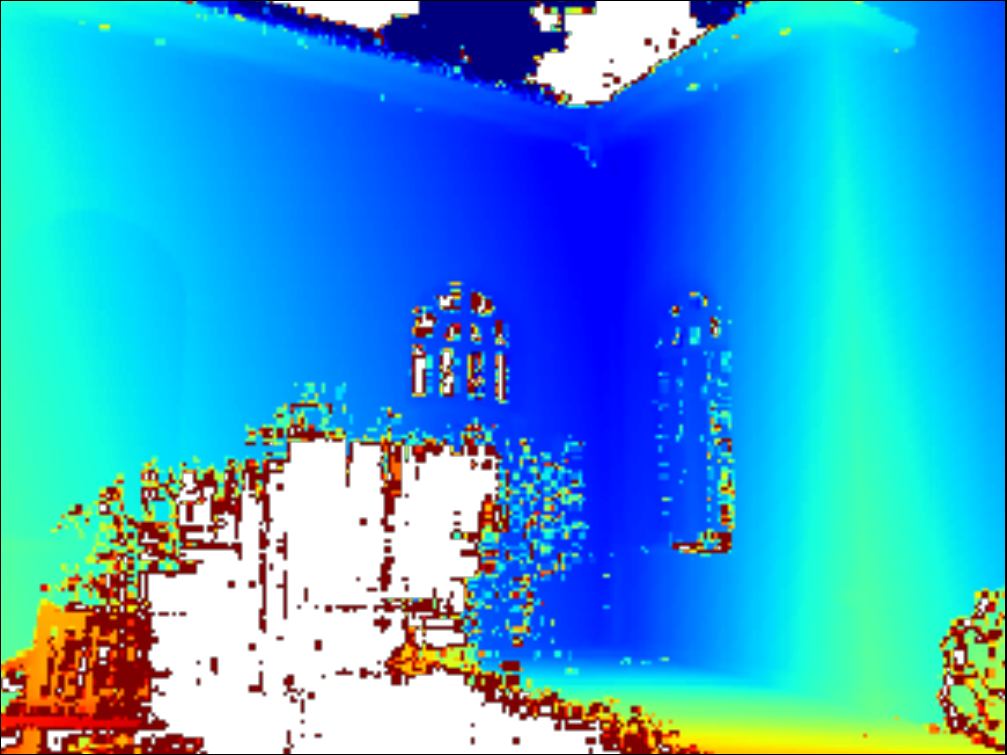}\\
      \includegraphics[width=1.0\textwidth]{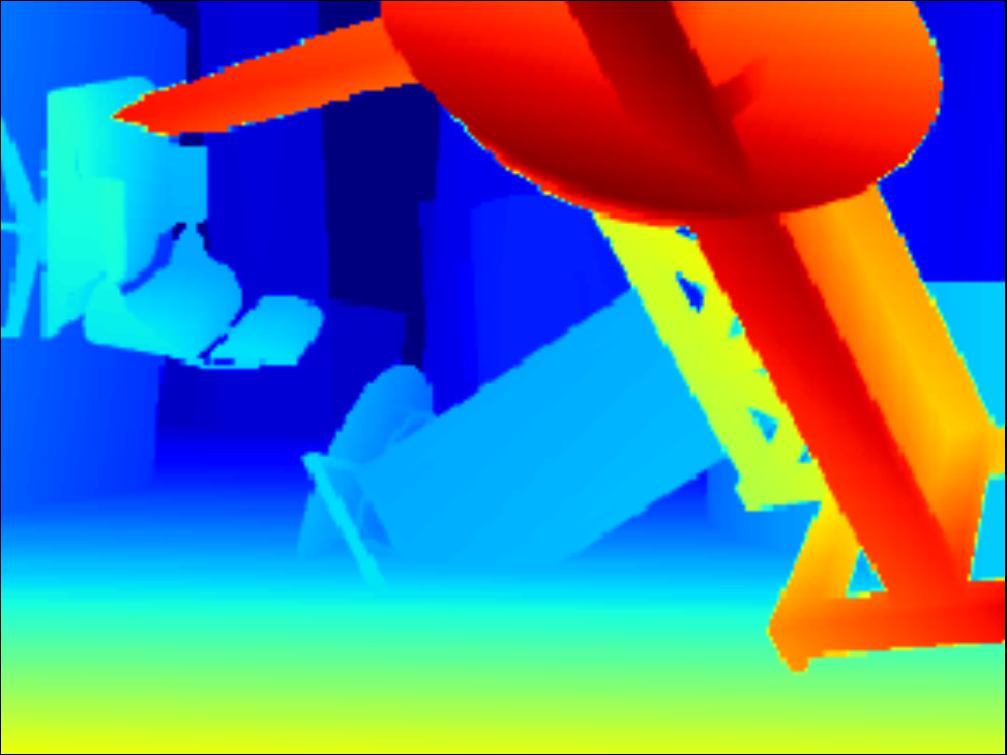}
   \end{tabular}}
\end{minipage}
\begin{minipage}[b]{0.15\hsize}
\centering
\subfloat[][\centering MVDepthNet

         	 \cite{wang18}]{
   \begin{tabular}{c}
      \includegraphics[width=1.0\textwidth]{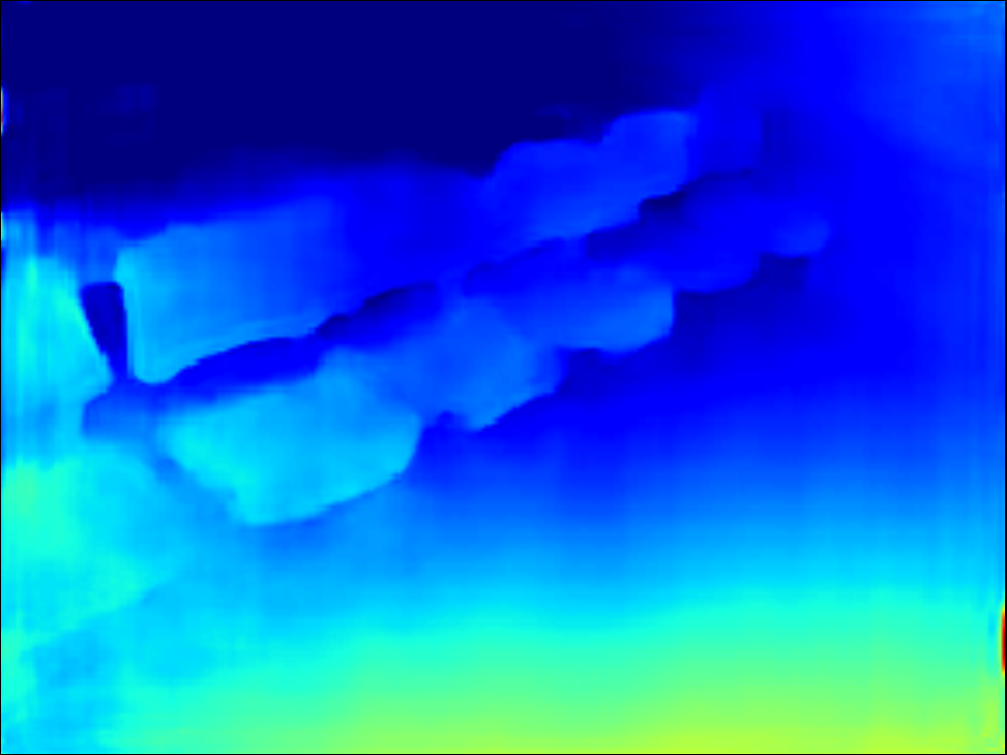}\\
      \includegraphics[width=1.0\textwidth]{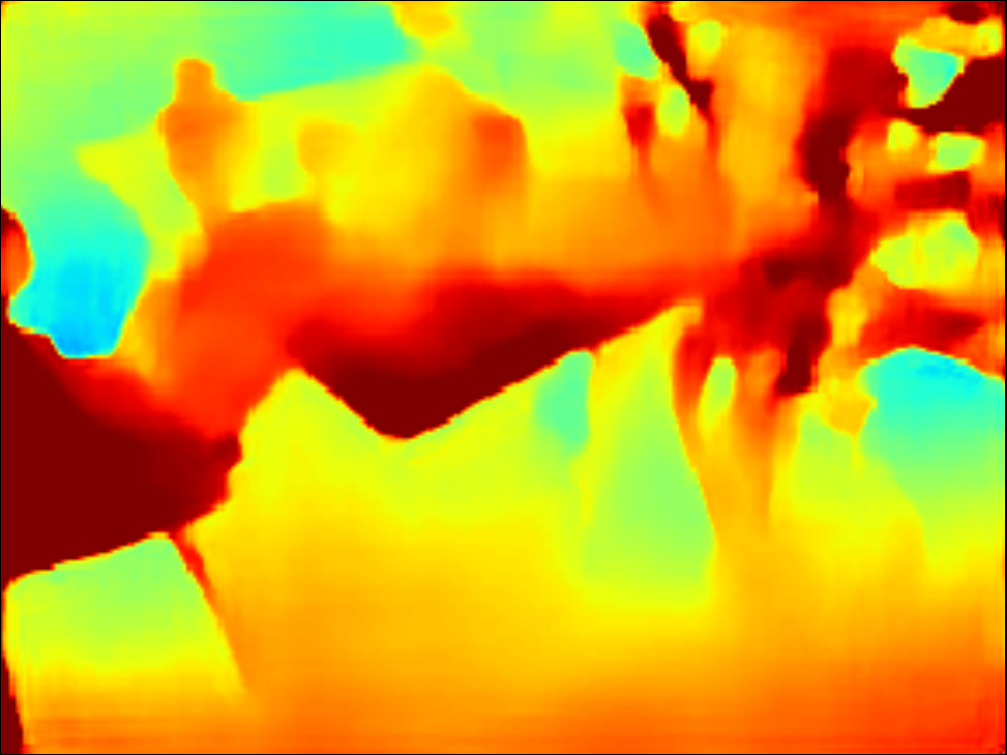}\\
      \includegraphics[width=1.0\textwidth]{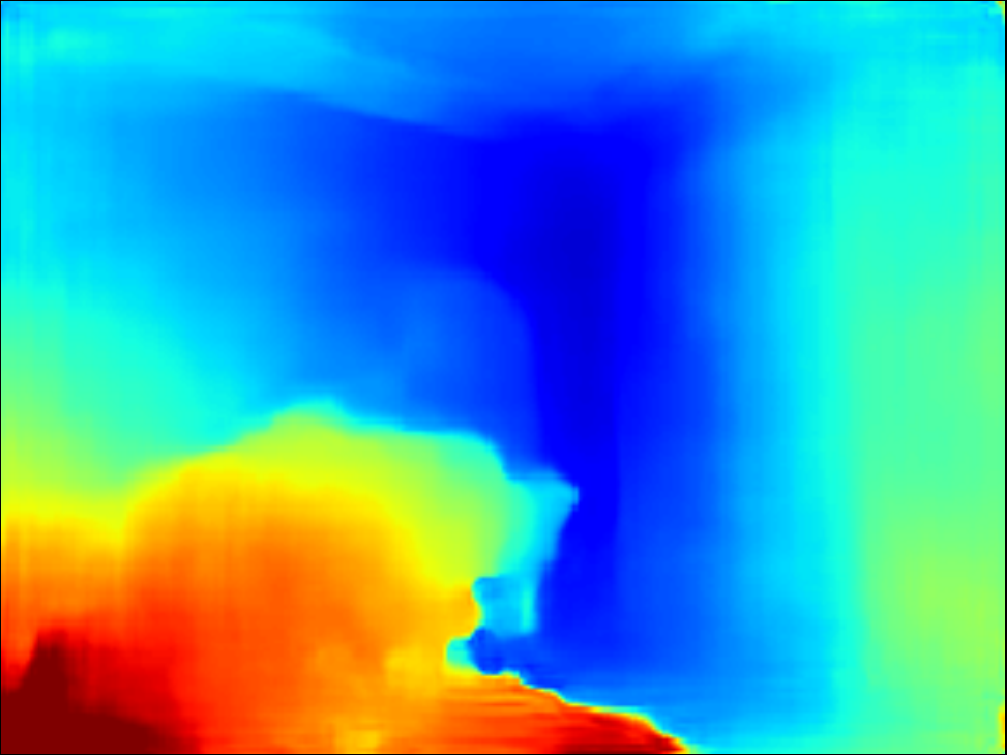}\\
      \includegraphics[width=1.0\textwidth]{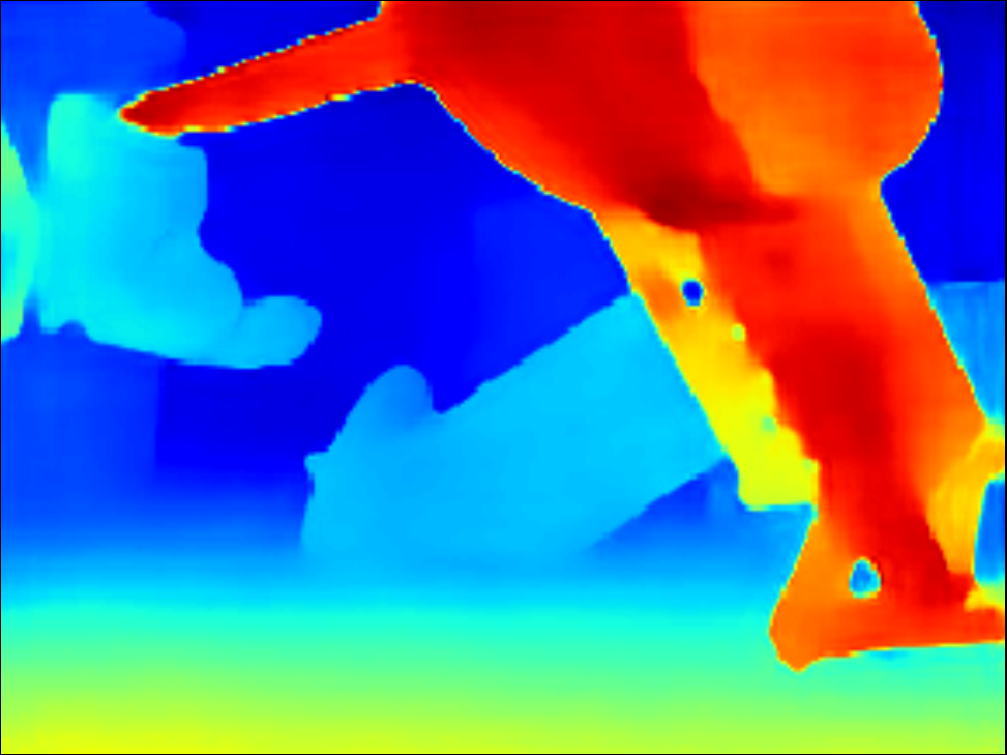}
   \end{tabular}}
\end{minipage}
\begin{minipage}[b]{0.15\hsize}
\centering
\subfloat[][\centering DPSNet

			\cite{im19}]{
   \begin{tabular}{c}
      \includegraphics[width=1.0\textwidth]{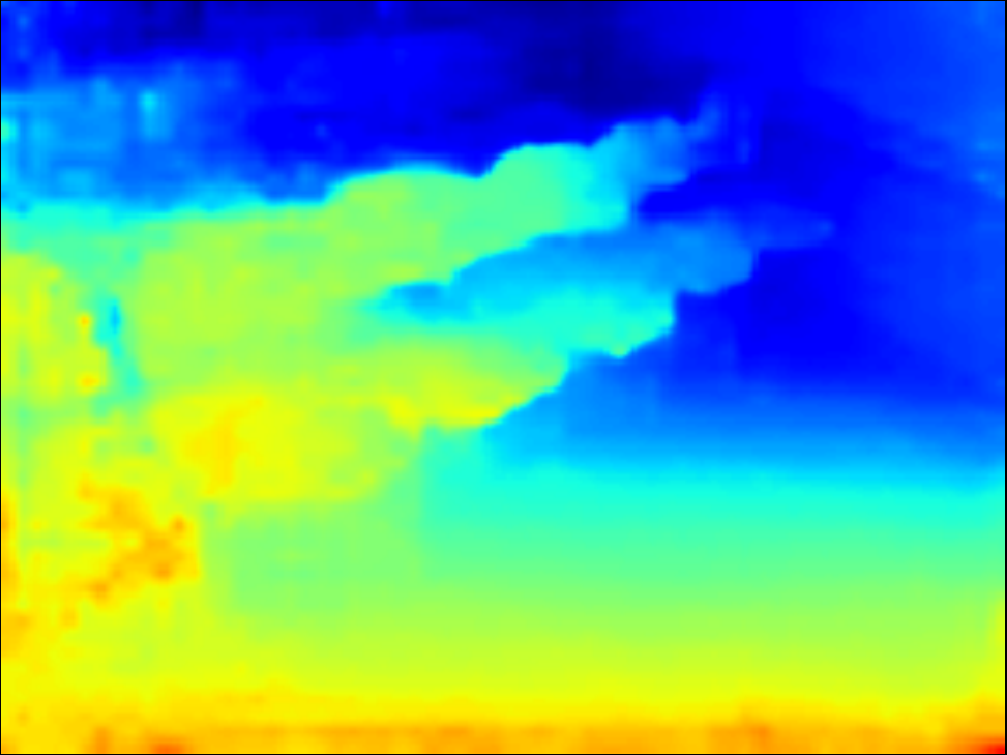}\\
      \includegraphics[width=1.0\textwidth]{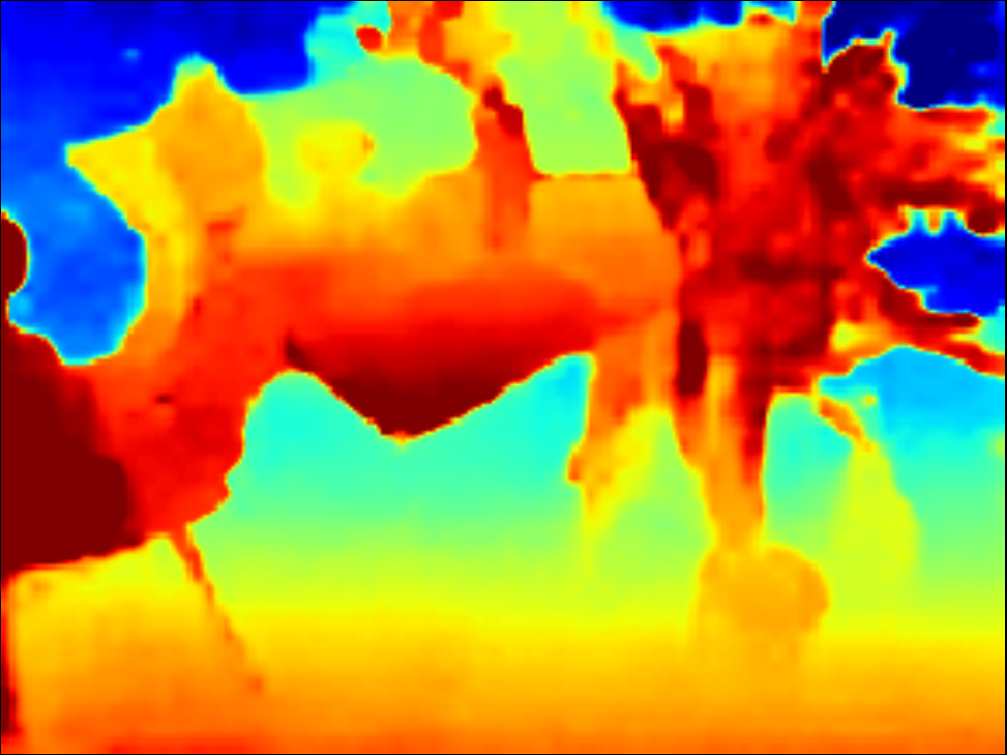}\\
      \includegraphics[width=1.0\textwidth]{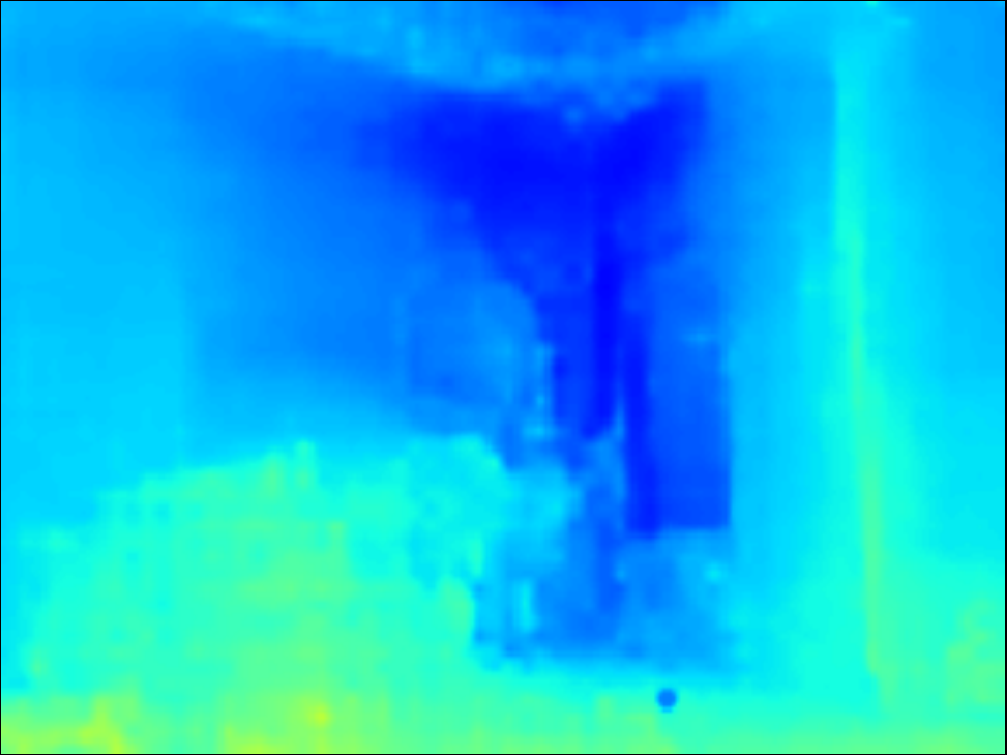}\\
      \includegraphics[width=1.0\textwidth]{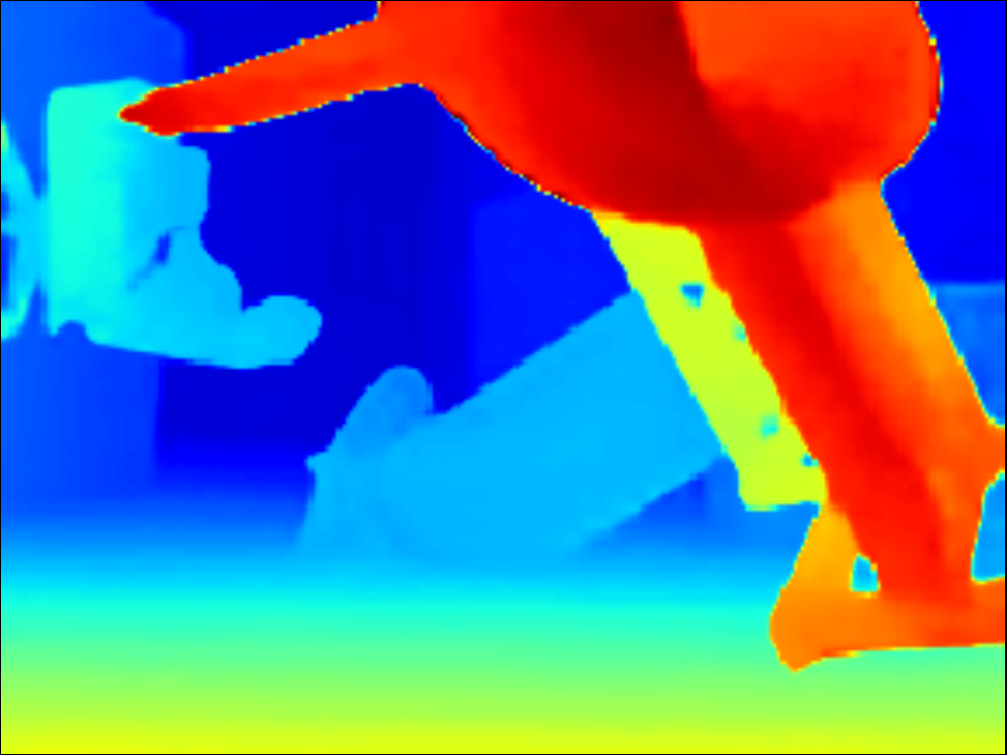}
   \end{tabular}}
\end{minipage}
\begin{minipage}[b]{0.15\hsize}
\centering
\subfloat[Proposed]{
   \begin{tabular}{c}
      \includegraphics[width=1.0\textwidth]{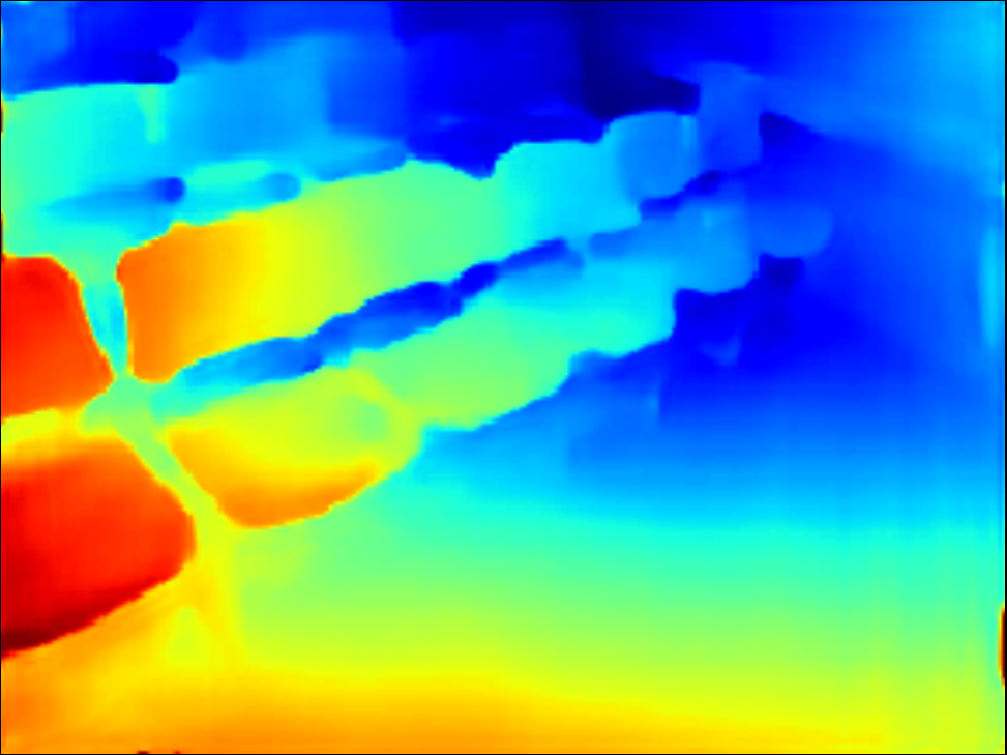}\\
      \includegraphics[width=1.0\textwidth]{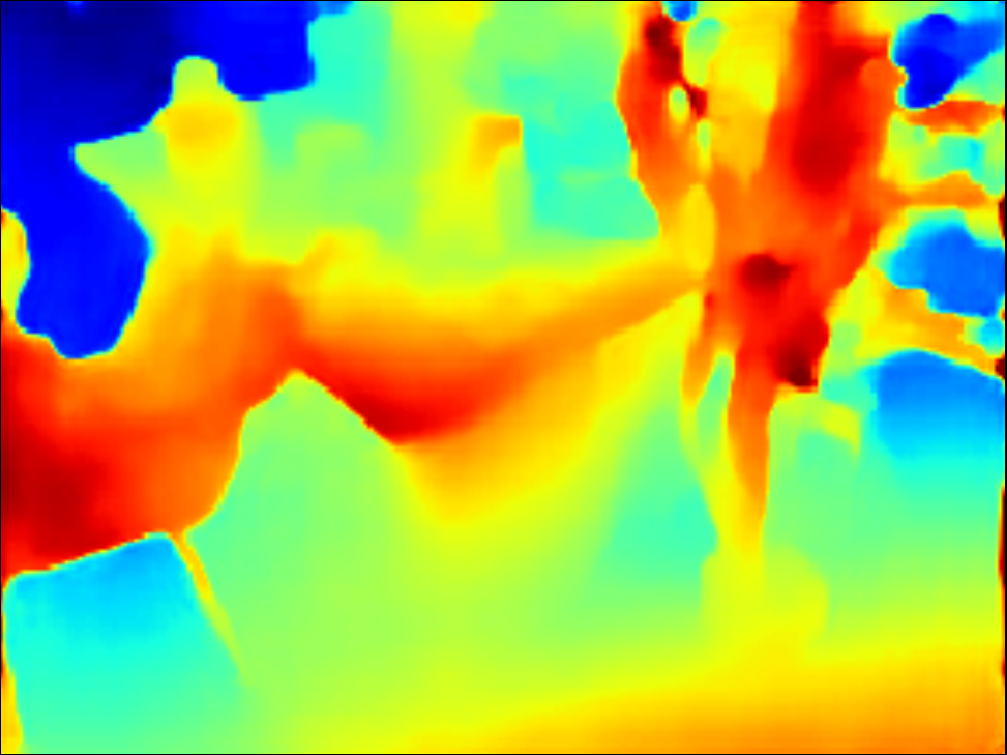}\\
      \includegraphics[width=1.0\textwidth]{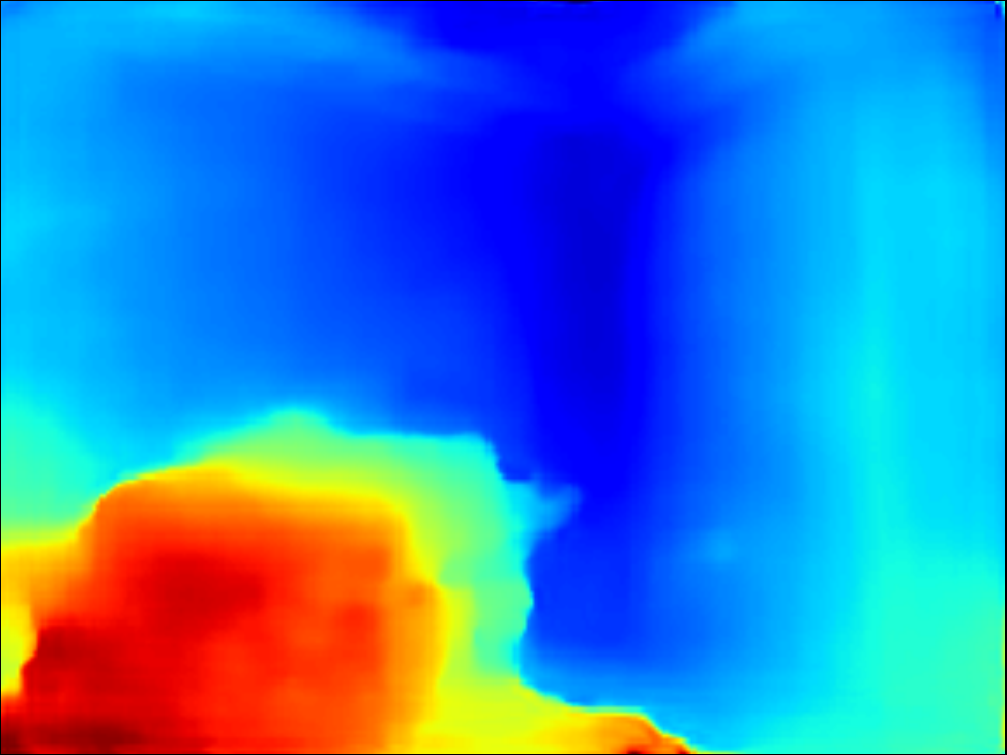}\\
      \includegraphics[width=1.0\textwidth]{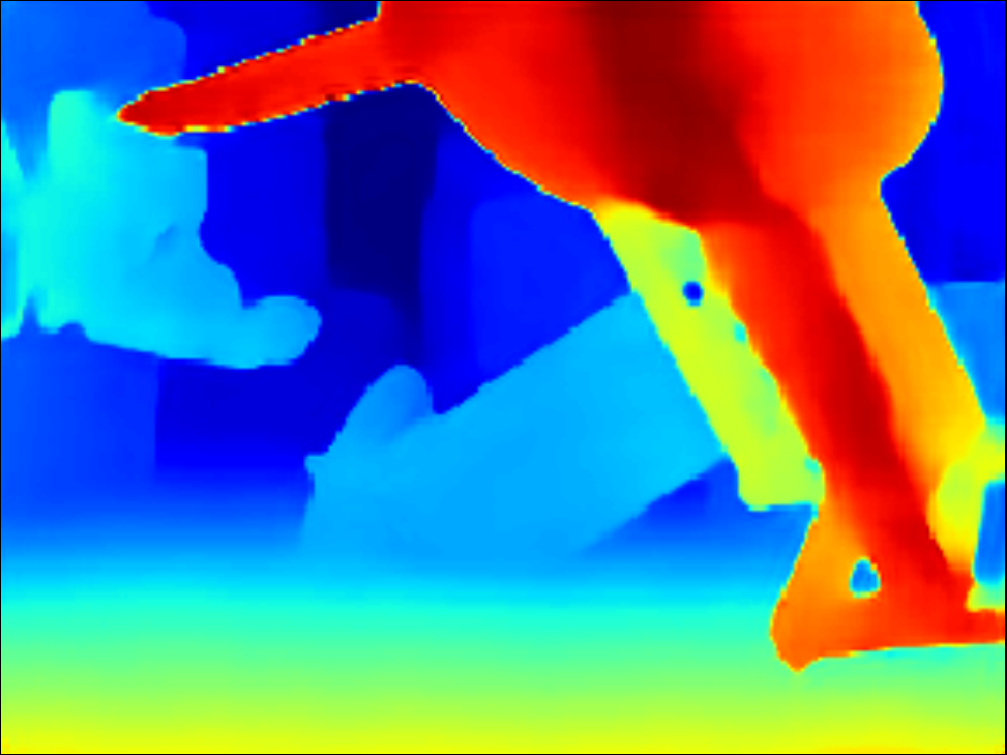}
   \end{tabular}}
\end{minipage}
\caption{Qualitative results. (a) clear image, (b) hazy input, (c) ground-truth depth, (d) output of fine-tuned MVDepthNet \cite{wang18}, (e) output of DPSNet \cite{im19}, and (f) output of proposed method. From top to bottom, each row shows results of input images in SUN3D, RGB-D SLAM, MVS, and Scenes11 datasets, respectively. }
\label{fig:results}
\end{figure*}

\begin{table*}[tb]
\centering
	\caption{Quantitative results of depth and scattering parameter estimation. ``MVDepthNet w/ dcv, pe'' denotes the proposed method with scattering parameter estimation. Red and blue values are best and second-best, respectively. As evaluation metric of $A$ and $\beta$, we used mean absolute error (MAE$_A$ and MAE$_\beta$).}
	\label{tab:params_result}
   \begin{tabular}{cccccccc} \hline 
   Dataset & Method & L1-rel & L1-inv & sc-inv & C.P. (\%) & MAE$_A$ & MAE$_\beta$\\ \hline \hline
   \multirow{5}{*}{\begin{tabular}{c} L1-rel $\leq 0.1$ \\ 1364 samples\end{tabular}}&FFA-Net + MVDepthNet  & 0.141 & 0.104 & 0.152 & 57.0 & - & - \\
   &MVDepthNet & 0.130 & 0.090 & 0.135 & 59.9 & - & - \\
   &DPSNet & 0.109 & 0.069 & 0.125 & 65.2 & - & - \\
   &MVDepthNet w/ dcv  & \textcolor{red}{0.069} & \textcolor{red}{0.043} & \textcolor{red}{0.104} & \textcolor{red}{80.7} & - & - \\ \cline{2-8}
   &MVDepthNet w/ dcv, pe & \textcolor{blue}{0.081} & \textcolor{blue}{0.050} & \textcolor{blue}{0.116} & \textcolor{blue}{76.3} & 0.028 & 0.043  \\ \hline \hline
   \multirow{5}{*}{\begin{tabular}{c} L1-rel $\leq 0.2$ \\ 2661 samples\end{tabular}}&FFA-Net  + MVDepthNet  & 0.154 & 0.102 & 0.172 & 52.4 & - & - \\
   &MVDepthNet & 0.138 & 0.088 & 0.152 & 56.0 & - & - \\
   &DPSNet & 0.120 & 0.072 & 0.138 & 61.1 & - & - \\
   &MVDepthNet w/ dcv  & \textcolor{red}{0.077} & \textcolor{red}{0.044} & \textcolor{red}{0.116} & \textcolor{red}{78.4} & - & - \\ \cline{2-8}
   &MVDepthNet w/ dcv, pe & \textcolor{blue}{0.092} & \textcolor{blue}{0.053} & \textcolor{blue}{0.132} & \textcolor{blue}{72.9} & 0.028 & 0.042 \\ \hline \hline
   \multirow{5}{*}{\begin{tabular}{c} L1-rel $\leq 0.3$ \\ 3157 samples\end{tabular}}&FFA-Net  + MVDepthNet  & 0.162 & 0.103 & 0.182 & 50.7 & - & - \\
   &MVDepthNet & 0.143 & 0.089 & 0.158 & 54.7 & - & - \\
   &DPSNet & 0.124 & 0.072 & 0.144 & 59.9 & - & - \\
   &MVDepthNet w/ dcv & \textcolor{red}{0.079} & \textcolor{red}{0.045} & \textcolor{red}{0.120} & \textcolor{red}{77.6} & - & - \\ \cline{2-8}
   &MVDepthNet w/ dcv, pe & \textcolor{blue}{0.100} & \textcolor{blue}{0.056} & \textcolor{blue}{0.141} & \textcolor{blue}{70.3} & 0.027 & 0.044 \\ \hline \hline
   \end{tabular}
\end{table*}

\begin{figure*}[tb]
\centering
\begin{minipage}{0.14\hsize}
	\centering
	\subfloat[]{
	\begin{tabular}{c}
		\includegraphics[width=1.0\textwidth]{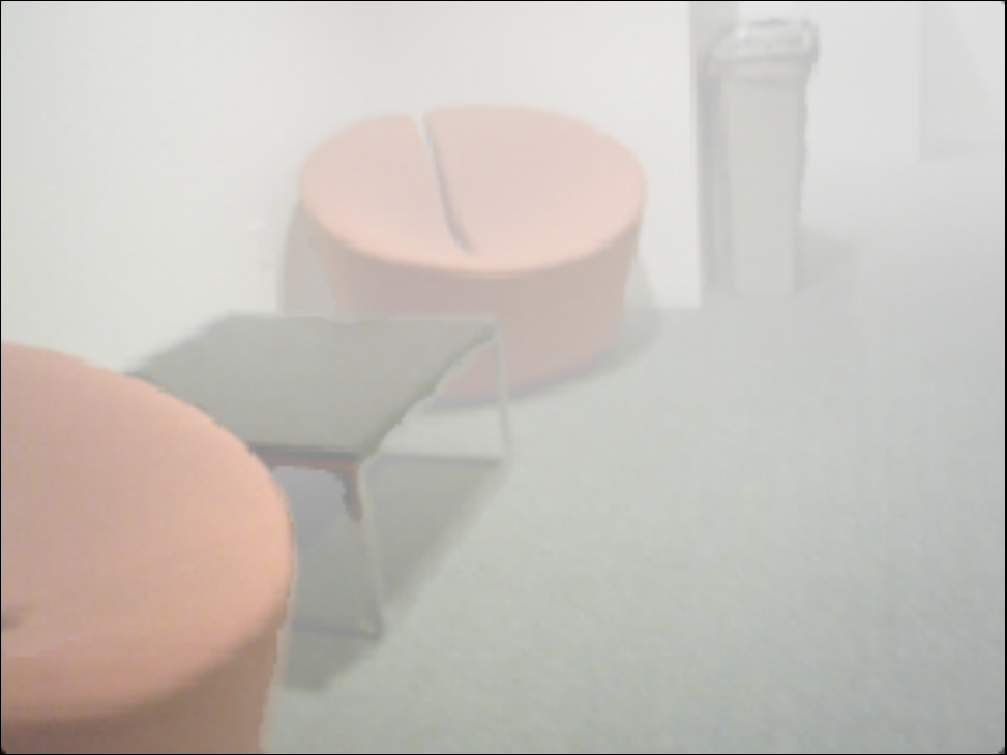}\\
		\includegraphics[width=1.0\textwidth]{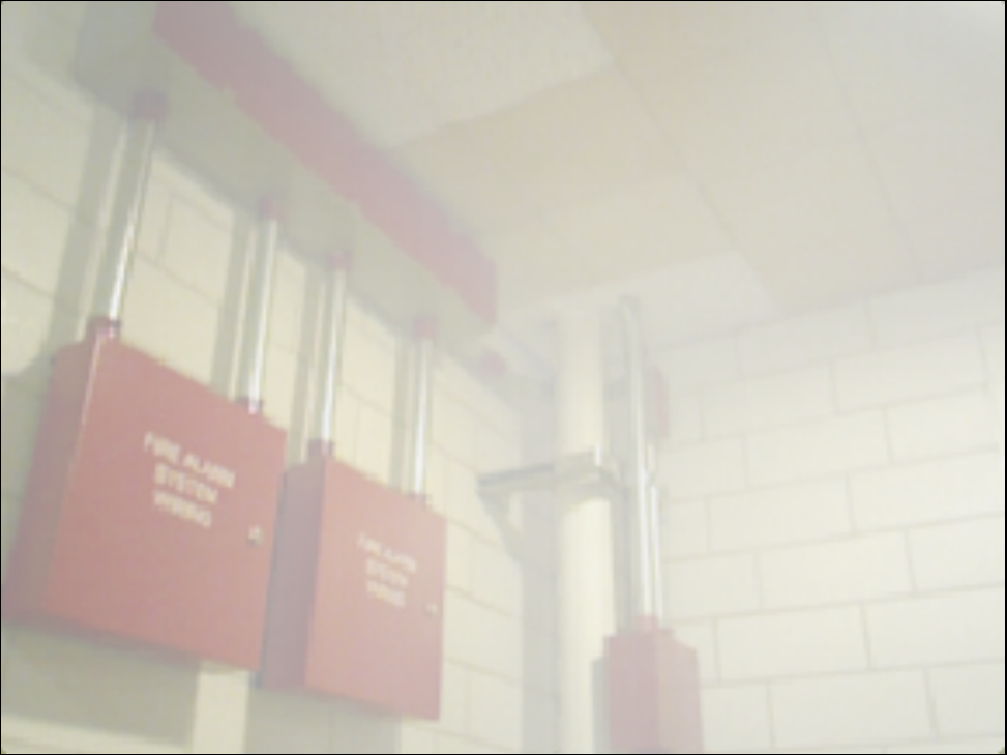}\\
		\includegraphics[width=1.0\textwidth]{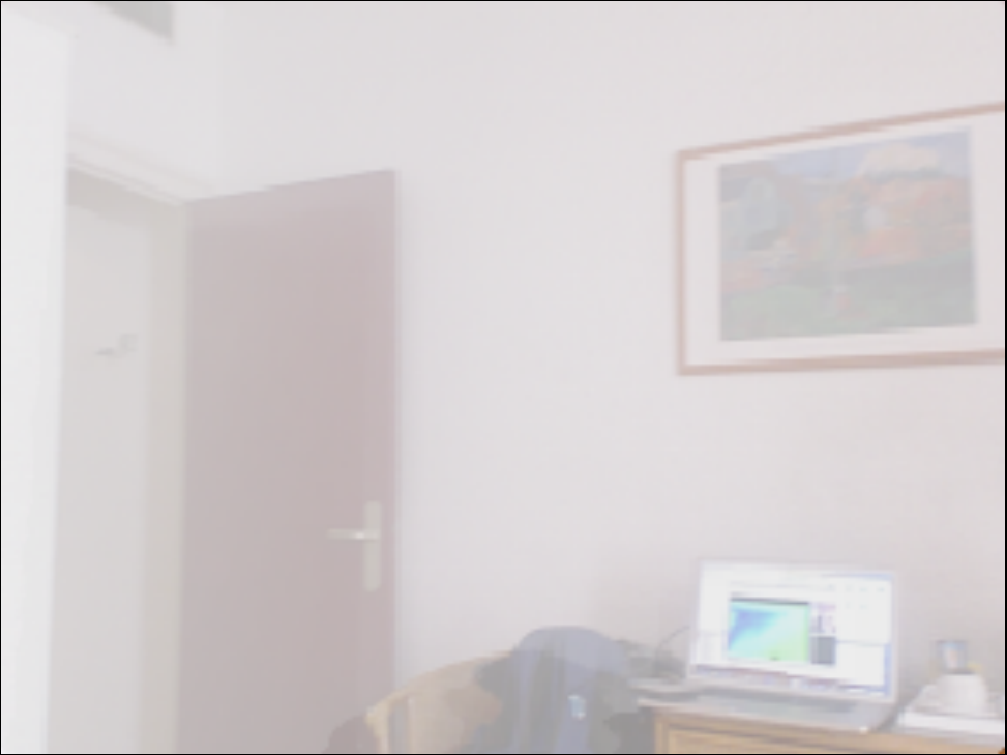}
	\end{tabular}
	}
\end{minipage}
\begin{minipage}{0.14\hsize}
	\centering
	\subfloat[]{
	\begin{tabular}{c}
		\includegraphics[width=1.0\textwidth]{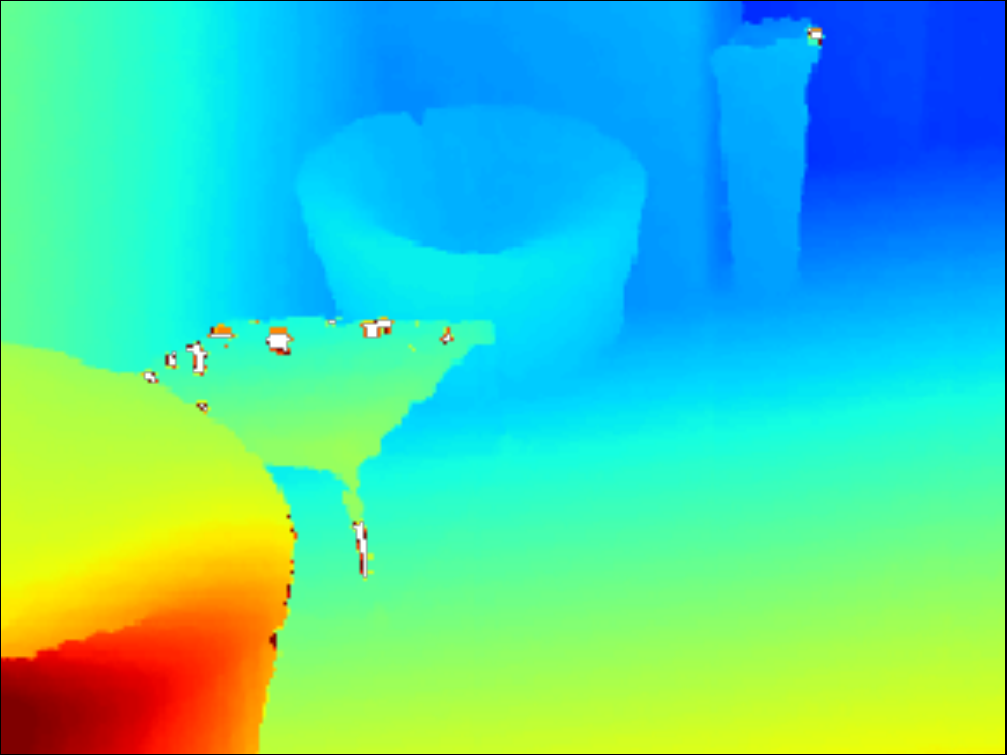}\\
		\includegraphics[width=1.0\textwidth]{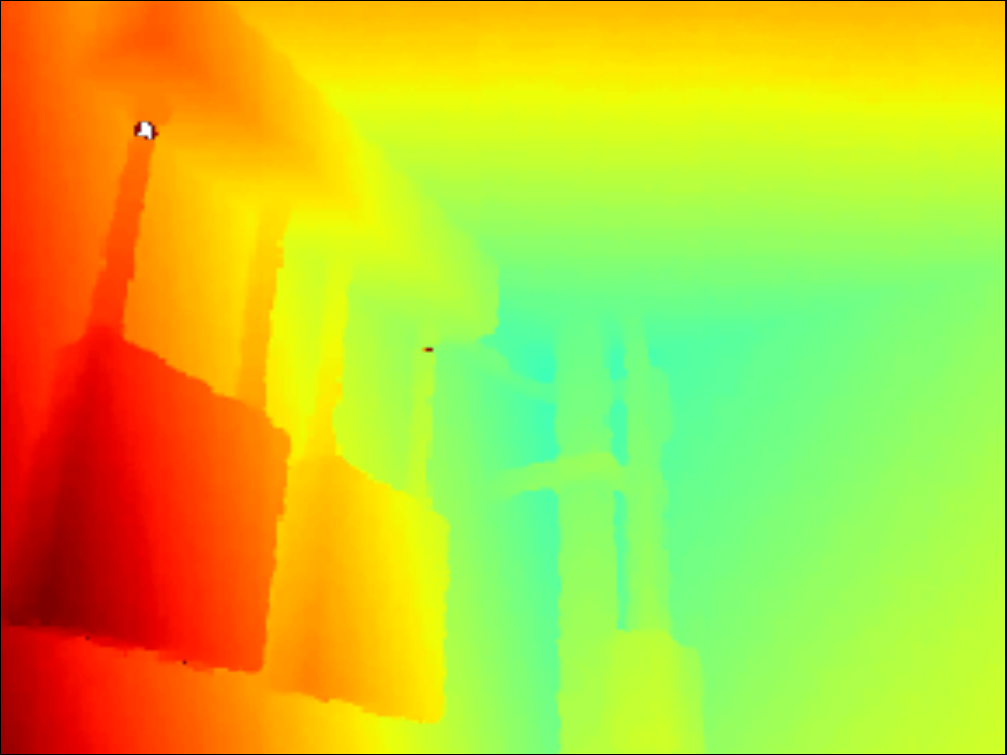}\\
		\includegraphics[width=1.0\textwidth]{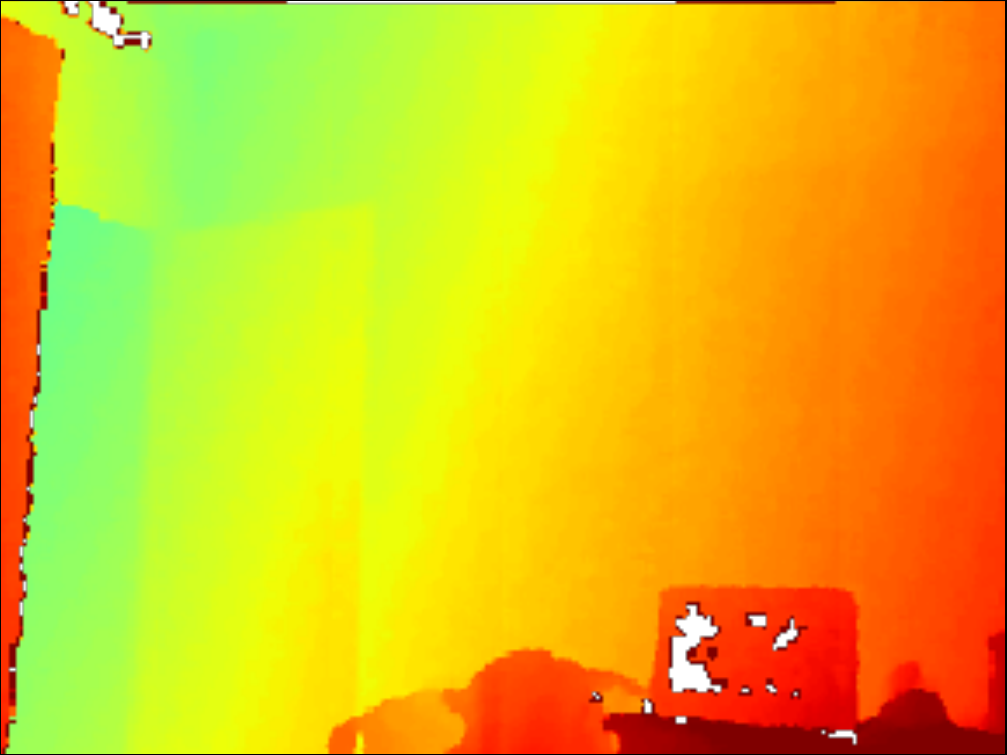}
	\end{tabular}
	}
\end{minipage}
\begin{minipage}{0.14\hsize}
	\centering
	\subfloat[]{
	\begin{tabular}{c}
		\includegraphics[width=1.0\textwidth]{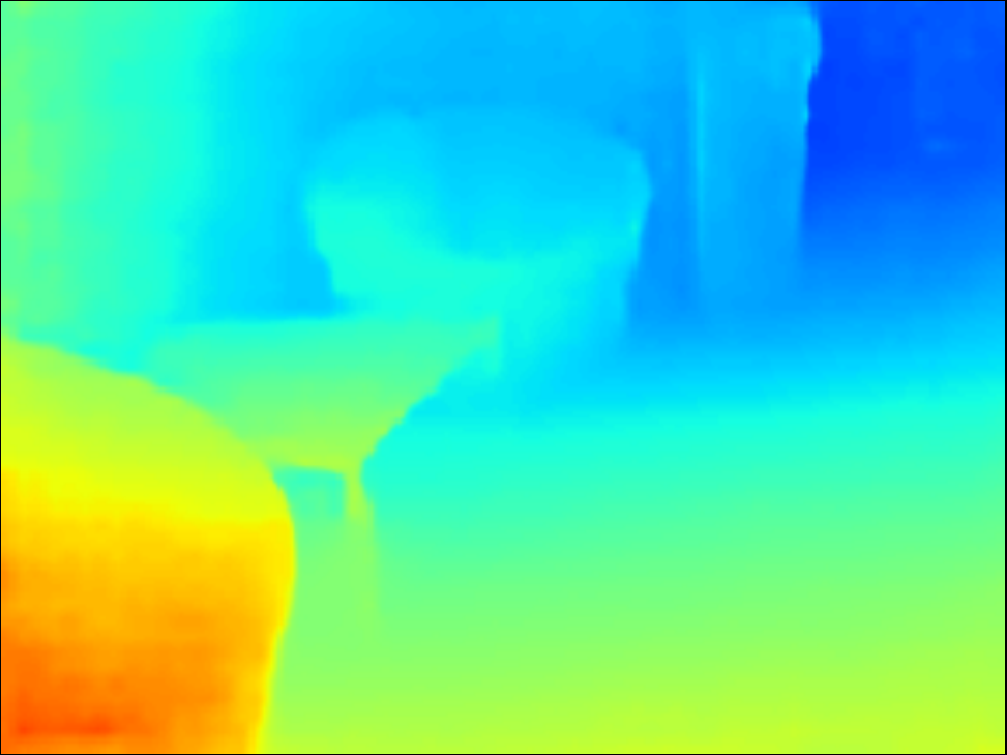}\\
		\includegraphics[width=1.0\textwidth]{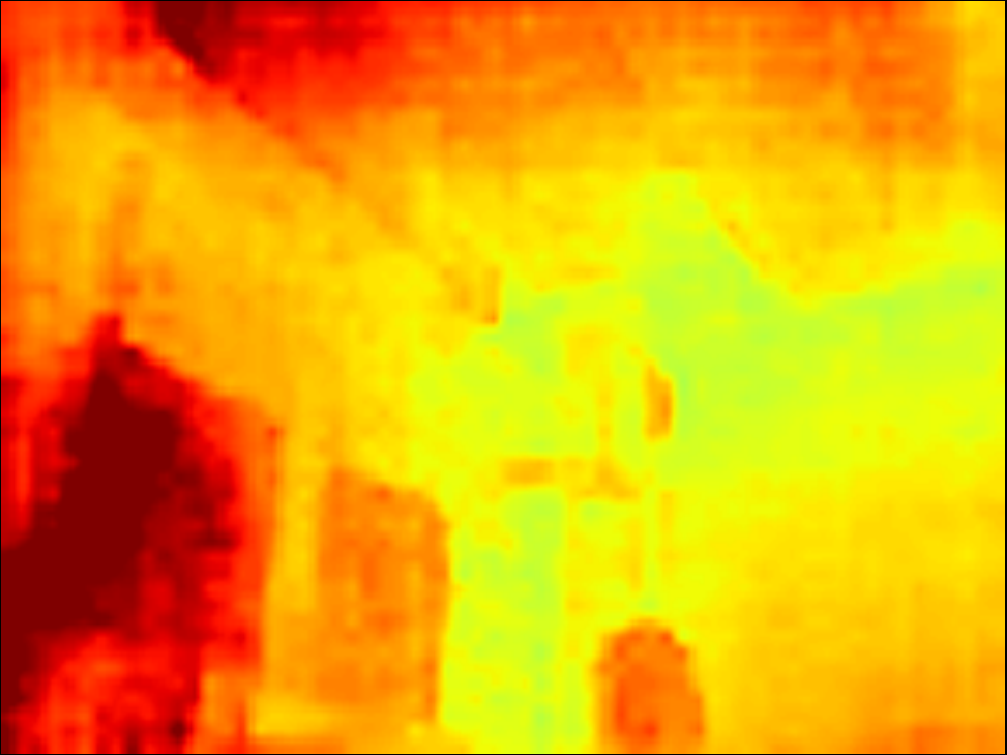}\\
		\includegraphics[width=1.0\textwidth]{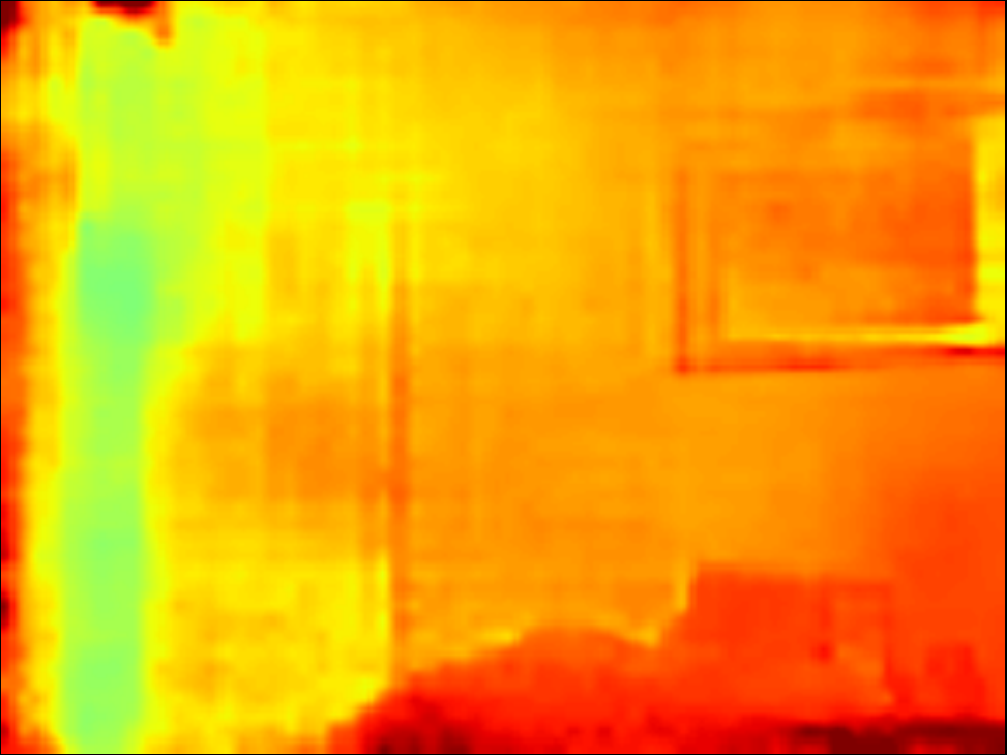}
	\end{tabular}
	}
\end{minipage}
\begin{minipage}{0.14\hsize}
	\centering
	\subfloat[]{
	\begin{tabular}{c}
		\includegraphics[width=1.0\textwidth]{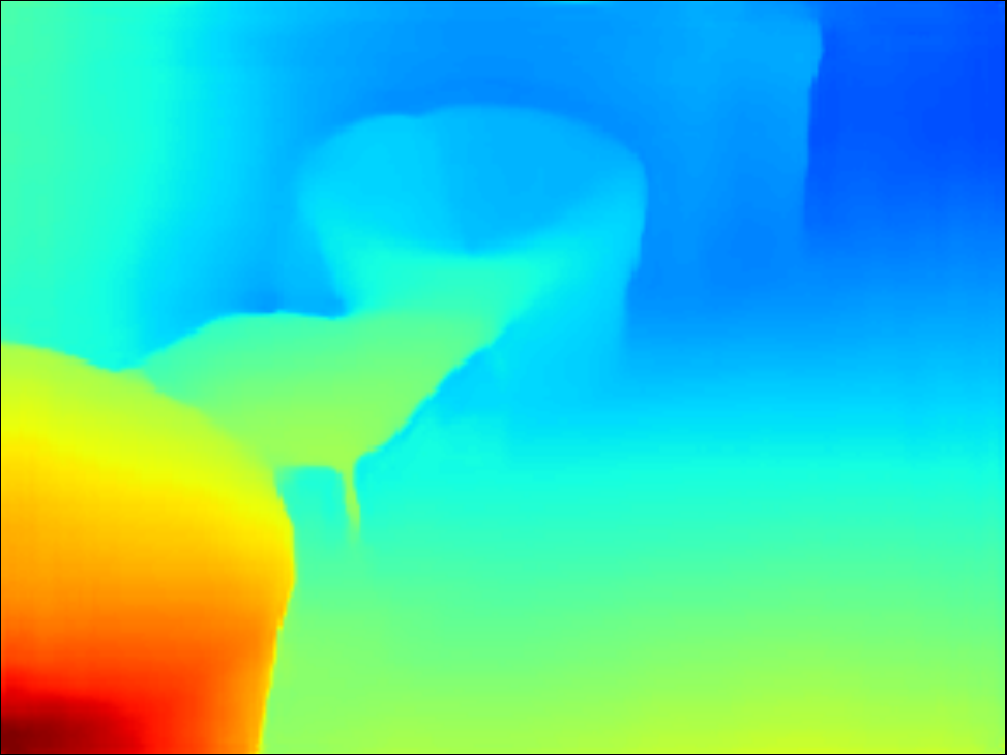}\\
		\includegraphics[width=1.0\textwidth]{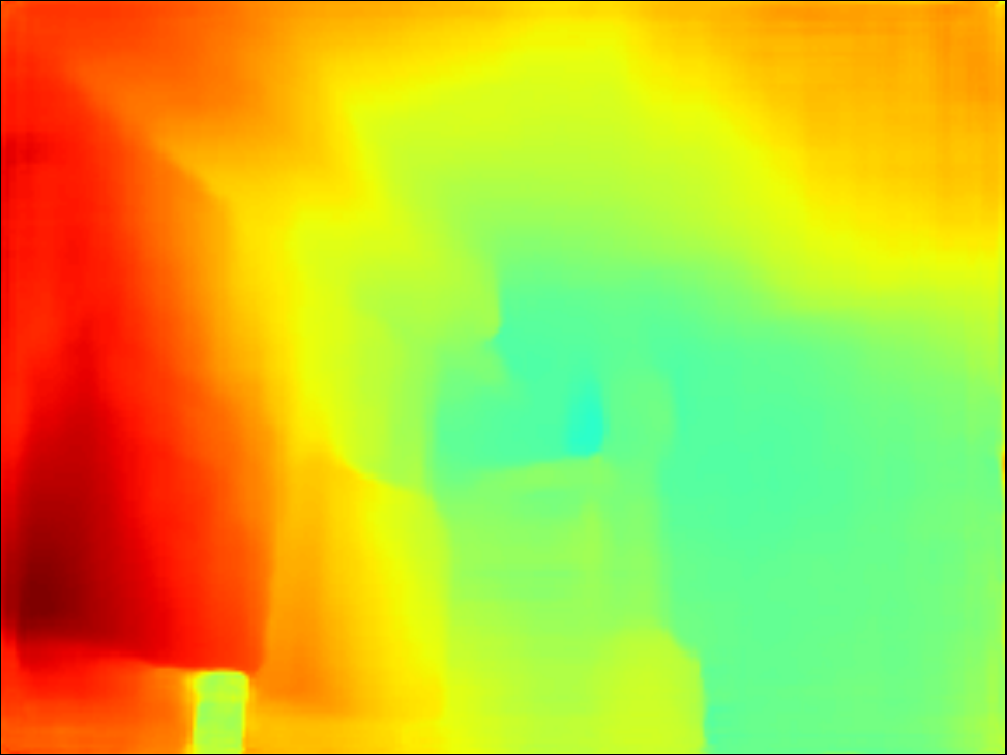}\\
		\includegraphics[width=1.0\textwidth]{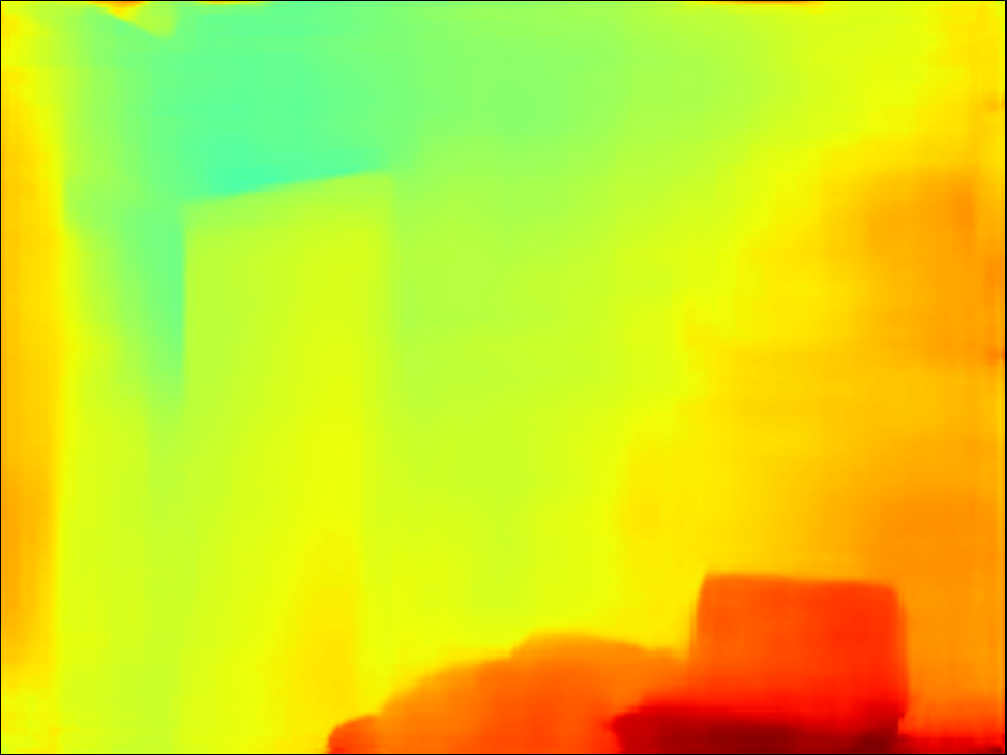}
	\end{tabular}
	}
\end{minipage}
\begin{minipage}{0.14\hsize}
	\centering
	\subfloat[]{
	\begin{tabular}{c}
		\includegraphics[width=1.0\textwidth]{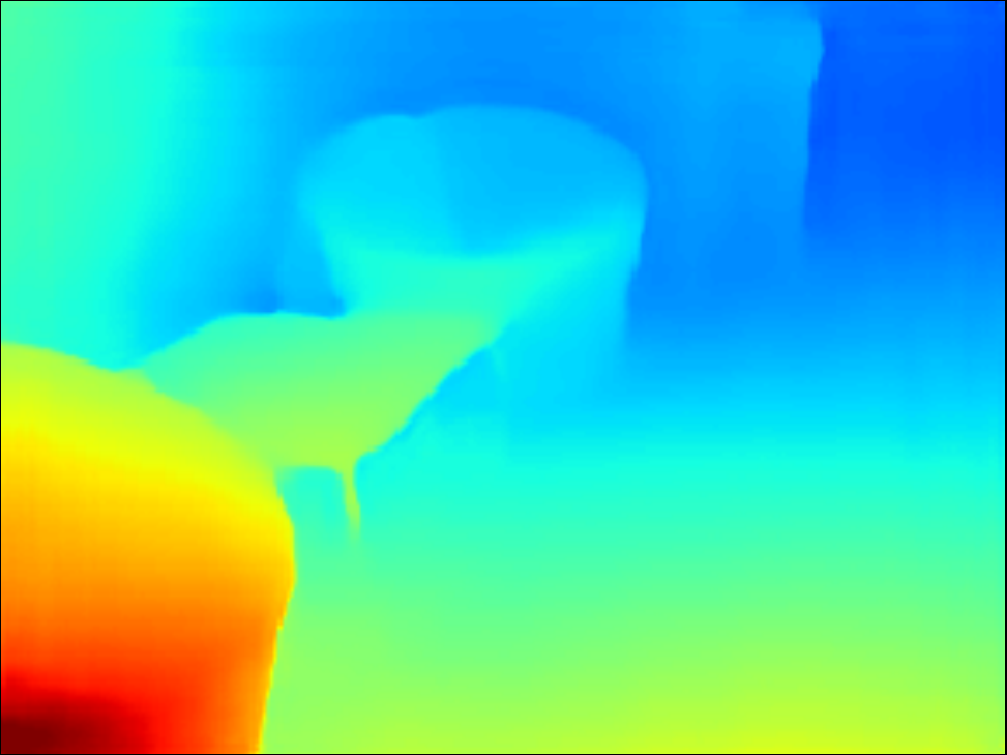}\\
		\includegraphics[width=1.0\textwidth]{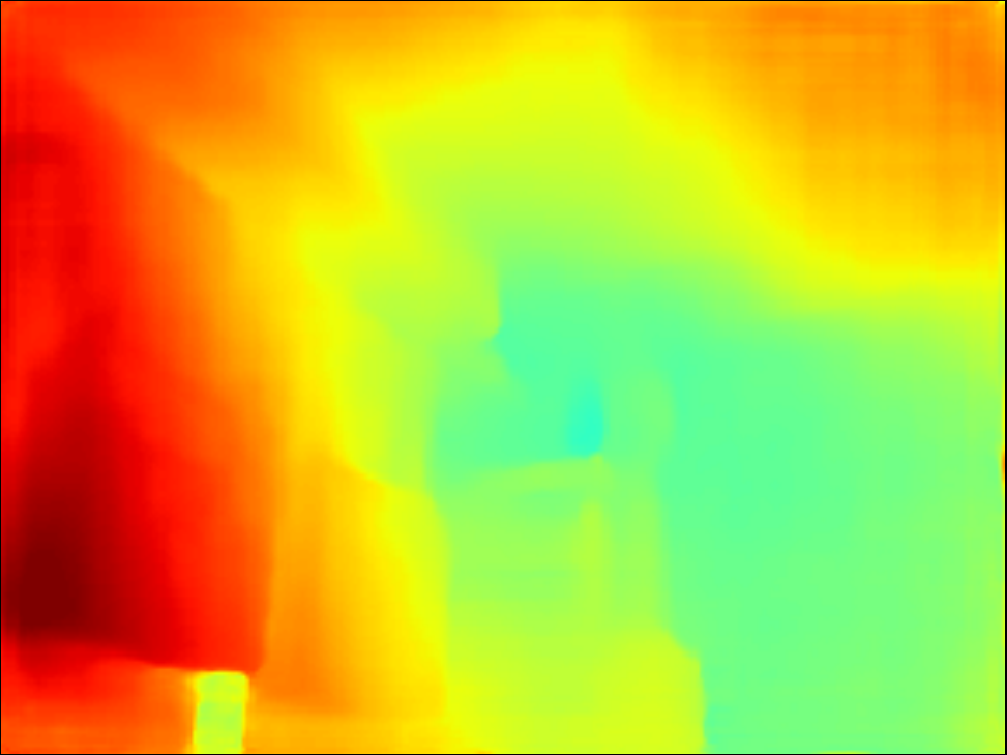}\\
		\includegraphics[width=1.0\textwidth]{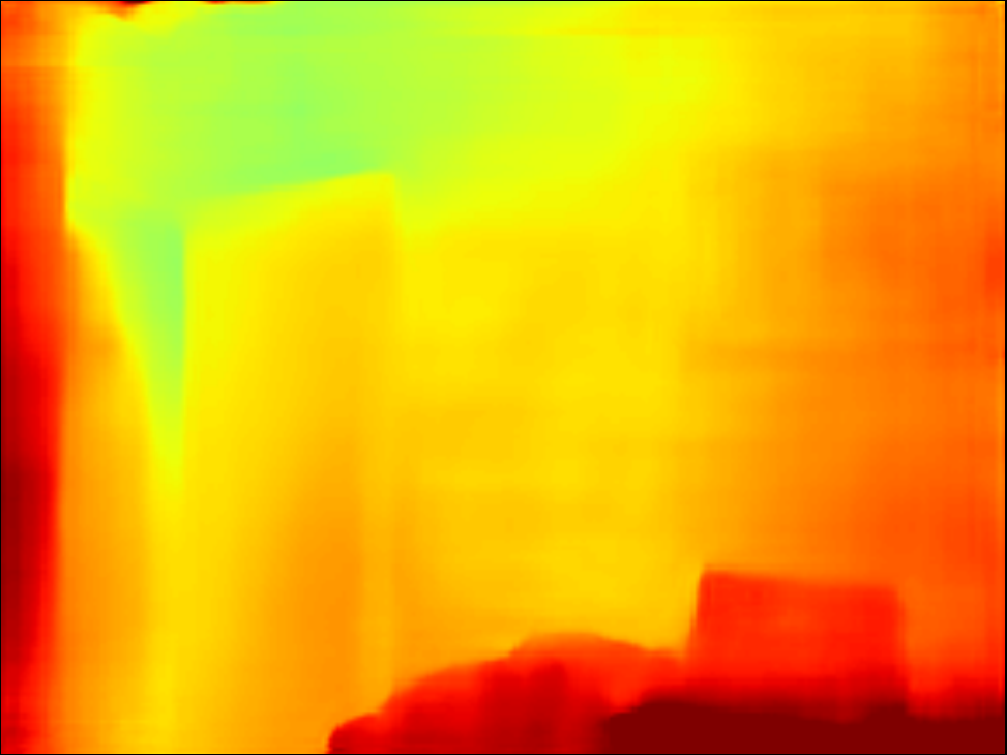}
	\end{tabular}
	}
\end{minipage}
\begin{minipage}{0.14\hsize}
	\centering
	\subfloat[]{
	\begin{tabular}{c}
		\includegraphics[width=1.0\textwidth]{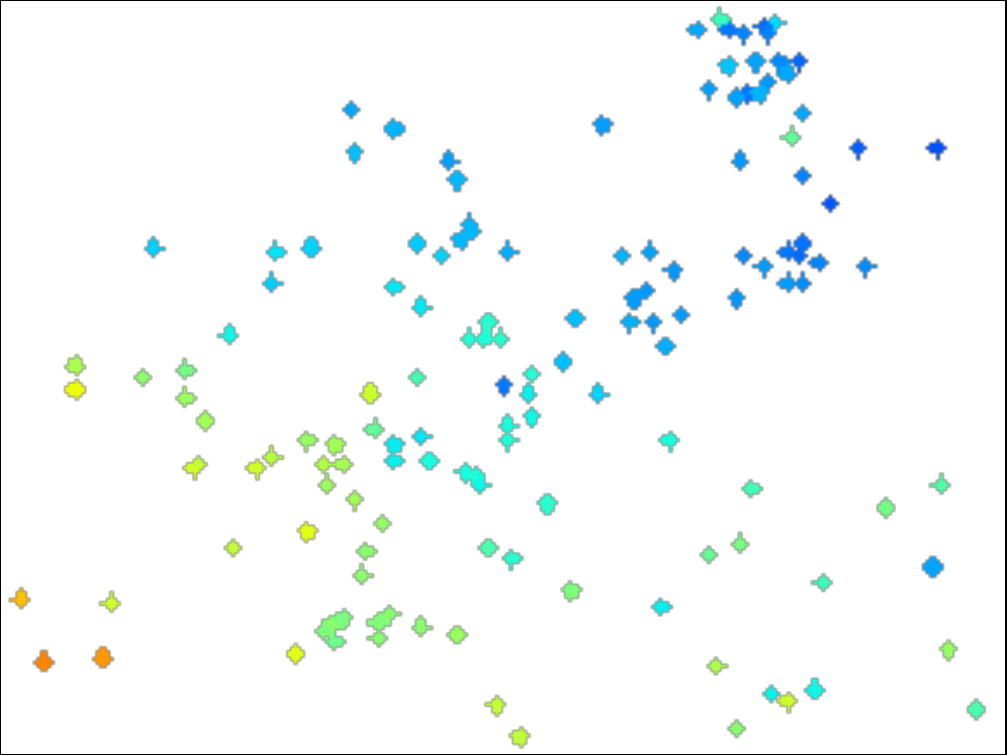}\\
		\includegraphics[width=1.0\textwidth]{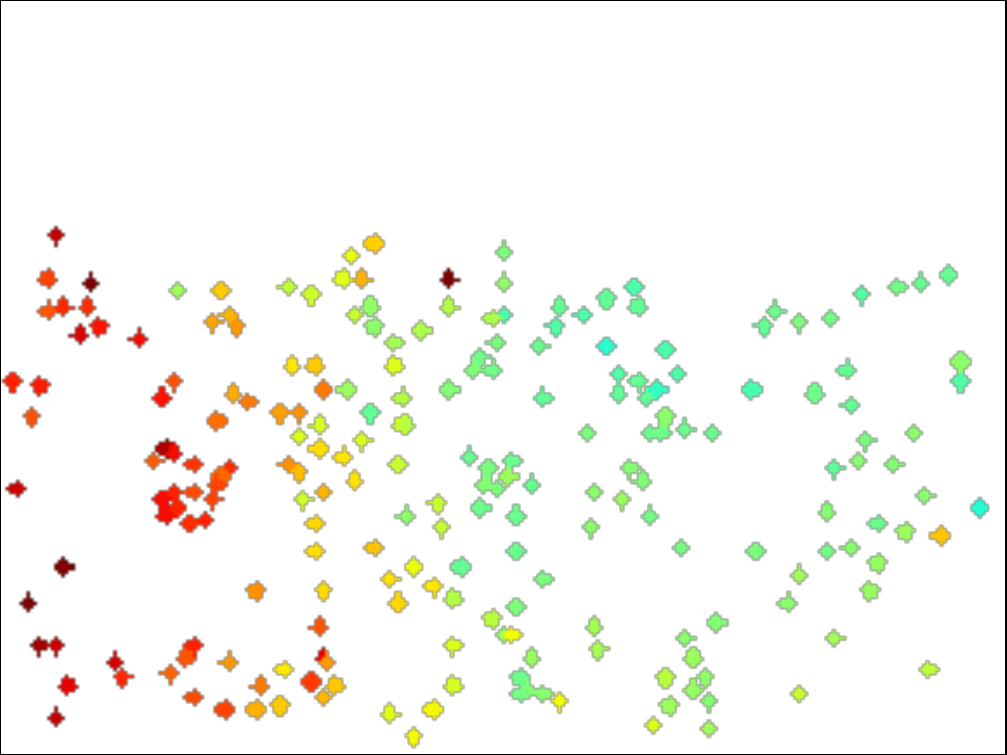}\\
		\includegraphics[width=1.0\textwidth]{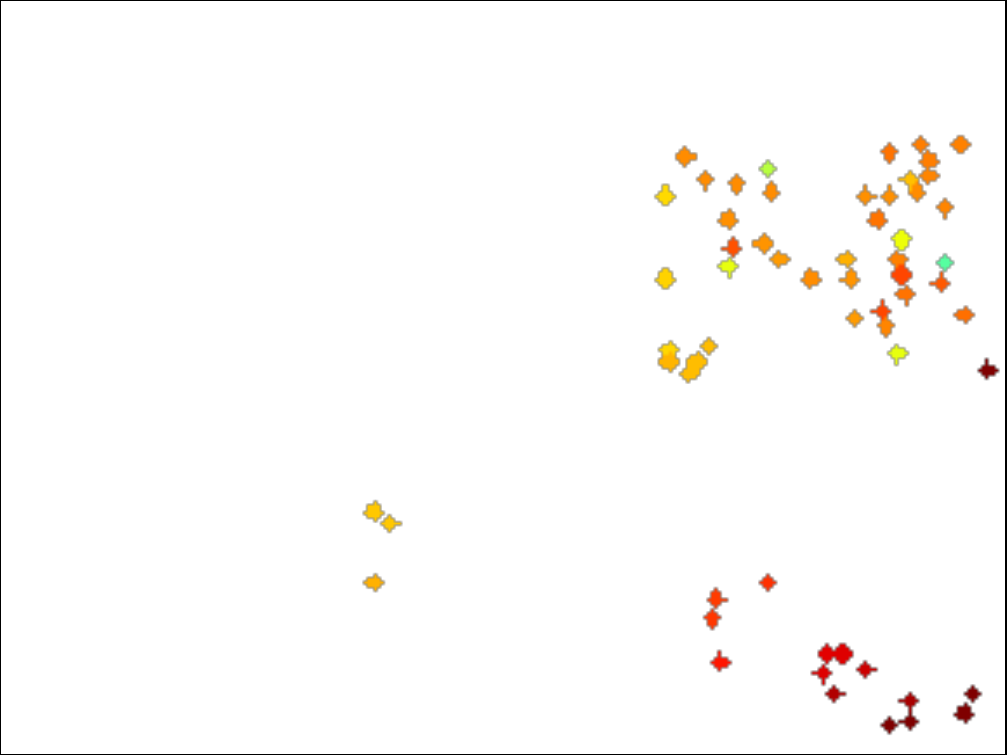}
	\end{tabular}
	}
\end{minipage}
\caption{Output depth after scattering parameter estimation. (a) Hazy input, (b) ground-truth depth, (c) DPSNet \cite{im19}, (d) proposed method with ground-truth scattering parameters, (e) proposed method with scattering parameter estimation, and (f) sparse depth obtained by SfM.}
\label{fig:pe_qualitative}
\end{figure*}

\begin{figure*}[tb]
\centering
\begin{minipage}{0.14\hsize}
	\centering
	\subfloat[]{
	\begin{tabular}{c}
		\includegraphics[width=1.0\textwidth]{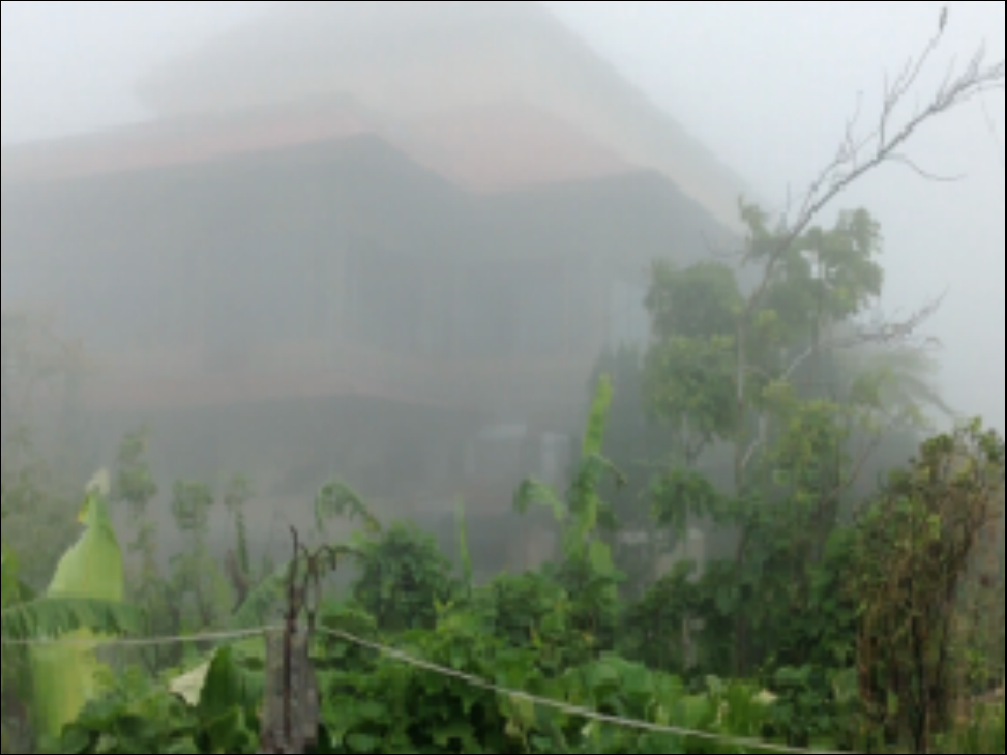}\\
		\includegraphics[width=1.0\textwidth]{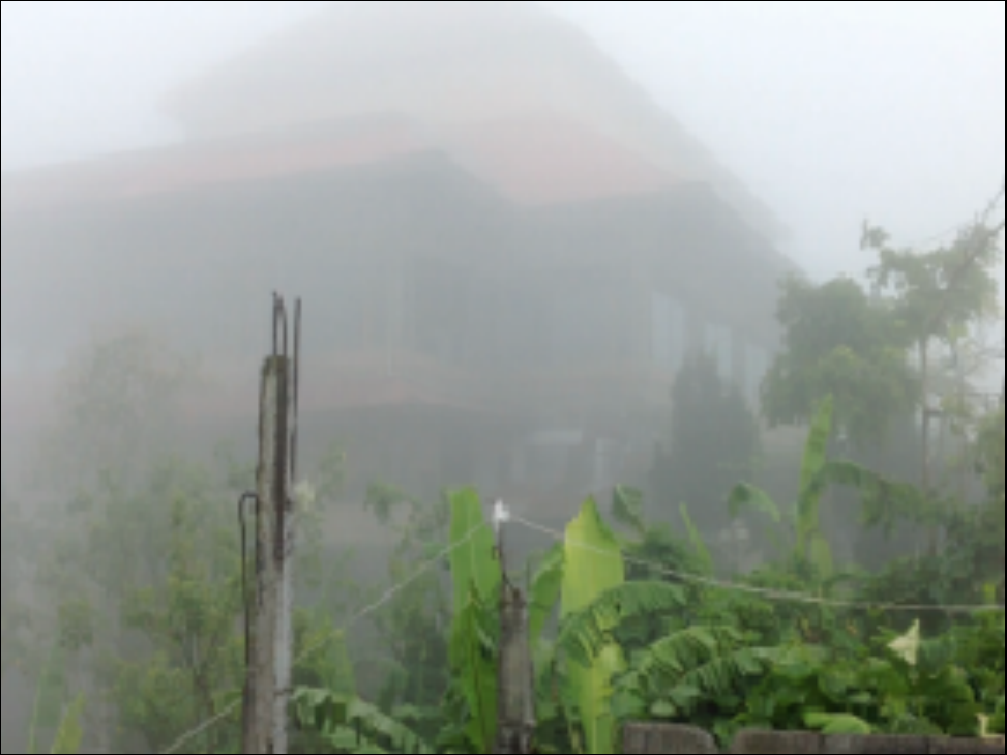}
	\end{tabular}
	}
\end{minipage}
\begin{minipage}{0.14\hsize}
	\centering
	\subfloat[]{
	\begin{tabular}{c}
		\includegraphics[width=1.0\textwidth]{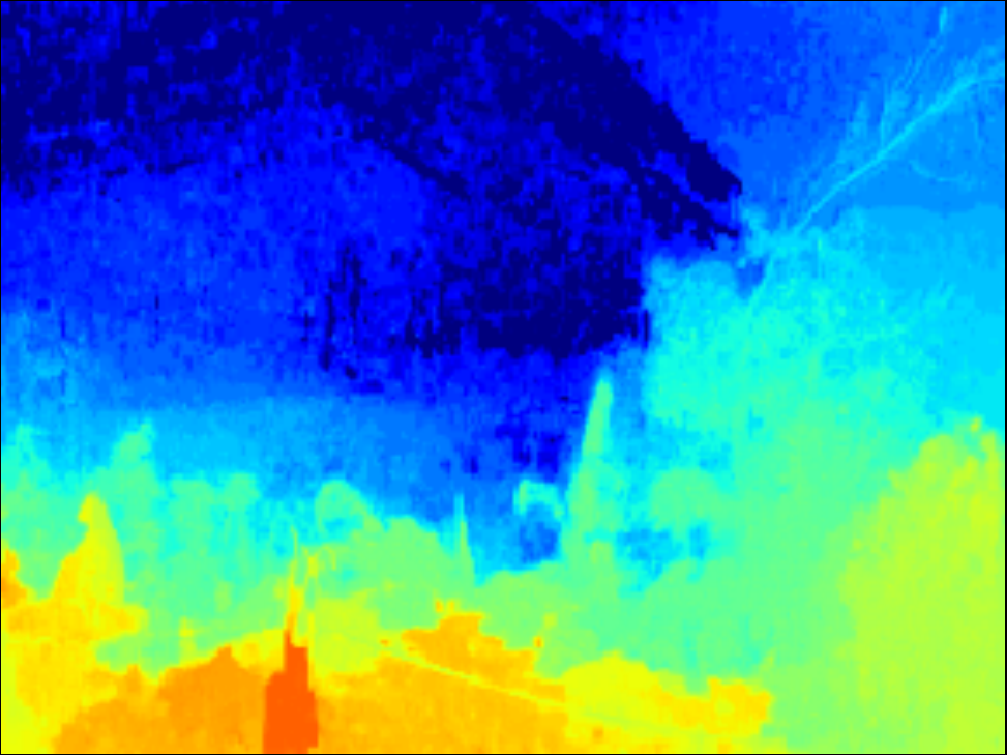}\\
		\includegraphics[width=1.0\textwidth]{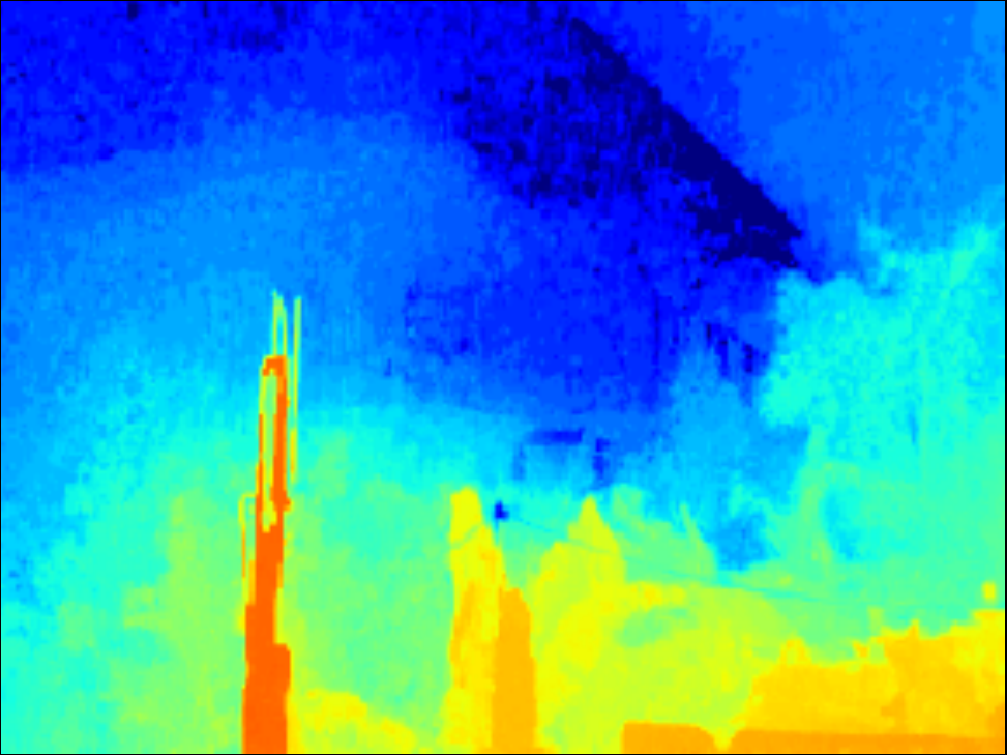}
	\end{tabular}
	}
\end{minipage}
\begin{minipage}{0.14\hsize}
	\centering
	\subfloat[]{
	\begin{tabular}{c}
		\includegraphics[width=1.0\textwidth]{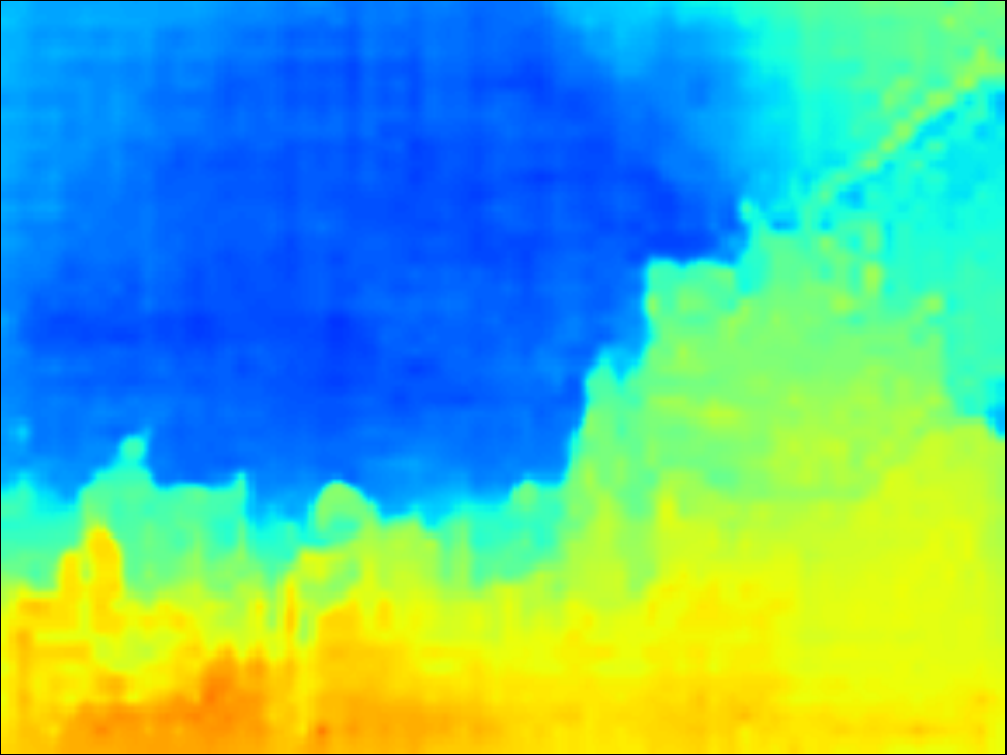}\\
		\includegraphics[width=1.0\textwidth]{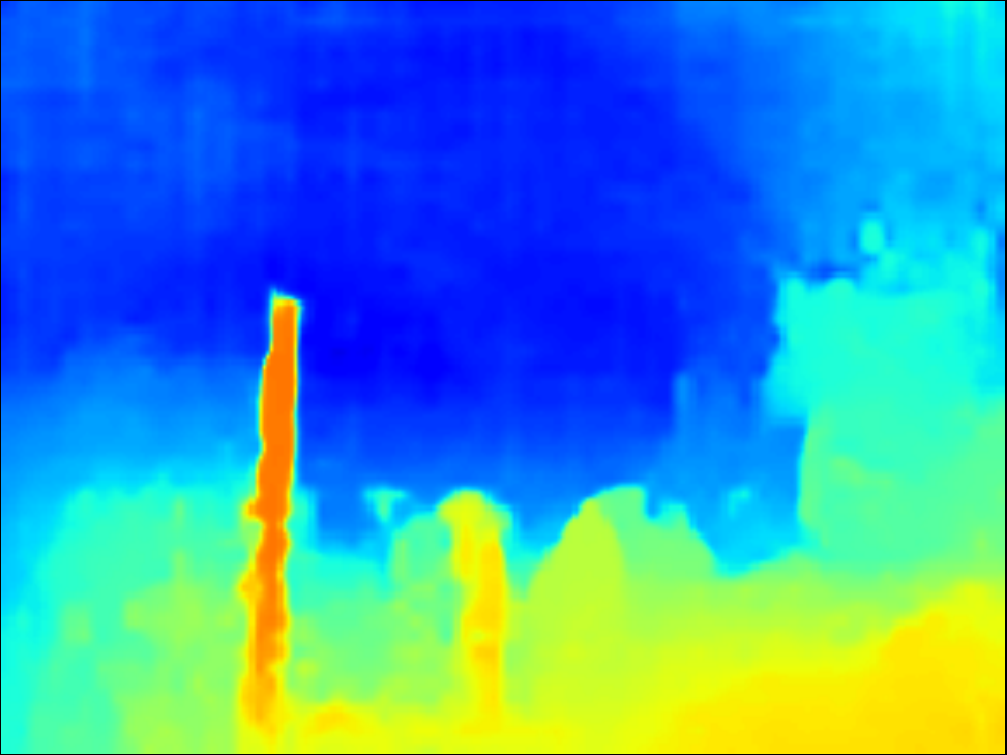}
	\end{tabular}
	}
\end{minipage}
\begin{minipage}{0.14\hsize}
	\centering
	\subfloat[]{
	\begin{tabular}{c}
		\includegraphics[width=1.0\textwidth]{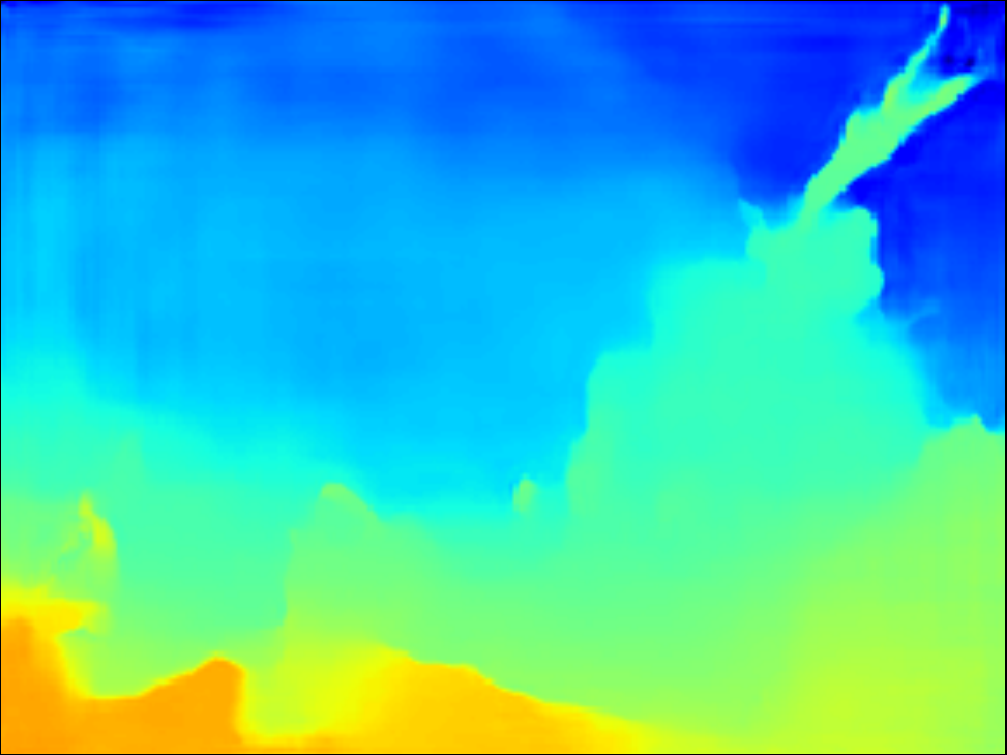}\\
		\includegraphics[width=1.0\textwidth]{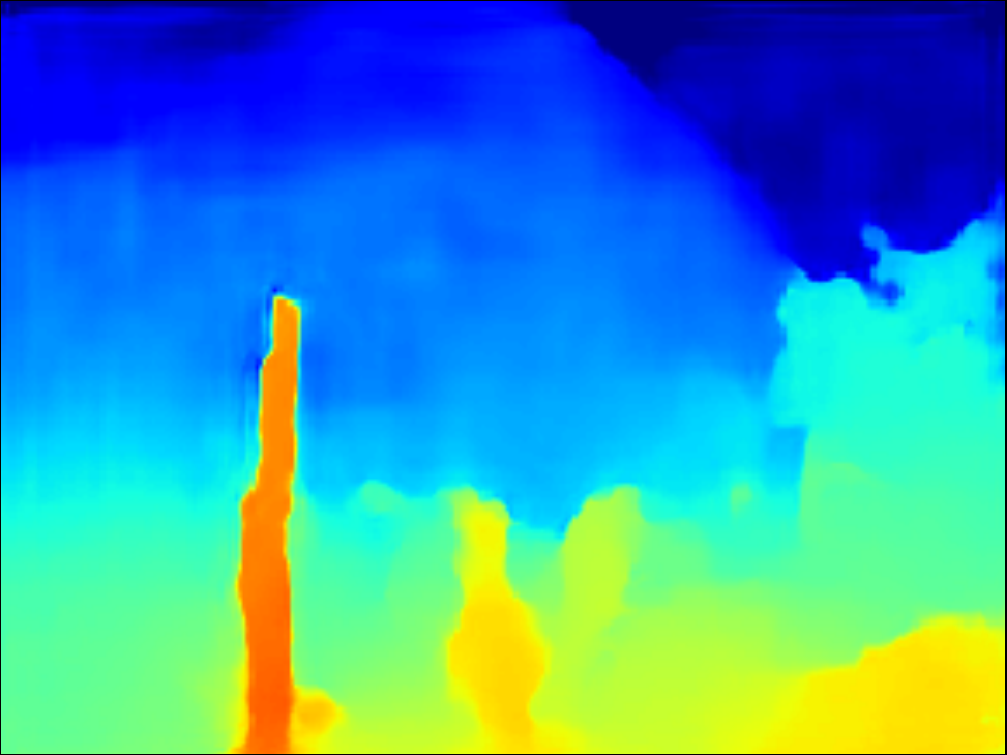}
	\end{tabular}
	}
\end{minipage}
\begin{minipage}{0.14\hsize}
	\centering
	\subfloat[]{
	\begin{tabular}{c}
		\includegraphics[width=1.0\textwidth]{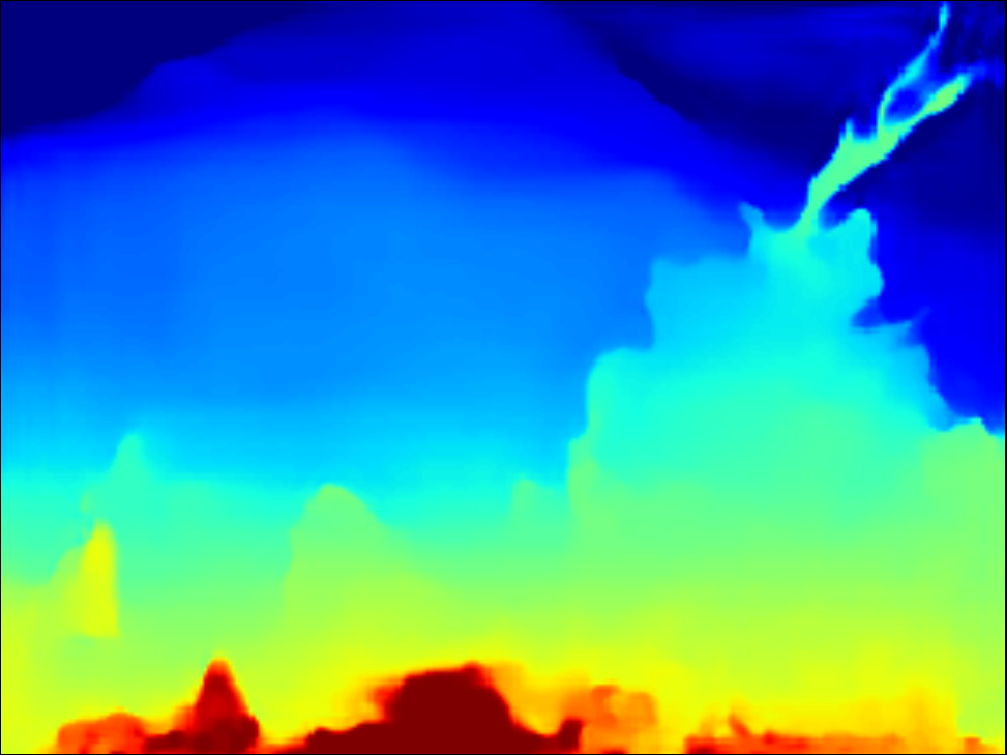}\\
		\includegraphics[width=1.0\textwidth]{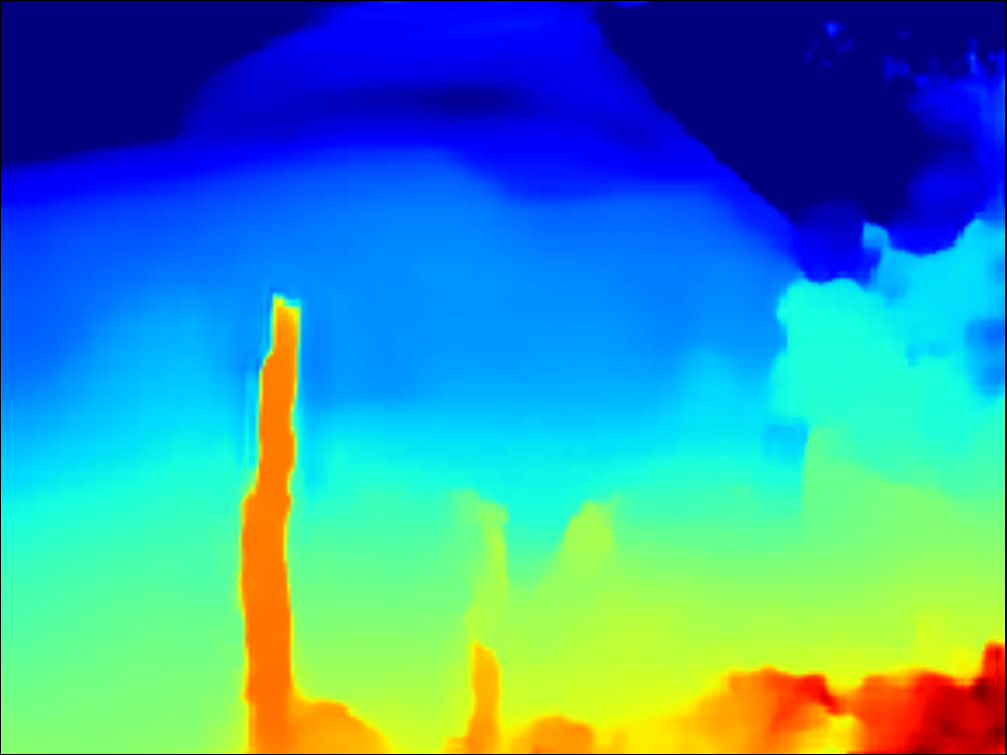}
	\end{tabular}
	}
\end{minipage}
\begin{minipage}{0.14\hsize}
	\centering
	\subfloat[]{
	\begin{tabular}{c}
		\includegraphics[width=1.0\textwidth]{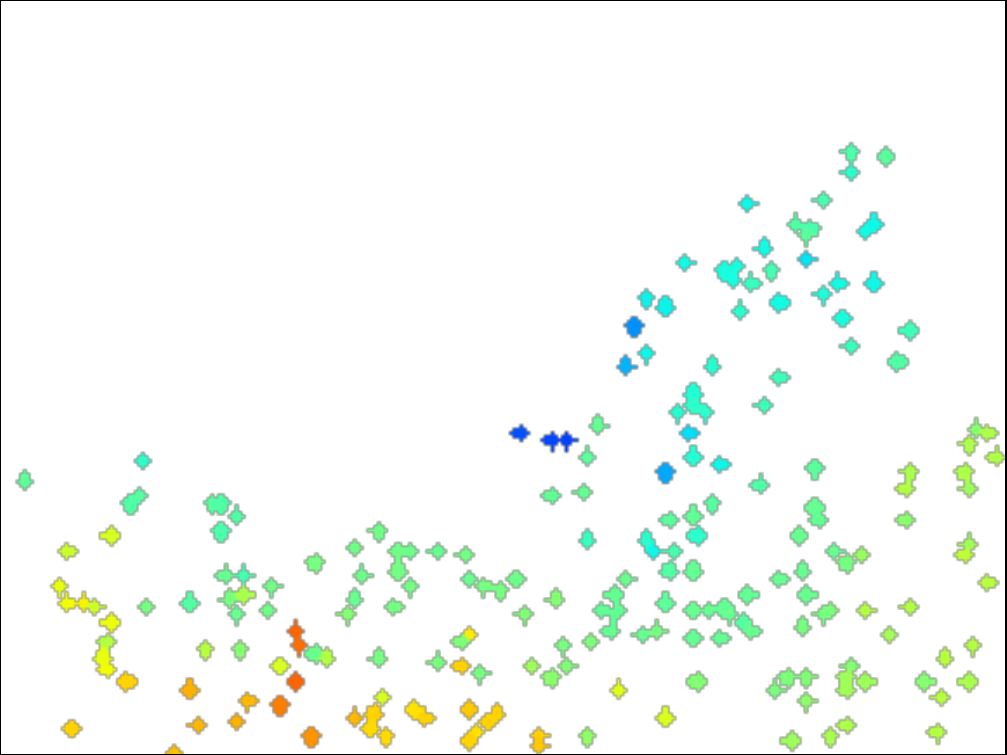}\\
		\includegraphics[width=1.0\textwidth]{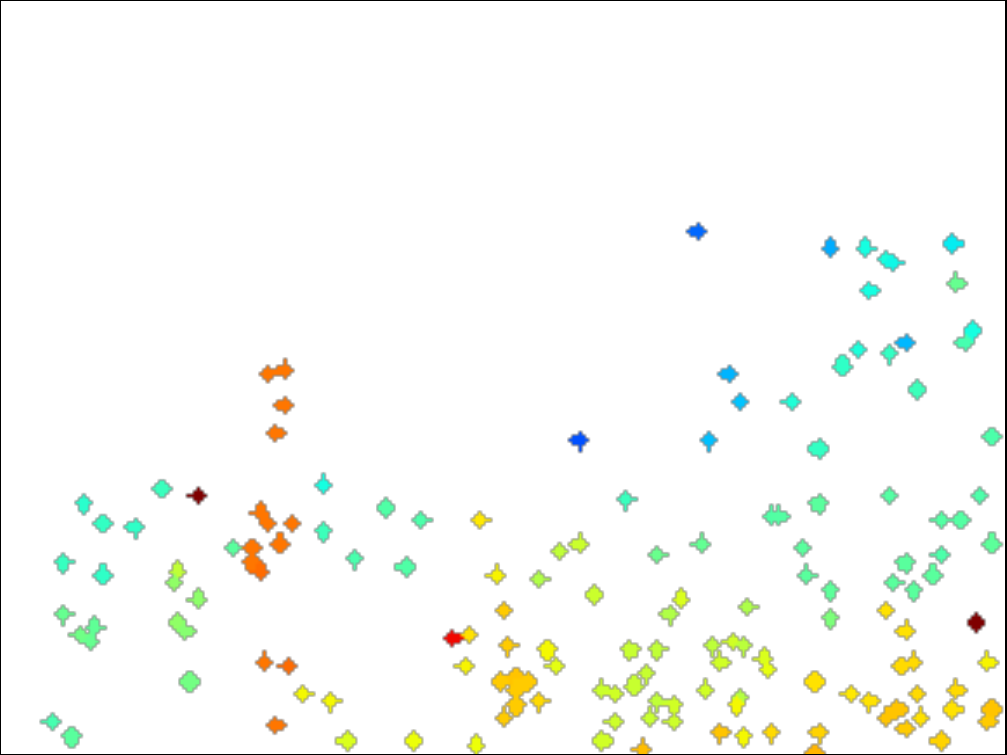}
	\end{tabular}
	}
\end{minipage}
\caption{Experimental results on {\it bali} \cite{li15}. (a) foggy input, (b) estimated depth of Li et al. \cite{li15}, (c) output of DPSNet \cite{im19}, (d) output of fine-tuned MVDepthNet \cite{wang18}, (e) output of proposed method with scattering parameter estimation, and (f) sparse depth obtained by SfM.}
\label{fig:fstereo_result}
\end{figure*}

\begin{figure*}[tb]
\centering
\begin{minipage}{0.14\hsize}
	\centering
	\subfloat[]{
	\begin{tabular}{c}
		\includegraphics[width=1.0\textwidth]{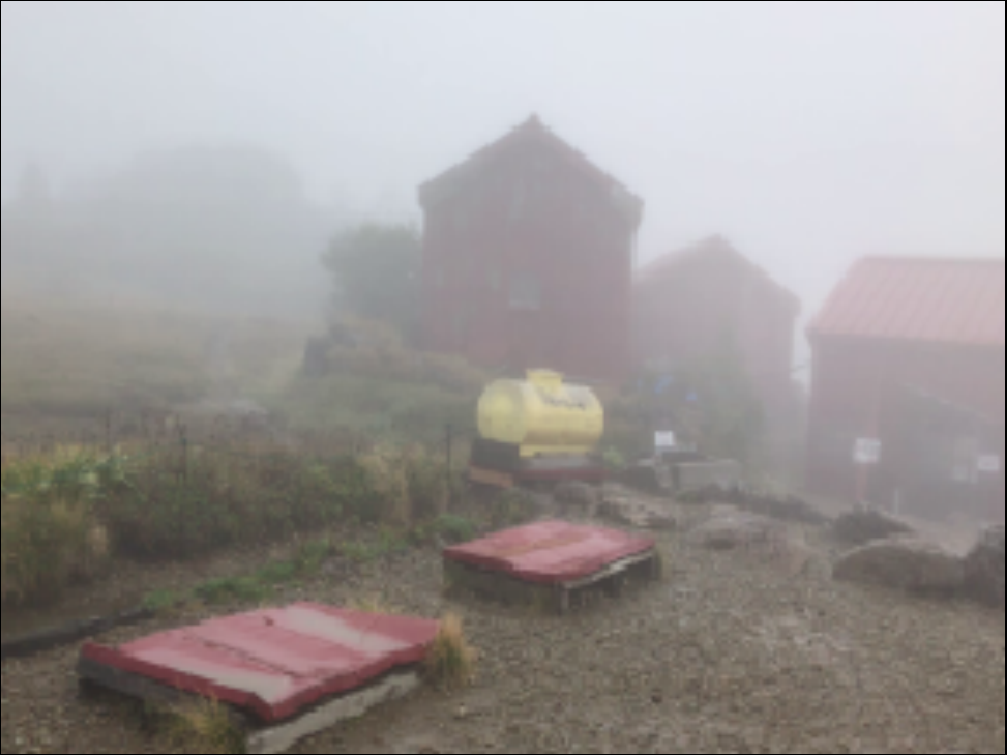}\\
		\includegraphics[width=1.0\textwidth]{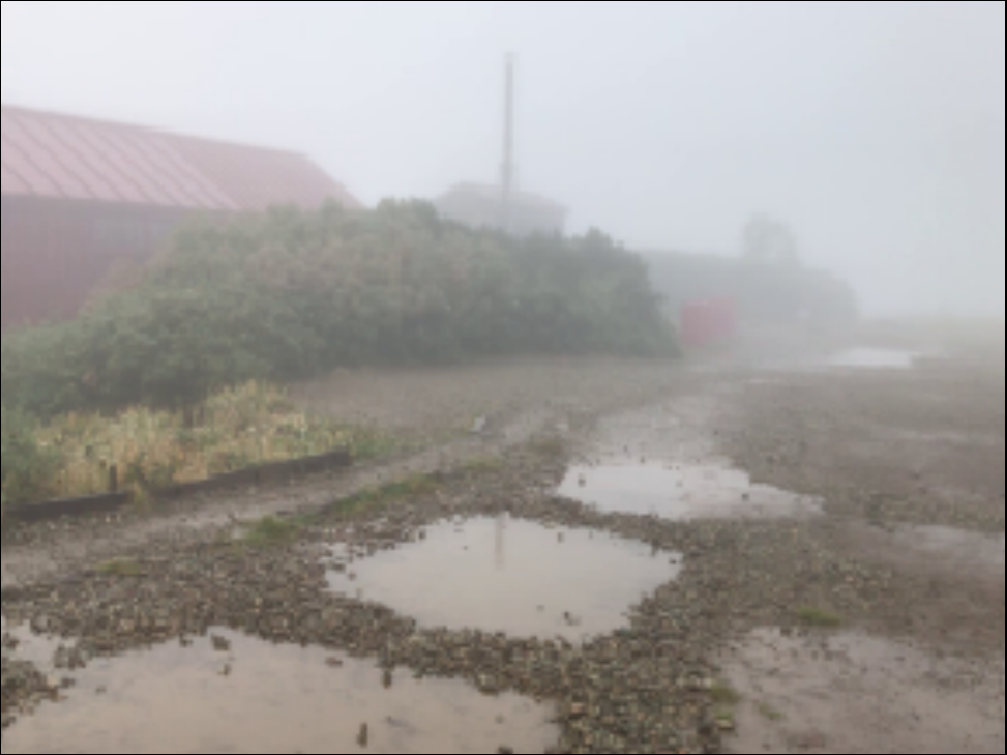}\\
		\includegraphics[width=1.0\textwidth]{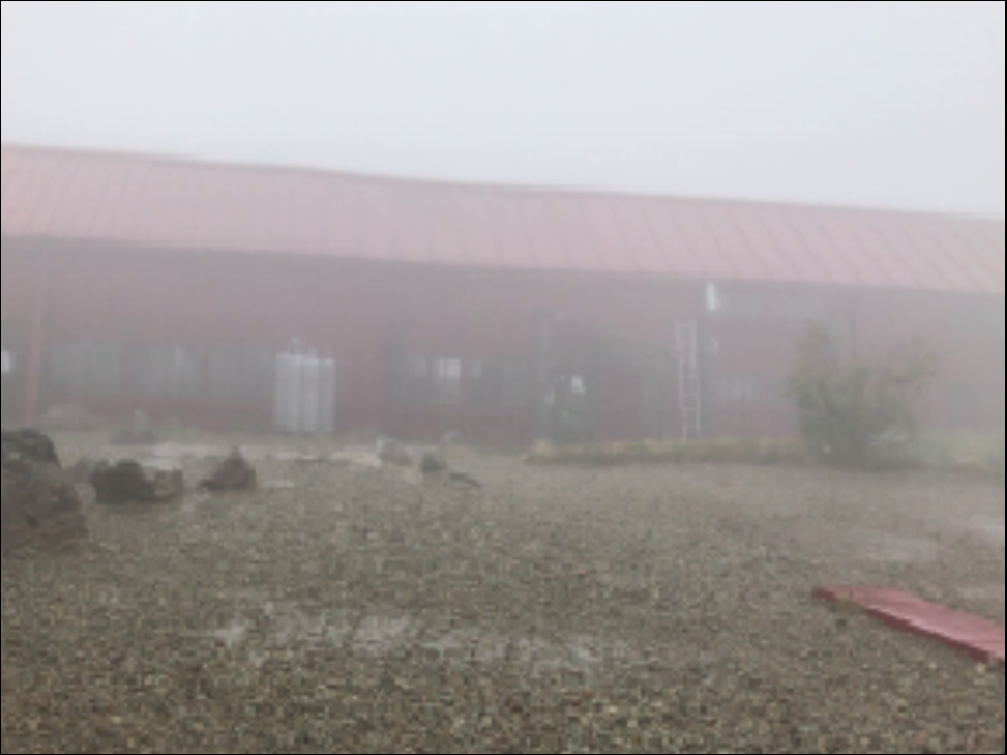}\\
		\includegraphics[width=1.0\textwidth]{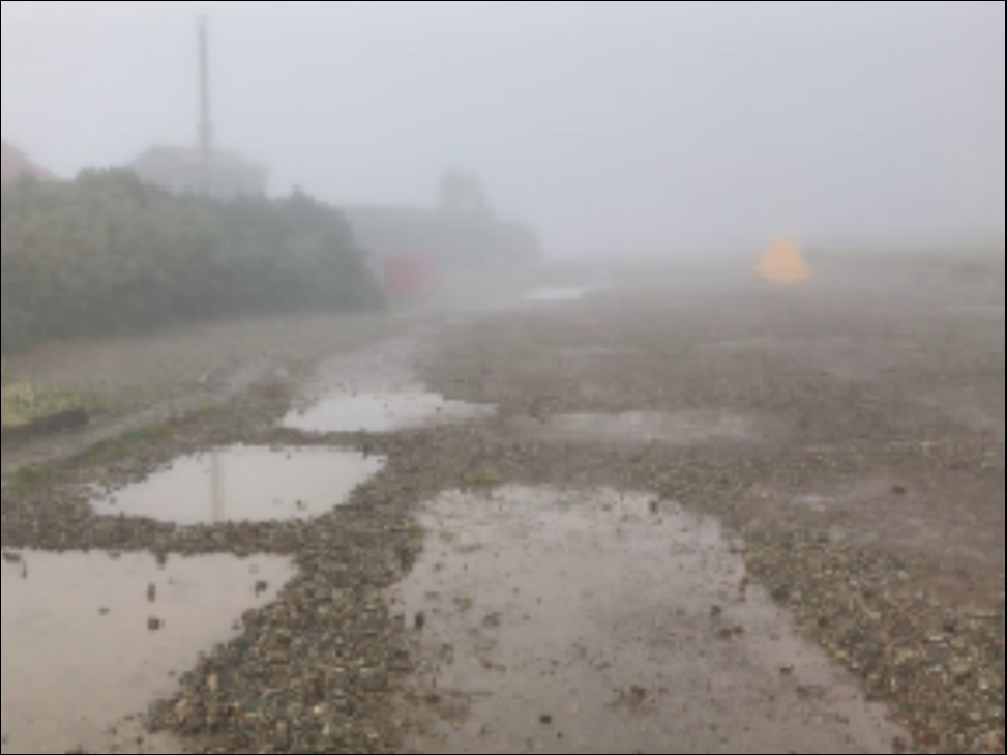}
	\end{tabular}
	}
\end{minipage}
\begin{minipage}{0.14\hsize}
	\centering
	\subfloat[]{
	\begin{tabular}{c}
		\includegraphics[width=1.0\textwidth]{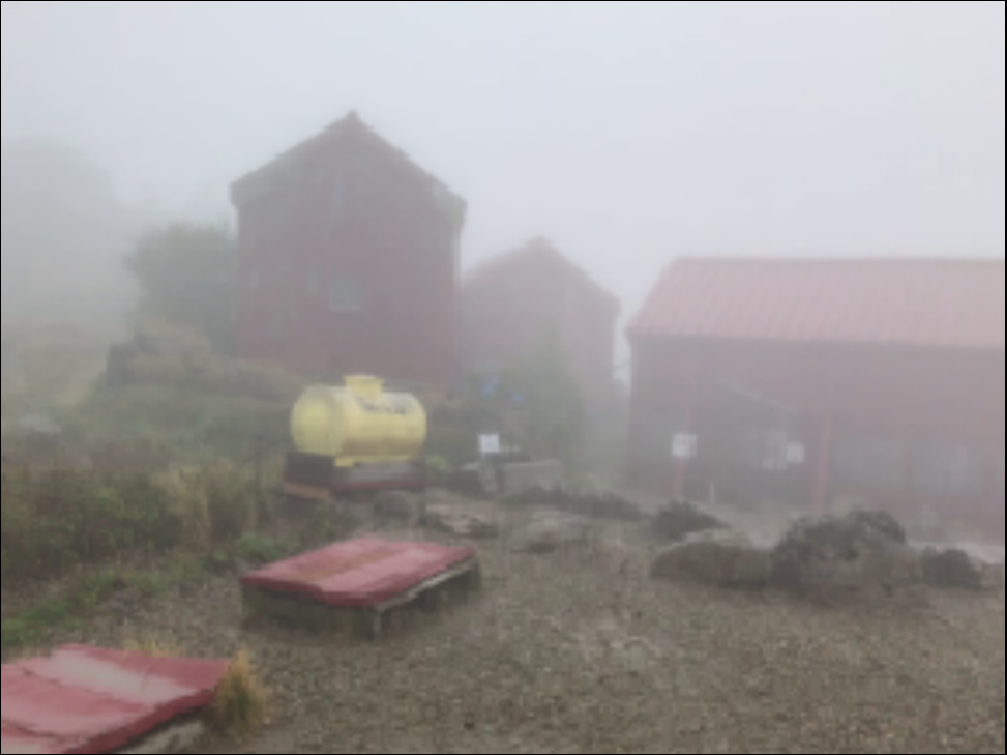}\\
		\includegraphics[width=1.0\textwidth]{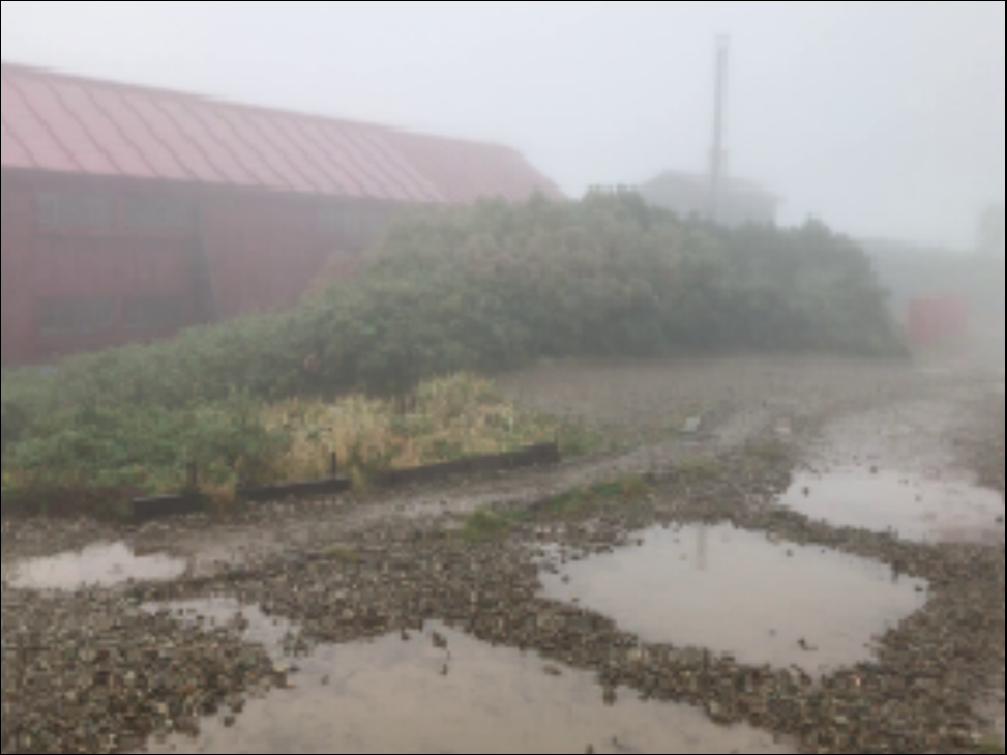}\\
		\includegraphics[width=1.0\textwidth]{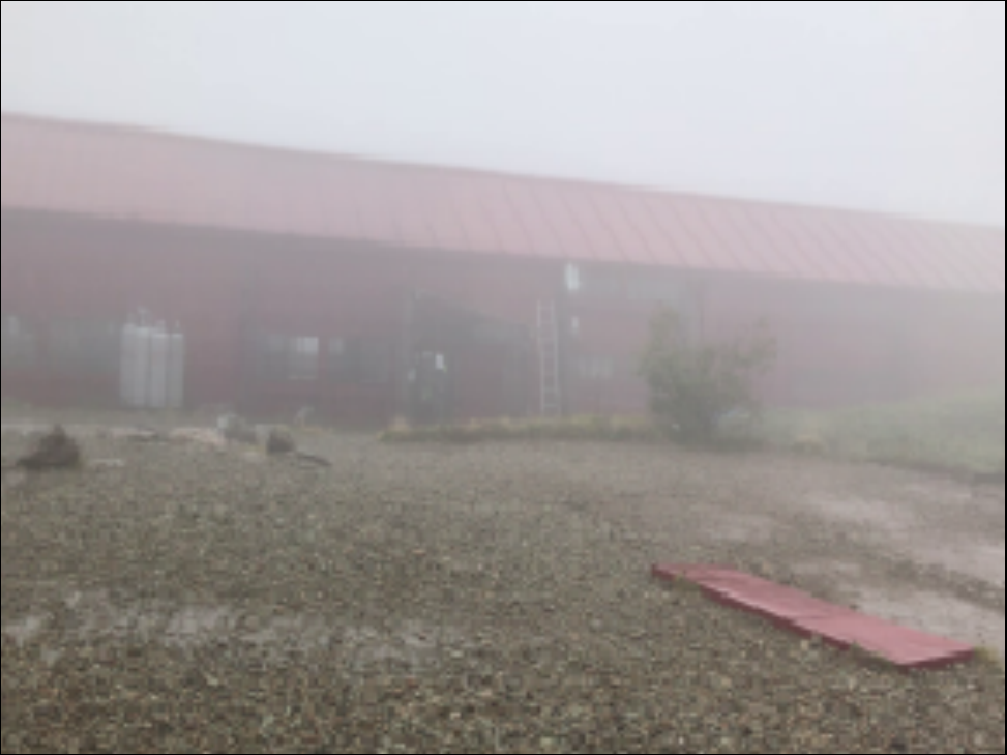}\\
		\includegraphics[width=1.0\textwidth]{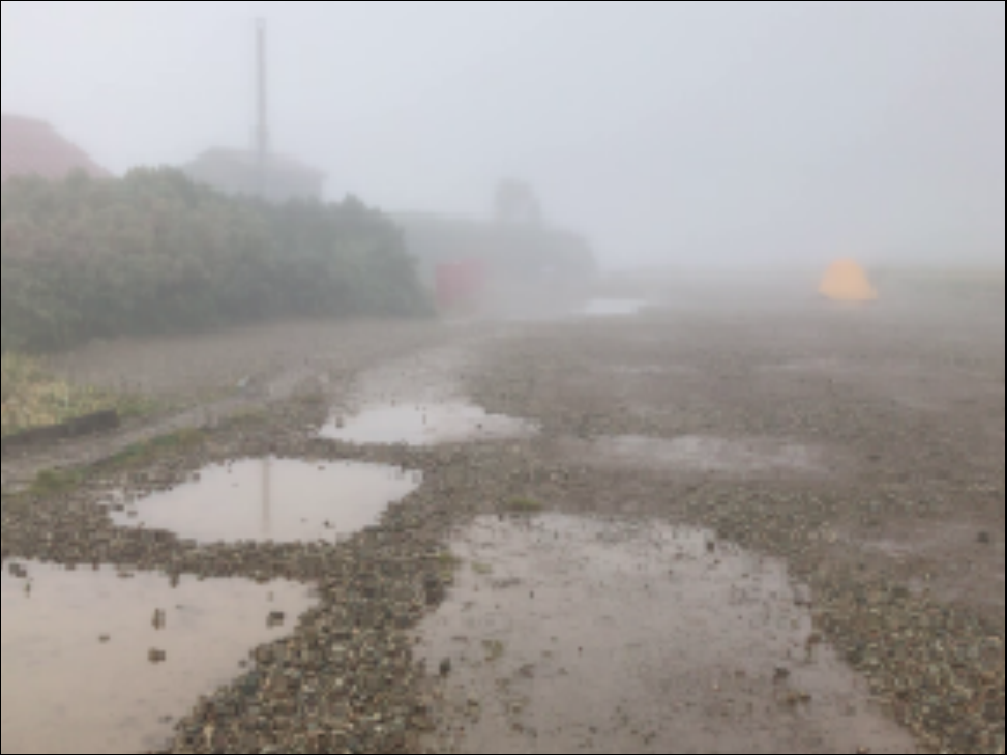}
	\end{tabular}
	}
\end{minipage}
\begin{minipage}{0.14\hsize}
	\centering
	\subfloat[]{
	\begin{tabular}{c}
		\includegraphics[width=1.0\textwidth]{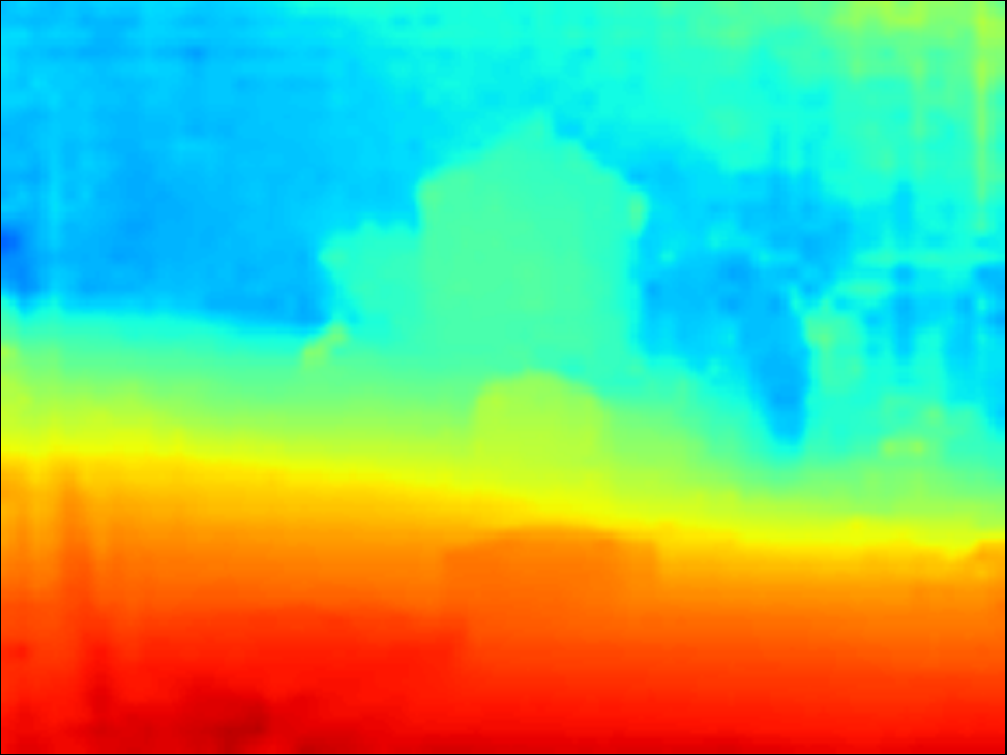}\\
		\includegraphics[width=1.0\textwidth]{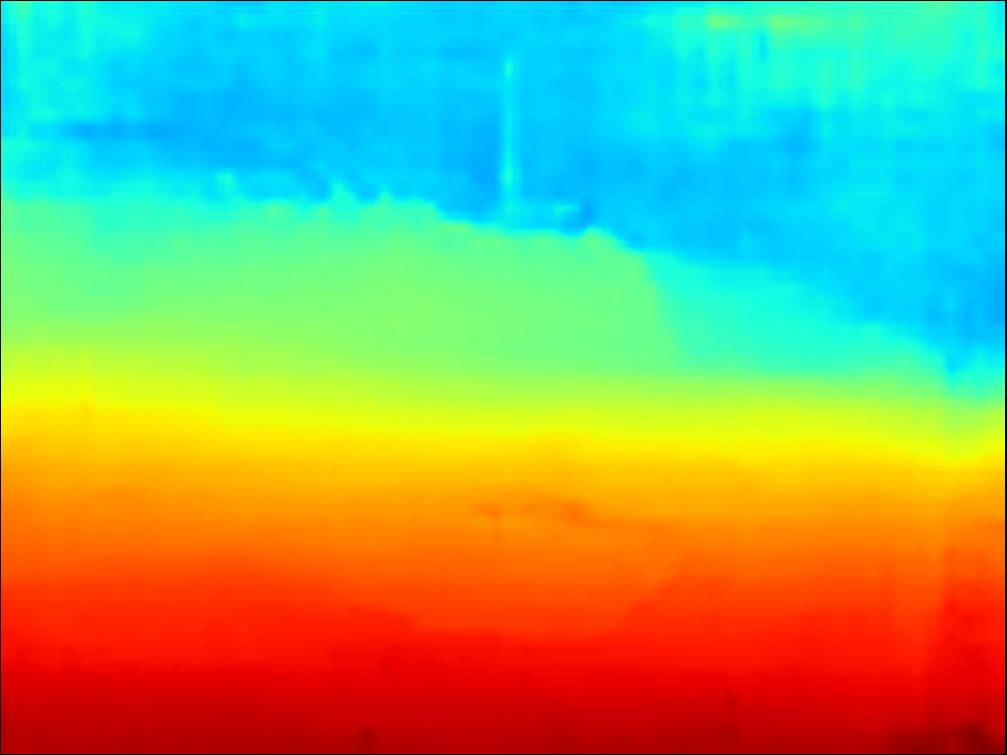}\\
		\includegraphics[width=1.0\textwidth]{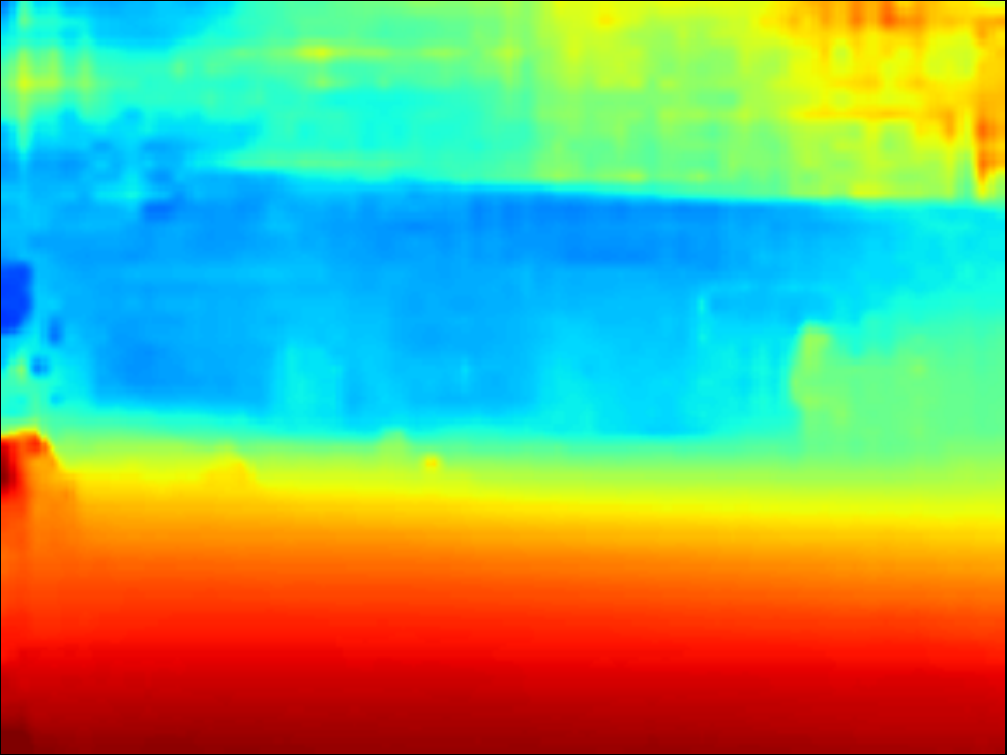}\\
		\includegraphics[width=1.0\textwidth]{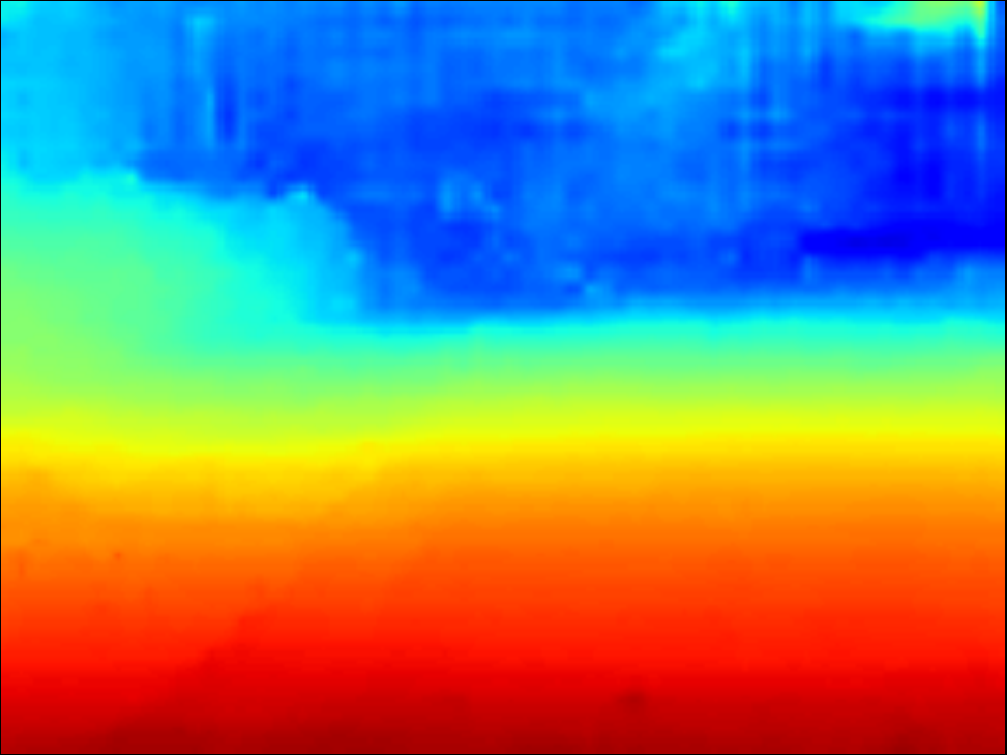}
	\end{tabular}
	}
\end{minipage}
\begin{minipage}{0.14\hsize}
	\centering
	\subfloat[]{
	\begin{tabular}{c}
		\includegraphics[width=1.0\textwidth]{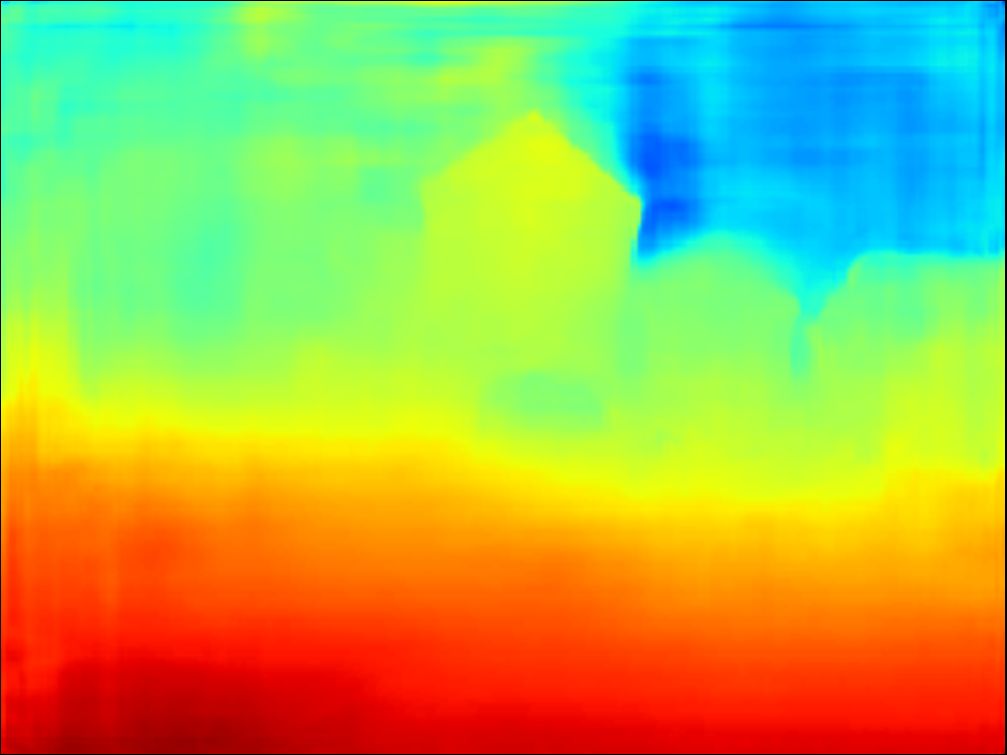}\\
		\includegraphics[width=1.0\textwidth]{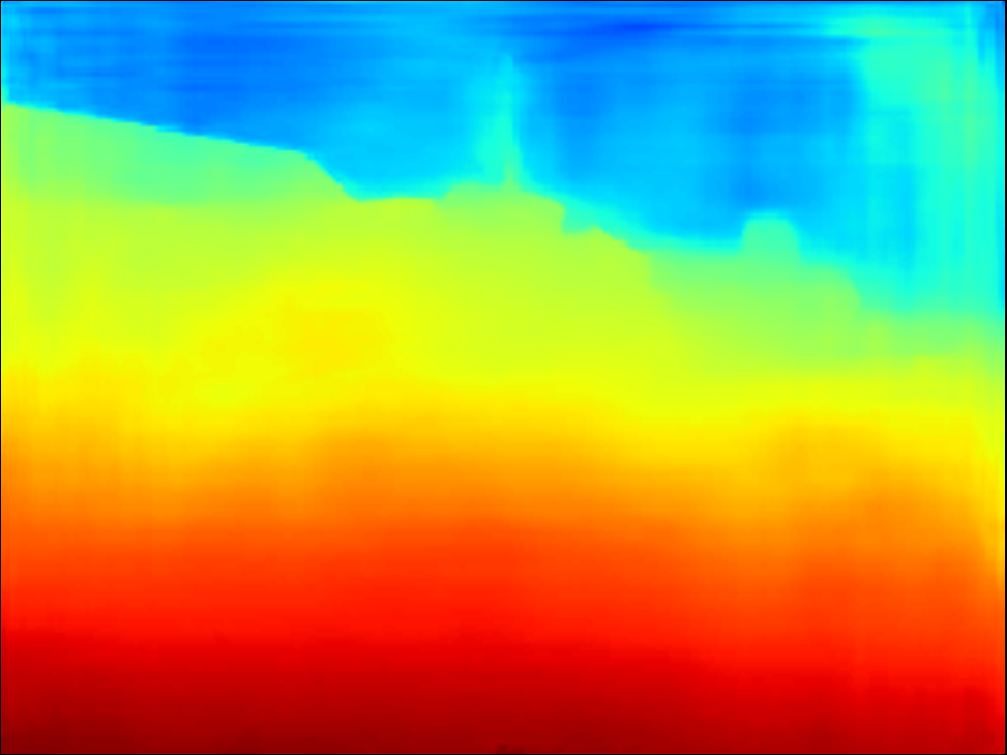}\\
		\includegraphics[width=1.0\textwidth]{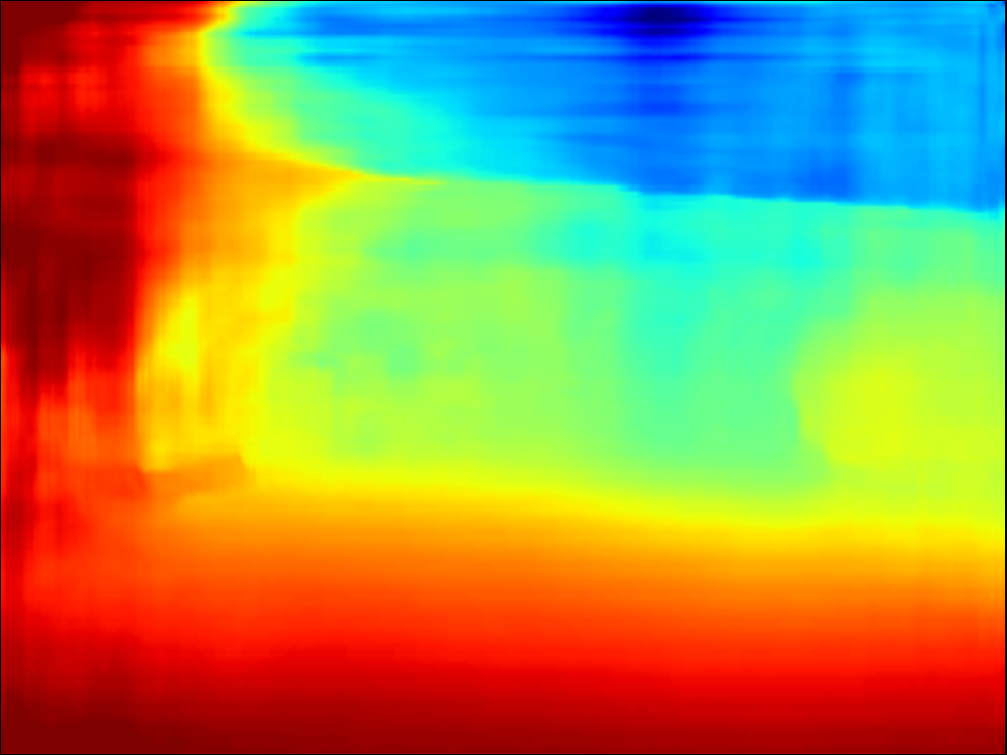}\\
		\includegraphics[width=1.0\textwidth]{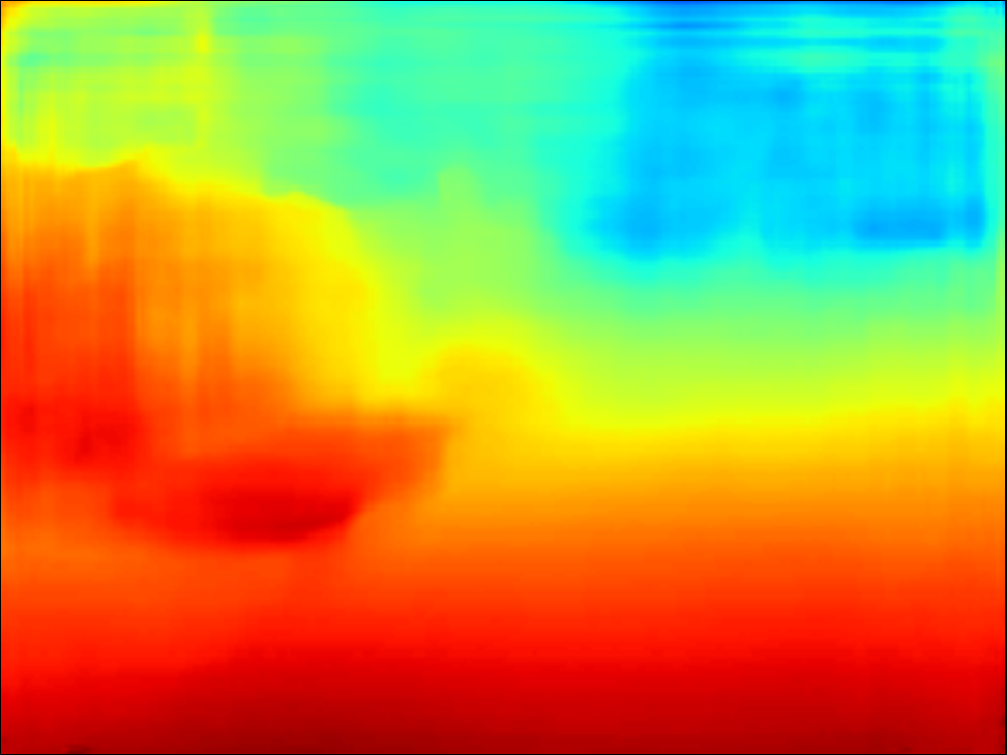}
	\end{tabular}
	}
\end{minipage}
\begin{minipage}{0.14\hsize}
	\centering
	\subfloat[]{
	\begin{tabular}{c}
		\includegraphics[width=1.0\textwidth]{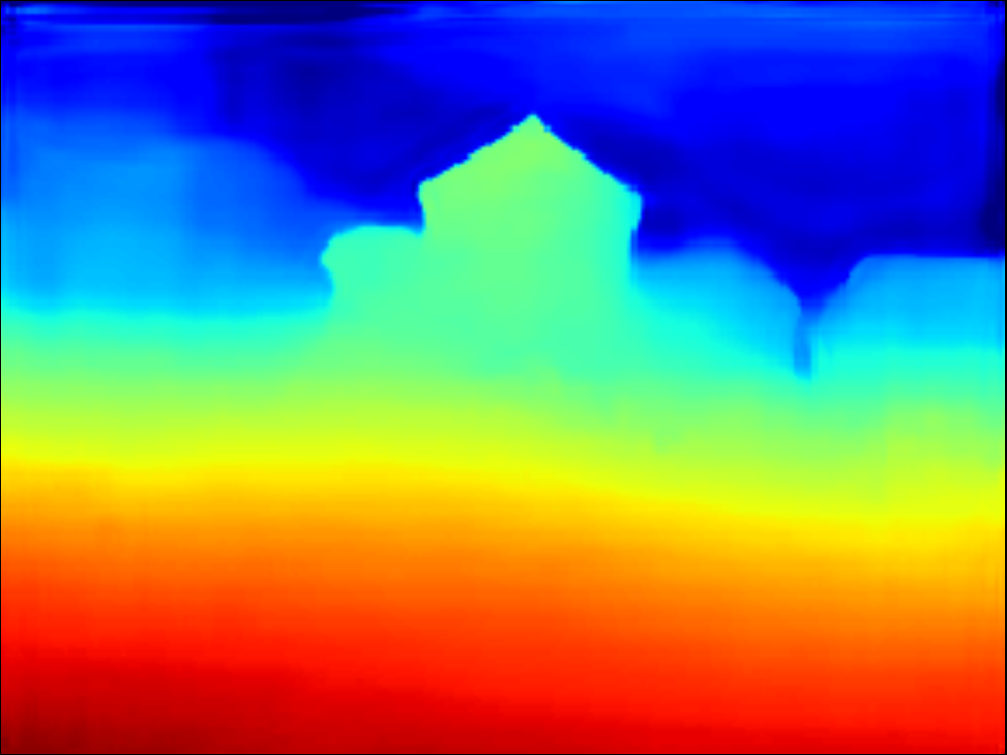}\\
		\includegraphics[width=1.0\textwidth]{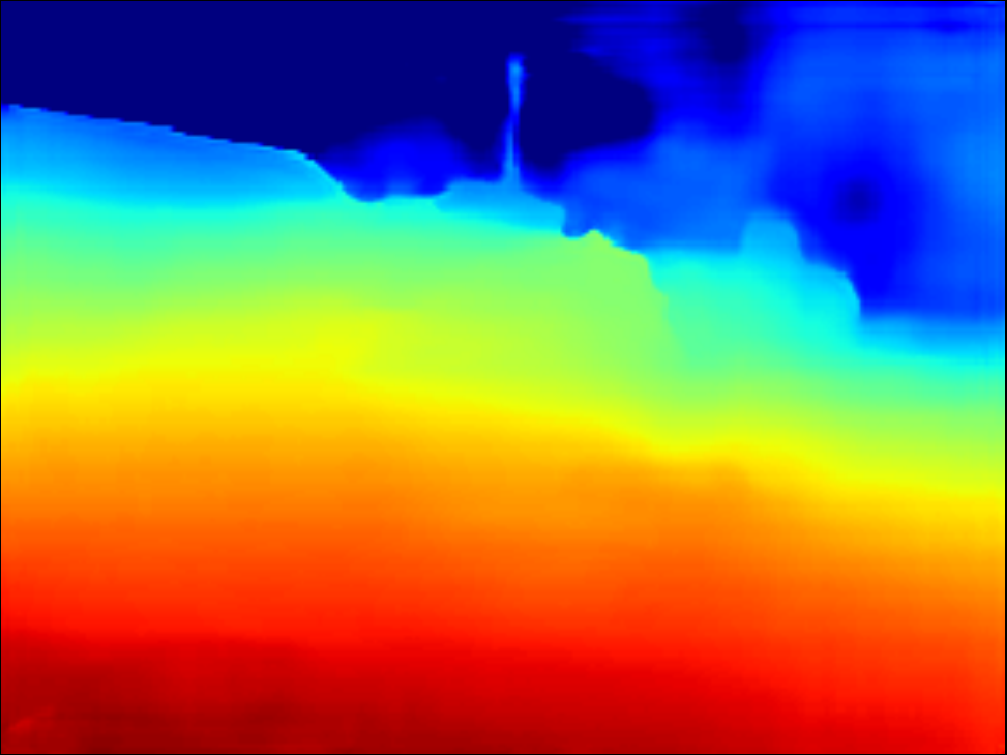}\\
		\includegraphics[width=1.0\textwidth]{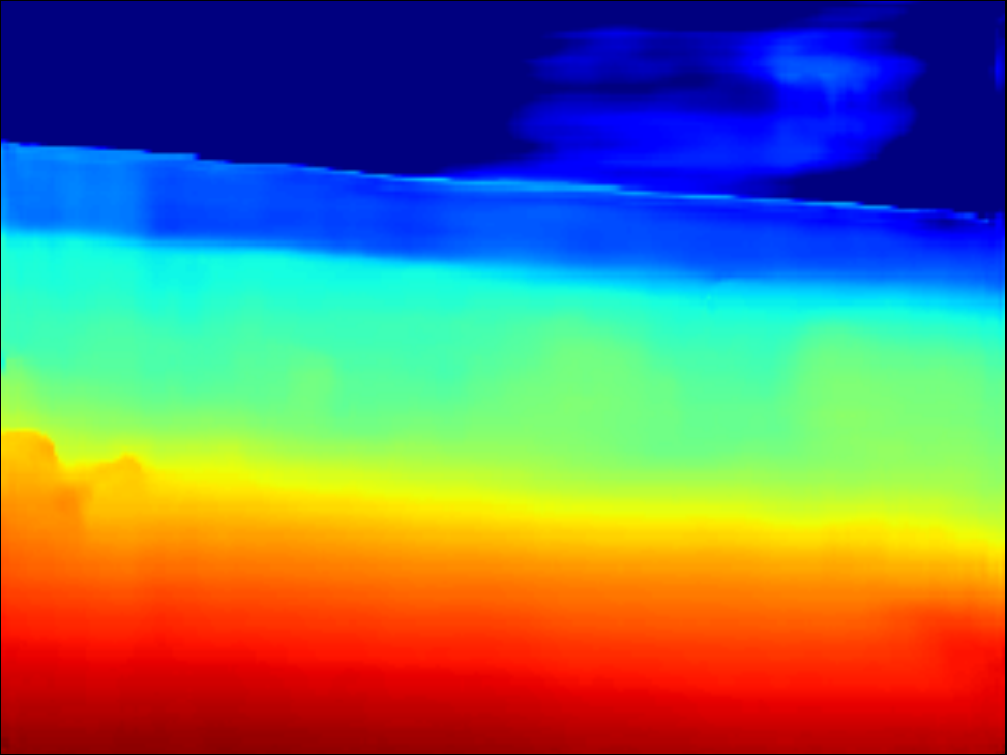}\\
		\includegraphics[width=1.0\textwidth]{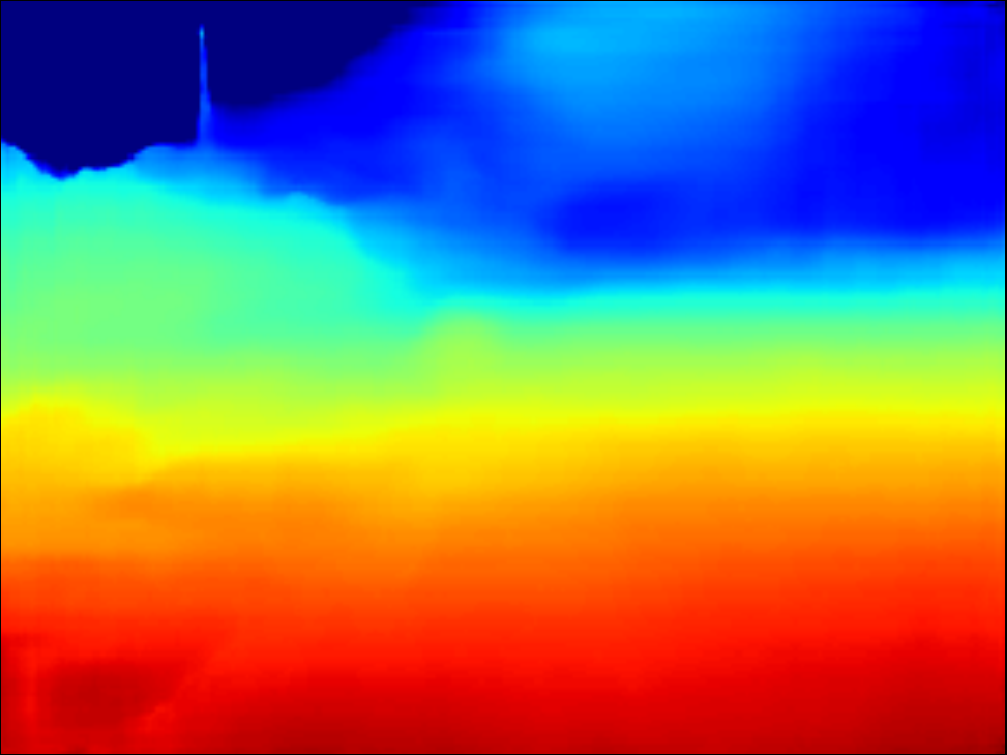}
	\end{tabular}
	}
\end{minipage}
\begin{minipage}{0.14\hsize}
	\centering
	\subfloat[]{
	\begin{tabular}{c}
		\includegraphics[width=1.0\textwidth]{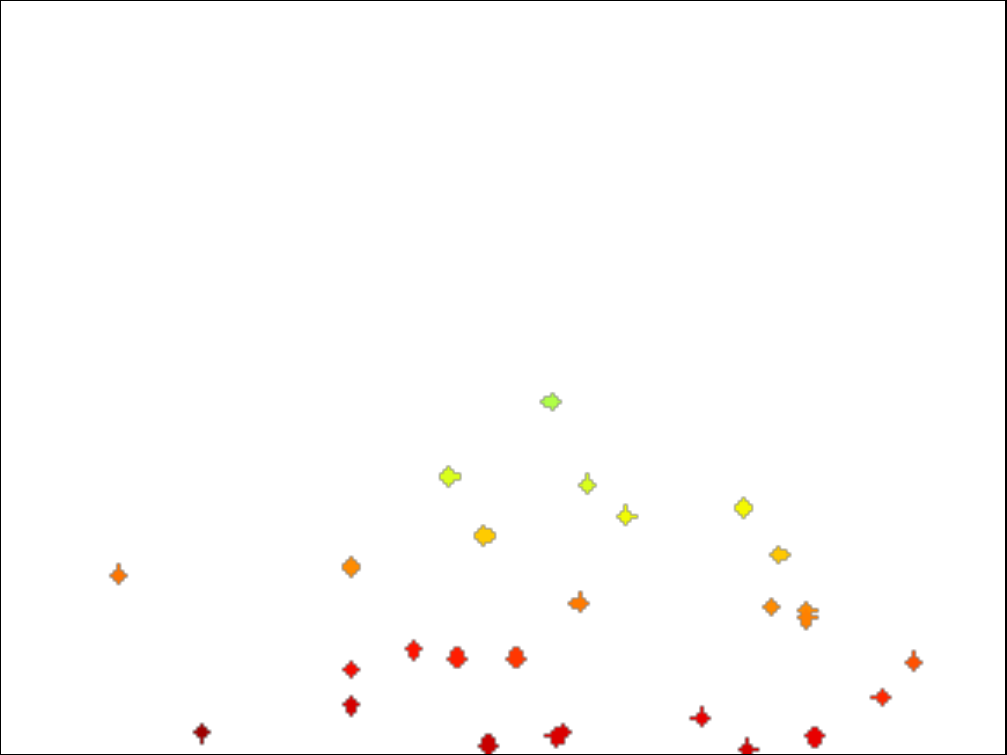}\\
		\includegraphics[width=1.0\textwidth]{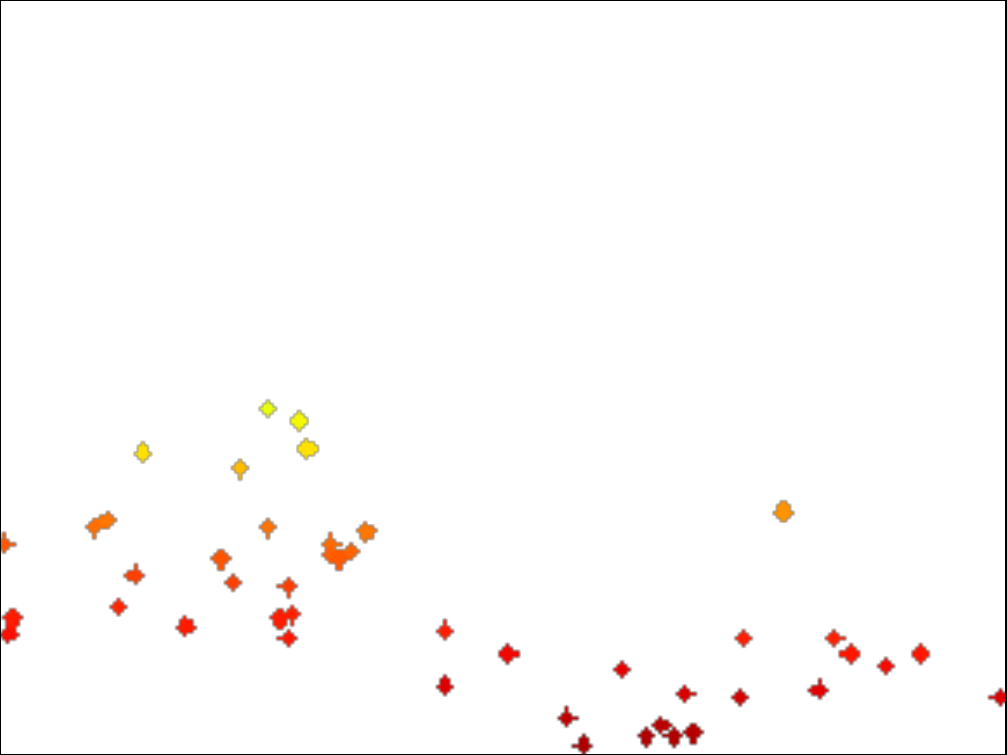}\\
		\includegraphics[width=1.0\textwidth]{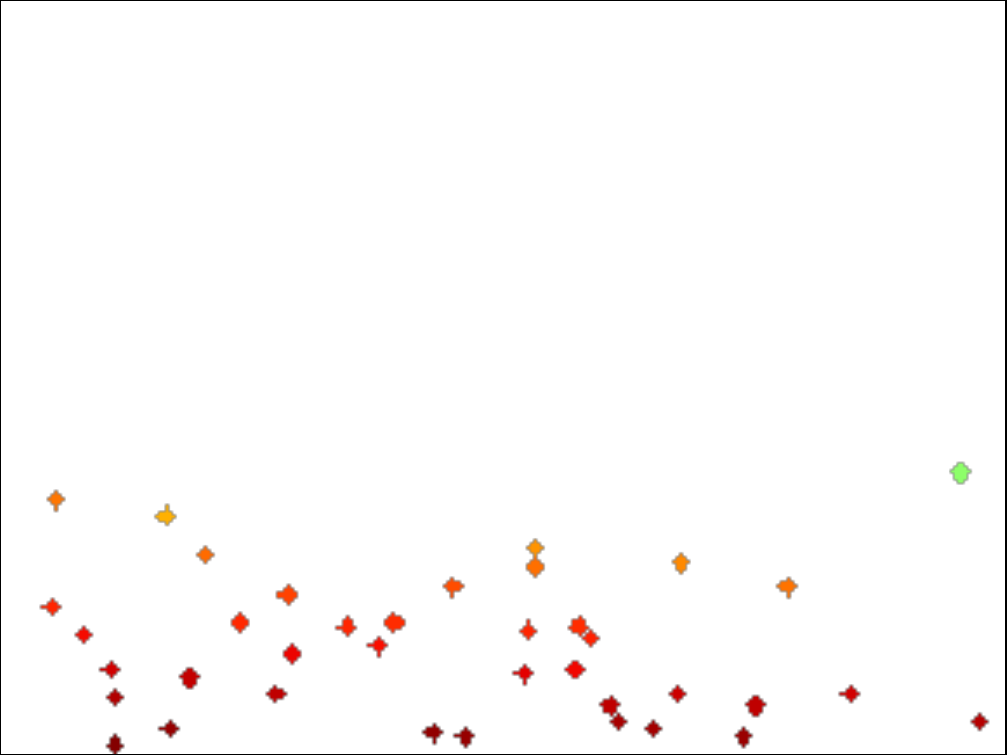}\\
		\includegraphics[width=1.0\textwidth]{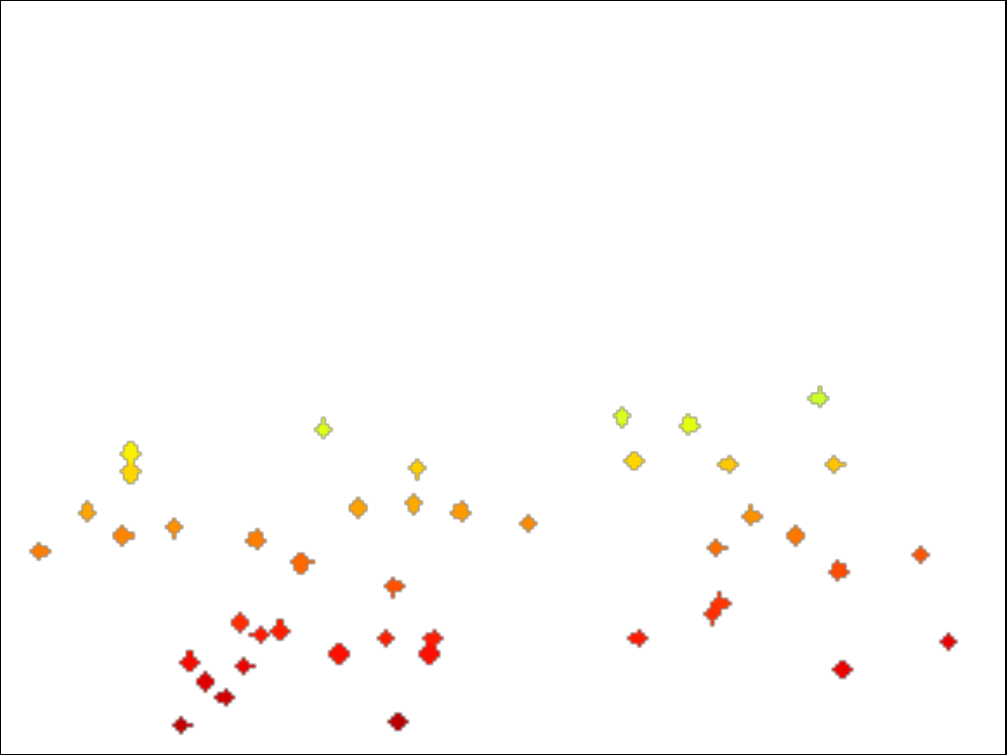}
	\end{tabular}
	}
\end{minipage}
\caption{Experimental results on our captured data in actual foggy scenes. (a) input reference image, (b) input source image, (c) output of DPSNet \cite{im19}, (d) output of fine-tuned MVDepthNet \cite{wang18}, (e) output of proposed method with scattering parameter estimation, and (f) sparse depth obtained by SfM.}
\label{fig:shirouma_result}
\end{figure*}

\subsection{Evaluation of dehazing cost volume}
We first evaluated our dehazing cost volume with ground-truth scattering parameters.
Table \ref{tab:results} shows the quantitative evaluation.
We used four evaluation metrics following Wang and Shen \cite{wang18}: L1-rel is the mean of the relative L1 error between the ground-truth depth and estimated depth, L1-inv is the mean of the L1 error between ground- truth inverse depth and estimated inverse depth, sc-inv is the scale-invariant error of depth proposed by Eigen et al. \cite{eigen14}, and 
correctly estimated depth percentage (C.P.) \cite{tateno17} is the percentage of pixels whose relative L1 error is within 10$\%$.
The red and blue values are the best and second-best, respectively. 

The proposed method (MVDepthMet w/ dcv, where ``dcv'' denotes our dehazing cost volume) was compared with MVDepthNet \cite{wang18} fine-tuned on hazy images (MVDepthNet), simple sequential methods of dehazing \cite{li17,qin20} and depth estimation with MVDepthNet \cite{wang18} (AOD-Net + MVDepthNet, FFA-Net + MVDepthNet), and DPSNet \cite{im19} trained on hazy images (DPSNet).

In most evaluation metrics, the proposed method outperformed the fine-tuned MVDepthNet, demonstrating the effectiveness of our dehazing cost volume.
For the RGB-D SLAM dataset, the fine-tuned MVDepthNet was comparable to the proposed method.
This is because many scenes in the RGB-D SLAM dataset are close to a camera.
In such case, the degradation of an observed image is small and exists uniformly in the image, which has little effect on photometric consistency. 

The proposed method also performed better than the sequential methods of dehazing \cite{li17,qin20} and MVDepthNet \cite{wang18}.
Therefore, we can see that the simultaneous modeling of dehazing and 3D reconstruction on the basis of our dehazing cost volume is effective. 
DPSNet \cite{im19} first extracts feature maps from input images, and then constructs a cost volume in the feature space. Thus, the feature extractor might be able to deal with image degradation caused by light scattering. Nevertheless, our dehazing cost volume allows the consideration of image degradation with a simple network architecture.

The output depth of each method is shown in Fig. \ref{fig:results}.
From top to bottom, each row shows the results of the input images in the SUN3D, RGB-D SLAM, MVS, and Scenes11 datasets, respectively.
DPSNet failed to construct correspondence in some scenes, although it has the multi-scale feature extractor.
Note that the results from the Scenes11 dataset indicate that the proposed method can reconstruct the 3D geometry of a distant scene where the image is heavily degraded due to scattering media. 

\subsection{Evaluation of scattering parameter estimation}
Next, we evaluated the proposed method with scattering parameter estimation.
Each sample of the test dataset presented above consists of image pairs.
Parameter estimation requires a 3D point cloud obtained by SfM.
To ensure the accuracy of SfM, which requires high visual overlap between images and a sufficient number of images observing the same objects, we created a new test dataset for the evaluation of the scattering parameter estimation.
From the SUN3D dataset \cite{xiao13}, we selected 68 scenes and extracted 80 frames from each scene.
The resolution of each image is $680 \times 480$.
We cropped the image patch with $512 \times 384$ from the center and downsized the resolution to $256 \times 192$ for the input of the proposed method.
Similar to the previous test dataset, missing regions were compensated with the output of MVDepthNet \cite{wang18}.
The scattering parameters were randomly sampled for each scene, where the sampling ranges were $A \in [0.7,1.0]$ and $\beta \in [0.4,0.8]$.
SfM \cite{schonberger16,schonberger16sfm} was applied to all 80 frames of each scene to estimate a sparse 3D point cloud, and then the proposed method took the image pair as input.
To evaluate the output depth on the ground-truth depth of the original SUN3D dataset, the sparse depth obtained by SfM was rescaled to match the scale of the ground-truth depth, and we used the camera parameters of the original SUN3D dataset.  

For the parameter search, we set the first $\beta$ range as $\beta_{min} = 0.4$ and $\beta_{max}=0.8$ with 10 steps for the grid search.
We then searched for $A$ and $\beta$ with the search range $\Delta_A=0.05$, $\Delta_\beta=0.05$ and $4 \times 4$ steps.
The total number of the forward computation of the network was $26$, and the total computation time was about $15$ seconds in our computational environment.

Table \ref{tab:params_result} shows the quantitative results of depth and scattering parameter estimation.
``MVDepthNet w/ dcv, pe'' denotes the proposed method with scattering parameter estimation.
As the evaluation metric of $A$ and $\beta$, we used mean absolute error (MAE$_A$ and MAE$_\beta$).
To evaluate the effect of the error at the SfM step, we created three test datasets, where the relative L1 error of the sparse SfM depth of the samples is less than $0.1$, $0.2$, and $0.3$, respectively, and show the number of samples in the table.
These results indicate that the proposed method with ground-truth scattering parameters (MVDeptNet w/ dcv) performed the best.
On the other hand, even when we incorporated scattering parameter estimation into the proposed methoed, it outperformed the other methods.
In addition, scattering parameter estimation is robust to the estimation error of the sparse depth at the SfM step since the MAE values for $A$ and $\beta$ did not vary so much for the three datasets with different SfM errors. 

The qualitative results of the following depth estimation after scattering parameter estimation are shown in Fig. \ref{fig:pe_qualitative}.
Figure \ref{fig:pe_qualitative}(f) shows the input sparse depth obtained by SfM.
Compared with the proposed method with ground-truth scattering parameters, the method with the scattering parameter estimation resulted in almost the same output depth.
In the third row in the figure, the left part in the image has slight error because no 3D sparse points were observed in that region.

\subsection{Experiments with actual foggy scenes}
Finally, we give an example of applying the proposed method to actual outdoor foggy scenes.
We used the image sequence {\it bali} \cite{li15} for the actual data.
This data consists of about 200 frames, and we applied the SfM method \cite{schonberger16,schonberger16sfm} to all these frames to obtain camera parameters and a sparse 3D point cloud. 
The proposed method took the estimated camera parameters, a sparse depth, and image pair as input.
We set the search space of the scattering parameter estimation as $\beta_{min} = 0.01$, $\beta_{max} = 0.1$, $\Delta_A = 0.05$, and $\Delta_\beta = 0.01$ with the same step size in the experiments of the synthesized data.

The results are shown in Fig. \ref{fig:fstereo_result}.
The output depths of the proposed method were rescaled to match the scale of the output of \cite{li15}, because the camera parameters were different between these methods.
Compared with \cite{li15}, the proposed method can reconstruct distant region, which have large image degradation due to light scattering, and the other learning-based methods also failed to reconstruct such distant regions.
Moreover, the proposed method could recover less noisy depth maps as a trade-off for loss of small details due to oversmoothing.
The method proposed by Li et al. \cite{li15} requires iterative graph-cut optimization, so it takes a few minutes to estimate depth for one image.
Our method, on the other hand, requires only a few seconds to estimate depth for one reference image after estimating scattering parameters.
Although scattering parameter estimation takes several ten of seconds, if we assume the medium density of a scene is homogeneous, the estimated scattering parameters at a certain frame can be used for another frame without additional parameter estimation.

We also captured a video with a smartphone camera in an actual foggy scene.
Similar to the previous experiments, we applied the SfM method \cite{schonberger16,schonberger16sfm} to all frames.
The proposed method took the estimated camera parameters, a sparse depth, and image pair as input, and the parameters search space was set as the same in the previous experiments.

The results are shown in Fig. \ref{fig:shirouma_result}.
Figures (a) and (b) show the input reference and source images, respectively.
This results also indicate that the proposed method can reconstruct distant regions with large image degradation due to light scattering.
These data are available at our project page \url{https://github.com/yfujimura/DCV-release}.

\section{Conclusion}
We proposed a learning-based MVS method with a novel cost volume, called the dehazing cost volume, which enables MVS methods to be used in scattering media.
Differing from the ordinary cost volume, our dehazing cost volume can compute the cost of photometric consistency by taking into account image degradation due to scattering media.
This is the first paper to solve the chicken-and-egg problem of depth and scattering estimation by computing the scattering effect using each swept plane in the cost volume without explicit scene depth.
We also proposed a method for estimating scattering parameters such as airlight and a scattering coefficient.
This method leverages geometric information obtained at an SfM step, and ensures the correctness of the following depth estimation.
The experimental results on synthesized hazy images indicate the effectiveness of our dehazing cost volume in scattering media.
We also demonstrated its applicability using images captured in actual foggy scenes.
For future work, we will extend the proposed method to depth-dependent degradation, other than light scattering, such as defocus blur \cite{gur19,maximov20}.

\section*{Acknowledgments}
This work was supported by JSPS KAKENHI Grant Number 18H03263 and 19J10003.

{\small
\bibliographystyle{ieee_fullname}
\bibliography{egbib}
}

\end{document}